\documentclass[10pt,twocolumn,letterpaper]{article}

\usepackage[pagenumbers]{cvpr} 

\definecolor{cvprblue}{rgb}{0.21,0.49,0.74}
\usepackage[pagebackref,breaklinks,colorlinks,allcolors=cvprblue]{hyperref}

\usepackage[accsupp]{axessibility}
\usepackage{graphicx}	
\usepackage{amsmath}	
\usepackage{amssymb}	
\usepackage{booktabs}
\usepackage{times}
\usepackage{microtype}
\usepackage{epsfig}
\usepackage{caption}
\usepackage{float}
\usepackage{placeins}
\usepackage{color, colortbl}
\usepackage{stfloats}
\usepackage{enumitem}
\usepackage{tabularx}
\usepackage{xstring}
\usepackage{multirow}
\usepackage{xspace}
\usepackage{url}
\usepackage{subcaption}
\usepackage{color}
\usepackage{xcolor}
\usepackage[hang,flushmargin]{footmisc}
\usepackage{pifont}
\usepackage{listings}
\usepackage{hyperref}
\usepackage{mathtools}
\usepackage{fontawesome5}
\usepackage{twemojis}
\usepackage{tikz}

\newif\ifreview 
\newif\ifarxiv 
\newif\ifcamera 
\newif\ifrebuttal 

\ifcamera \usepackage[accsupp]{axessibility} \fi
\ifarxiv  \fi

\newcommand{\nbf}[1]{{\noindent\textbf{#1}.}}
\newcommand{\cmark}{\textcolor{green!60!black}{\ding{51}}}
\newcommand{\xmark}{\textcolor{red!80!black}{\ding{55}}}

\colorlet{punct}{red!60!black}
\definecolor{background}{HTML}{F7F7F7}
\definecolor{delim}{RGB}{20,105,176}
\colorlet{cmnt}{magenta!60!black}

\newcommand{\myfootnote}[1]{\footnotemark[6]}
\lstdefinelanguage{json}{
    basicstyle=\normalfont\ttfamily,
    stepnumber=1,
    numbersep=8pt,
    showstringspaces=false,
    breaklines=true,
    frame=lines,
    backgroundcolor=\color{background},
    morecomment=[s]{/*}{*/},
    commentstyle=\color{cmnt}\ttfamily,
    literate=
     *{:}{{{\color{punct}{:}}}}{1}
      {,}{{{\color{punct}{,}}}}{1}
      {\{}{{{\color{delim}{\{}}}}{1}
      {\}}{{{\color{delim}{\}}}}}{1}
      {[}{{{\color{delim}{[}}}}{1}
      {]}{{{\color{delim}{]}}}}{1},
}

\def\paperID{} 
\def\confName{CVPR}
\def\confYear{2026}

\title{
SciGA: A Comprehensive Dataset \\ for Designing Graphical Abstracts in Academic Papers
}

\author{Takuro Kawada, \quad Shunsuke Kitada, \quad Sota Nemoto, \quad Hitoshi Iyatomi\\
Hosei University, Tokyo, Japan\\
{\tt\small takuro.kawada@gmail.com, info@shunk031.me, sota.nemoto.5s@gmail.com, iyatomi@hosei.ac.jp}\\
{\small \href{https://iyatomilab.github.io/SciGA}{https://iyatomilab.github.io/SciGA}}
}

\begin{document}

\maketitle

\begin{tikzpicture}[remember picture,overlay]
  \node[anchor=north west] at ([xshift=2.7cm,yshift=-3.1cm]current page.north west)
    {\includegraphics[width=2cm]{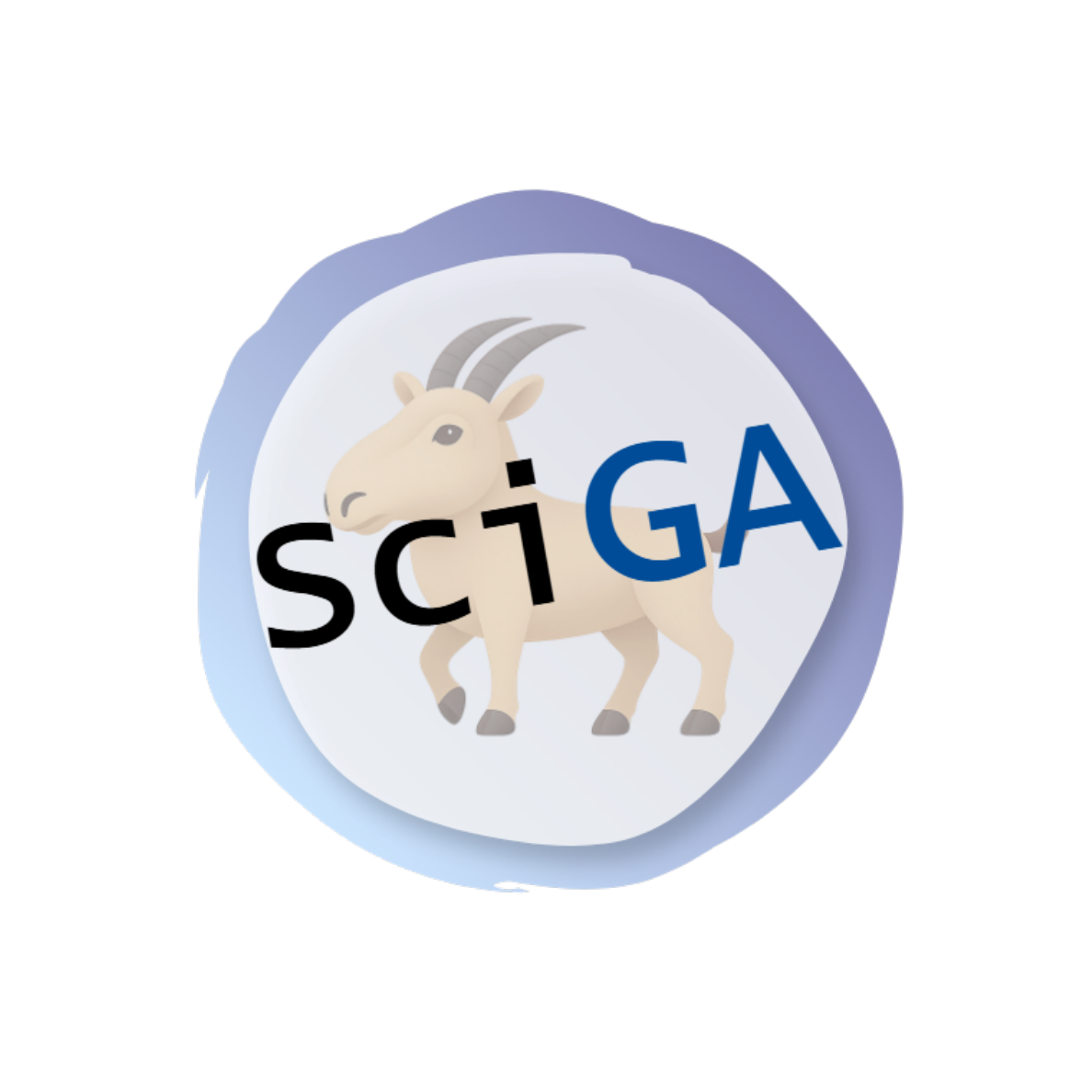}};
\end{tikzpicture}

\begin{abstract}

Graphical Abstracts (GAs) play a crucial role in visually conveying the key findings of scientific papers. Although recent research increasingly incorporates visual materials such as \textit{Figure~1} as \textit{de facto} GAs, their potential to enhance scientific communication remains largely unexplored. Designing effective GAs requires advanced visualization skills, hindering their widespread adoption. To tackle these challenges, we introduce SciGA-145k, a large-scale dataset comprising approximately 145,000 scientific papers and 1.14 million figures, specifically designed to support GA selection and recommendation, and to facilitate research in automated GA generation. As a preliminary step toward GA design support, we define two tasks: 1) Intra-GA Recommendation, identifying figures within a given paper well-suited as GAs, and 2) Inter-GA Recommendation, retrieving GAs from other papers to inspire new GA designs. Furthermore, we propose Confidence Adjusted top-1 ground truth Ratio (CAR), a novel recommendation metric for fine-grained analysis of model behavior. CAR addresses limitations of traditional rank-based metrics by considering that not only an explicitly labeled GA but also other in-paper figures may plausibly serve as GAs. Benchmark results demonstrate the viability of our tasks and the effectiveness of CAR. Collectively, these establish a foundation for advancing scientific communication within AI for Science. \protect \footnotemark[1]

\vspace{-6mm}

\end{abstract}

\footnotetext[1]{
    Code and data is available at:
    \faGithub~\href{https://github.com/IyatomiLab/SciGA}{IyatomiLab/SciGA},
    \raisebox{-0.16em}{\twemoji[height=1.0em]{1f917}} \href{https://huggingface.co/datasets/iyatomilab/SciGA}{iyatomilab/SciGA}
}

\section{Introduction}
\label{sec:introduction}

Scientific discovery and communication of its findings are fundamental to advancing knowledge, yet both are often constrained by researchers' limited resources, such as background knowledge or time.
Historically, research has focused on automating the discovery process to accelerate knowledge generation~\cite{lenat1977math,lenat1983am,buchanan1978dendral}.
More recently, the emergence of AI-driven approaches in science has gained significant attention, driving applications in research automation, including hypothesis generation~\cite{merchant2023scaling, pyzer-Knapps2022accelerating, meincke2023llm-idea-gen} and experimental design~\cite{szymanski2023auto-lab, baek2025research-agent}.
While scientific discovery progresses through automation, communicating research findings remains an equally critical challenge.
Recent advancements in AI-assisted paper writing~\cite{wen2024overleafcopilot, lu2024ai-scientist} and presentation material generation~\cite{fu2022doc2ppt, pang2025paper2poster} have improved the efficiency of scientific communication.
However, effectively conveying complex research ideas visually remains an open challenge.

\begin{figure}[!t]
    \centering
    \includegraphics[width=\linewidth]{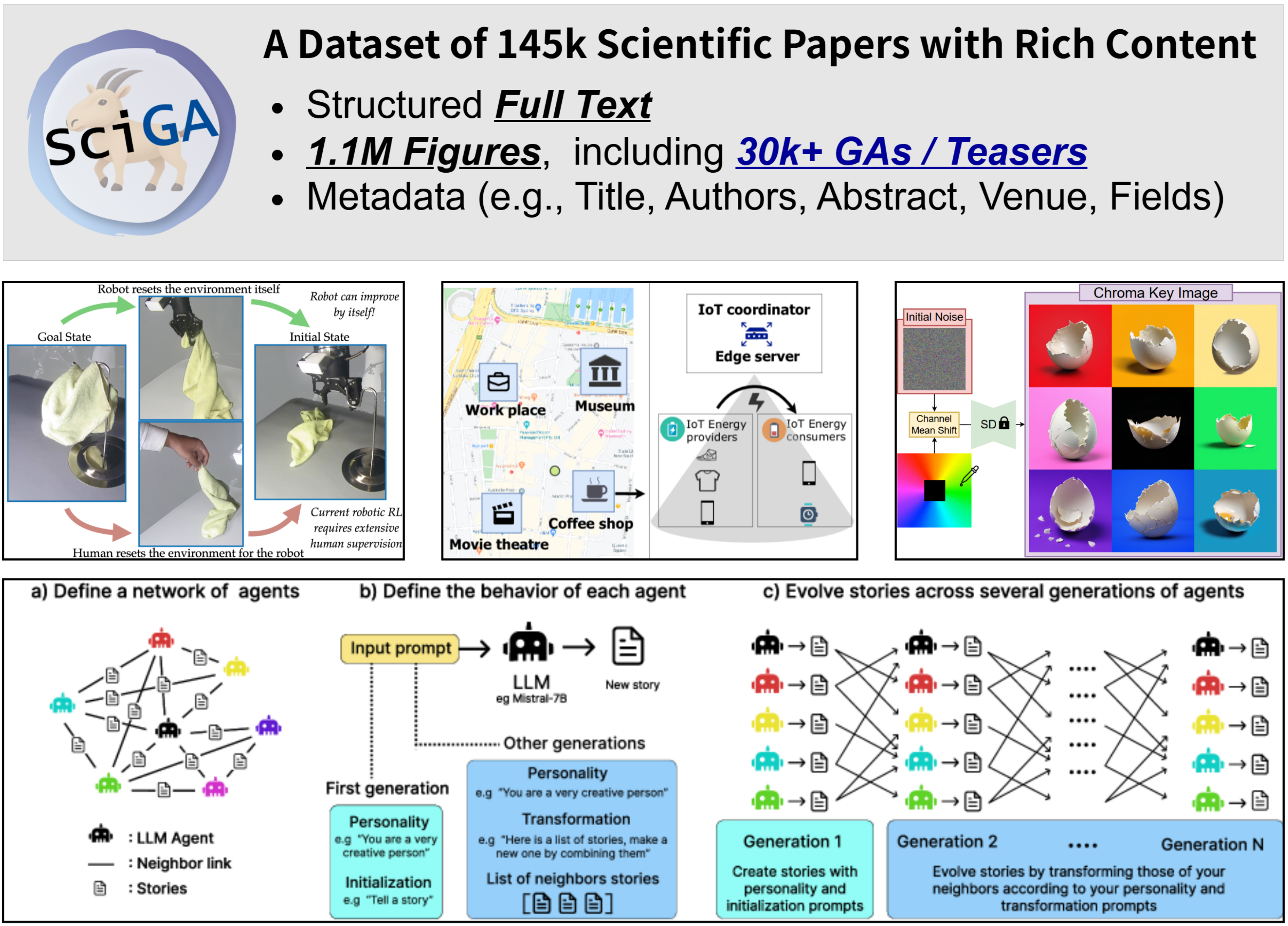}
    \vspace{-6mm}
    \caption{
        Example GAs or teasers in our SciGA-145k.\protect \footnotemark[2]
        These visual summaries highlight key contributions of scientific papers.
        In addition, SciGA-145k provides access to structured full texts, rich metadata, and non-GA figures, enabling comprehensive visual-language analysis in AI for Science.
    }
    \label{fig:GA_example}
    \vspace{-2mm}
\end{figure}

\footnotetext[2]{
\footnotesize
\begin{tabular}[t]{@{}ll@{}}
arXiv ID: 
\href{https://arxiv.org/abs/2303.01488}{2303.01488},
\href{https://arxiv.org/abs/2111.06064}{2111.06064},
\href{https://arxiv.org/abs/2411.15580}{2411.15580},
\href{https://arxiv.org/abs/2403.08882}{2403.08882}
\end{tabular}
}

\begin{figure*}[t]
    \centering
    \resizebox{0.95\textwidth}{!}{
    \begin{minipage}{0.49\linewidth}
        \centering
        \includegraphics[width=\textwidth]{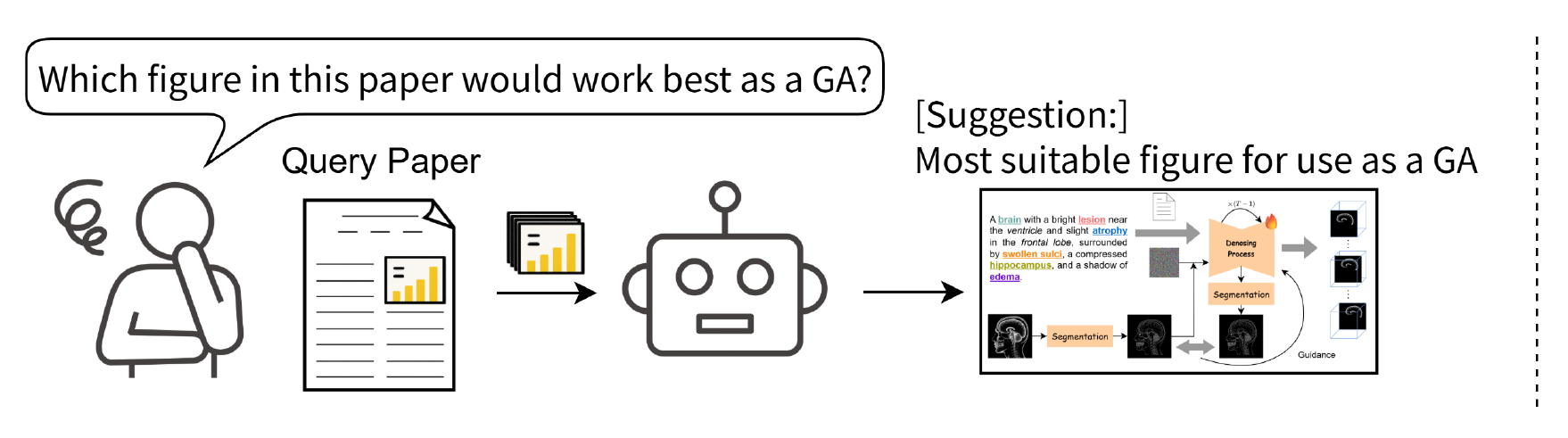}
        \vspace{-7mm}
        \subcaption{Intra-GA Recommendation}
        \label{fig:task-overview:a}
    \end{minipage}
    \begin{minipage}{0.49\linewidth}
        \centering
        \includegraphics[width=\textwidth]{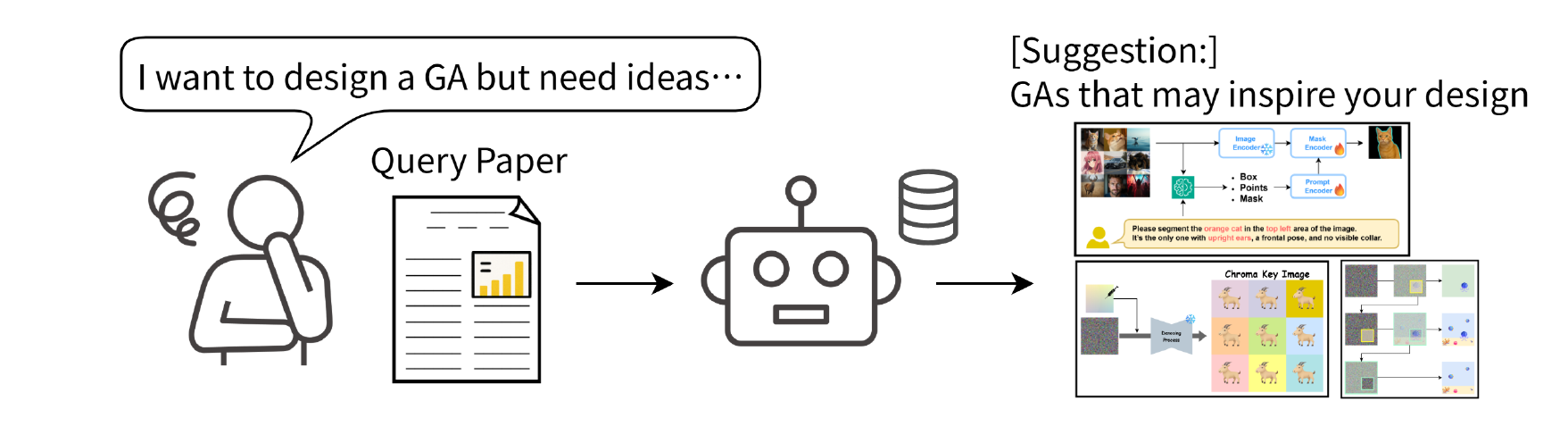}
        \vspace{-7mm}
        \subcaption{Inter-GA Recommendation}
        \label{fig:task-overview:b}
    \end{minipage}
    }
    \vspace{-2mm}
    \caption{
        Overview of our two GA design support tasks.
        (a) Intra-GA Recommendation identifies the most suitable figure as its GA within a paper.
        It supports authors who already have figures but are unsure which best represents their work, and promotes the use of visual summaries even on academic platforms that do not require explicit GA submission.
        (b) Inter-GA Recommendation retrieves GAs or teasers from other papers to inspire new designs.
        It supports authors during the design process by providing concrete references tailored to the paper’s topic and style.  
        Together, these tasks promote effective visual communication.
    }
    \label{fig:task_overview}
    \vspace{-2mm}
\end{figure*}

\begin{figure*}[t]
    \centering
    \includegraphics[width=0.95\linewidth]{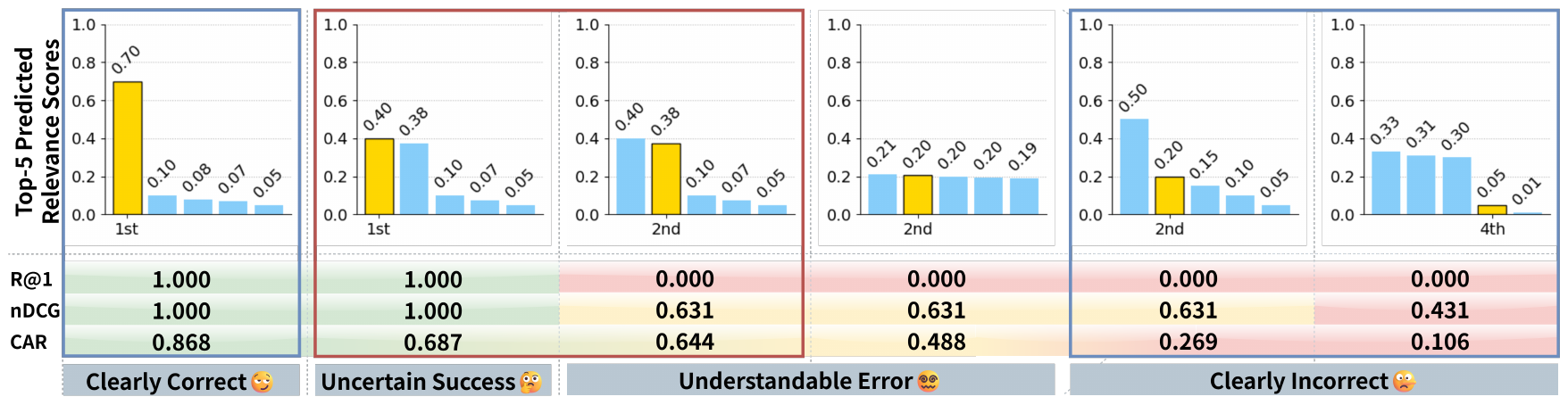}
    \vspace{-2mm}
    \caption{
        Illustration of CAR, our proposed recommendation metric.
        Each column shows predicted top-5 scores, with the GT highlighted in yellow.
        CAR assigns partial credit to understandable errors, similar to cases where the outcome is uncertain but still successful (red box), and evaluates clearly correct or incorrect predictions appropriately (blue box).
        Unlike R@$k$ or nDCG, CAR assigns instance-level continuous scores without graded labels, based on the full score distribution, not just GT rank.
    }
    \label{fig:CAR_example}
    \vspace{-2mm}
\end{figure*}

Graphical Abstracts (GAs) have emerged as a crucial tool for visually communicating key findings of scientific papers.
Here, we define GAs as journal-hosted visual summaries submitted alongside papers, distinguishing them from in-paper figures.
In recent years, researchers frequently use teasers (\textit{Figure~1}, prominently displayed, often full-width, on or near the first page and cited in the \textit{Introduction}) as de facto GAs, even in cases where a formal GA is not adopted.
They contribute to greater online impact~\cite{bennett2023ga}.
Despite this growing trend, methodologies for effectively designing and utilizing such visual materials remain underdeveloped.
Furthermore, creating compelling GAs requires advanced skills in visualizing key contributions~\cite{lee2023ga, jeyaraman2023attract}, which is challenging for many researchers.

To address the gaps identified above, we introduce SciGA-145k, the first large-scale dataset designed to support GA design.
SciGA-145k comprises approximately 145,000 scientific papers (including full text and metadata) and 1.14 million figures (including both in-paper figures and GAs), released under the C-UDA 1.0 license.
The dataset preserves rich structural elements such as section hierarchies, subfigure structures, math expressions, and semantic tags, enabling downstream tasks on GAs and other visual-language content in scientific papers.
We collected GAs published alongside papers in journals and identified teasers from in-paper figures.
\cref{fig:GA_example} shows example GAs or teasers in SciGA-145k.

To promote scientific communication, we define two tasks:
1) Intra-GA Recommendation, identifying in-paper figures that effectively capture key contributions as GAs;
and 2) Inter-GA Recommendation, retrieving GAs from other papers to inspire new GA designs (see \cref{fig:task_overview}).
A successful solution to Intra-GA Recommendation would allow academic platforms (e.g., journals, preprint servers) to suggest alternative GA options, such as embedding them when linking papers on social media.
Meanwhile, a successful solution to Inter-GA Recommendation would support researchers in designing more impactful GAs by leveraging existing designs from other papers.
We evaluated a group of representative models based on classification, including SwinTransformerV2~\cite{liu2022swin-transformer}, ConvNeXtV2~\cite{woo2023convnext}; and contrastive learning, including CLIP~\cite{radford2022clip}, Long-CLIP~\cite{zhang2024long-clip}, SigLIP2~\cite{tschannen2025siglip2}.

Conventional recommendation metrics such as recall@$k$~(R@$k$), or normalized discounted cumulative gain~(nDCG)~\cite{burges2005ndcg} have failed to account for scenarios in Intra-GA Recommendation where multiple plausible candidates exist beyond the ground truth~(GT).
To address these limitations, we introduce Confidence Adjusted top-1 GT Ratio@$k$~(CAR@$k$), a novel metric that considers the model's confidence, as illustrated in \cref{fig:CAR_example}.

Our contributions are summarized as follows:
\begin{itemize}
    \item We introduce SciGA-145k, the first dataset explicitly designed for GA design support, providing a foundation for advancing scientific communication through GA research.
    \item We define Intra/Inter-GA Recommendation, two complementary tasks that facilitate broader adoption of GA-based scientific communication and support researchers refining their visual abstracts for enhanced clarity and impact.
    \item We propose CAR, a novel recommendation metric for scenarios where multiple items beyond the explicitly GT may serve as viable alternatives, offering a more nuanced view of model performance in handling soft relevance. 
\end{itemize}

\begin{table*}[t]
    \caption{
        Comparison of SciGA-145k with existing scientific paper datasets. 
        Our dataset uniquely provides full-text content, comprehensive figure support, and explicit GA/teaser annotations, features missing in previous datasets.
        CentralFigure~\cite{yang2019identifying} includes teaser captions and their referenced textual contexts (7,295 entries), but does not contain the corresponding figure images or broader contextual information,  limiting its use for visual or structural analysis.  
        With 145k papers and 1.1M figures, SciGA-145k offers the complete foundation needed for advancing scientific visual communication research.
    }
    \label{tbl:datasets}
    \centering
    \small
    \resizebox{0.8\linewidth}{!}{
        \begin{tabular}{lccrrlrr}
            \toprule
            & \multicolumn{2}{c}{\textbf{Support Contents}} & \multicolumn{2}{c}{\textbf{Annotations}} & \multirow{2}{*}{\textbf{Source Format}} & \multirow{2}{*}{\textbf{\#Papers}} & \multirow{2}{*}{\textbf{\#Figures}}\\
            \cmidrule(lr){2-3} \cmidrule(lr){4-5}
            & Full-text & Figures & \#GA & \#Teaser & & \\
            
            \midrule
            
            S2ORC~\cite{lo2020s2orc}                        & \cmark & \xmark & N/A & N/A    & PDF\&HTML & 81,100k &    N/A \\
            unarXive 2022~\cite{saier2023unarxive}          & \cmark & \xmark & N/A & N/A    & PDF       &  1,900k &    N/A \\
            Paper2fig100k~\cite{rodriguez2023ocr-vggan}     & \xmark & \cmark & N/A & N/A    & \TeX      &     69k &   102k \\
            ArxivCap~\cite{li2024multi-arxiv}               & \xmark & \cmark & N/A & N/A    & \TeX      &    572k & 6,400k \\
            CentralFigure~\cite{yang2019identifying}        & \xmark & \xmark & N/A & 7,295  & PDF       &      7k &    N/A \\
            MMSci~\cite{li2023mmsci}                        & \cmark & \cmark & N/A & N/A    & HTML      &    131k &   742k \\
            
            \cmidrule(lr){1-1}
            \cmidrule(lr){2-2}
            \cmidrule(lr){3-3}
            \cmidrule(lr){4-4}
            \cmidrule(lr){5-5}
            \cmidrule(lr){6-6}
            \cmidrule(lr){7-7}
            \cmidrule(lr){8-8}
            
            \textbf{SciGA-145k (ours)}                      & \cmark & \cmark & 309 & 30,724 & HTML      &  145k   & 1,100k \\
            
            \bottomrule
        \end{tabular}
    }

    \vspace{-2mm}

\end{table*}

\section{Related Work}
\label{sec:related-work}

\nbf{The Importance of GA and its Design}
GAs enhance Altmetric Attention Scores and social media engagement, acting as effective entry points to scientific content~\cite{kim2022Seeing, huang2018the-effect, kunze2021infographics, bennett2023ga, hoffberg2020beyond, chapman2019randomizad, ibrahim2017va}.
At the same time, overly abstracted GAs can lead to misinterpretations~\cite{jeyaraman2023ga, jeyaraman2023attract}.
While structured design guidelines~\cite{millar2022the-role} and reusable design patterns~\cite{jeyaraman2023ga, jeyaraman2023attract} have been explored, these are often too rigid to address the diverse needs across research contexts.
Furthermore, automatic GA generation has been suggested as a potential future direction~\cite{lee2023ga, kirukowski2023potential}, but its development remains limited due to the lack of accessible, large-scale GA collections, especially those restricted behind journal paywalls.
Our work pioneers a data-driven approach to GA research, laying the groundwork for systematic exploration.
Rather than imposing static design guidelines, we adopt a flexible, recommendation-based framework.
To support this, we collect both GAs from open-access journals and teasers, which we treat as de facto GAs, from within papers.
This broader scope enhances accessibility and diversity for large-scale analysis.

\nbf{Scientific Visual Communication}
Recent studies support scientific communication by generating figures via diffusion models~\cite{rodriguez2023fig-gen} and automating the creation of slides~\cite{fu2022doc2ppt, zheng2025pptagent} or posters~\cite{qiang2019learning, tanaka2024sci-post-layout, rodriguez2023fig-gen}.
Central-figure selection~\cite{yang2019identifying, yamamoto2021visual} and multimodal summarization systems~\cite{yamamoto2019automatic, tan2025enhancing} both tackle selecting a representative figure, comparable to a teaser in our setting.
However, existing problem formulations remain coarse:
central-figure selection reduces the task to choosing a single figure within a paper based largely on textual cues, assuming a unique ground-truth and overlooking multiple plausible candidates; 
meanwhile, summarization pipelines treat figure choice only implicitly, lacking an explicit candidate set or a principled ranking formulation suited for GA design support.
As a result, neither line of work captures the multimodal and ambiguous nature of GA selection.
We reinterpret these approaches as precursors to Intra-GA Recommendation and generalize them into a multimodal ranking task by introducing a novel metric, CAR, suited to recommendation settings.
Finally, we introduce Inter-GA Recommendation as a complementary task that supports more design-oriented and creative use cases.

\section{Proposed Dataset, Tasks, and Metric}
\label{sec:proposed}


\subsection{SciGA-145k Dataset}
\label{sec:proposed:sciga}

SciGA-145k is the largest publicly available scientific paper dataset with both full texts and figures.
Our dataset is the first to provide enriched annotations for GAs and teaser images, facilitating research on scientific visual communication.
As summarized in \cref{tbl:datasets}, prior datasets often lack one or more of the following key components: full text, figures, or GA-related annotations.
SciGA-145k addresses these limitations by offering a comprehensive source for GA design research.

SciGA-145k comprises 145,080 seed papers and 1,148,882 figures, collected from articles published between January 2021 and November 2025 on arXiv.
Each entry includes the full text, figure captions, and metadata such as titles, authors, submission dates, research fields, author comments, DOIs, published journals, and accepted conferences.
SciGA-145k preserves section hierarchies, subfigure compositions, mathematical expressions, footnotes, and tags.
These elements are encapsulated with special tokens (\texttt{<MATH>}, \texttt{<NOTE>}, \texttt{<TAG>}) to facilitate preprocessing and accurate information extraction.
Research fields are categorized using the arXiv hierarchical taxonomy,\footnotemark[3] which assigns at least one primary subject to every paper in our dataset.
For detailed data structures and statistics, refer to the Appendix.
\footnotetext[3]{
\href{https://arxiv.org/category_taxonomy}{https://arxiv.org/category\_taxonomy}
}

Among these papers, we identified 309 with GAs and 30,724 with teasers, far exceeding the sample sizes used in prior GA studies.
These were obtained through exhaustive scanning of the entire corpus:
GAs were collected via journal versions linked from arXiv metadata (e.g., IEEE Access, Nature Chemistry), 
while teasers were automatically identified by analyzing \TeX{}-derived HTML structures.

\nbf{Data Collection}
Textual data in SciGA-145k, including abstracts, full texts, and figure captions, was obtained from HTML renderings of \TeX\ sources using the ar5iv:04:2024 dataset~\cite{ginev2024ar5iv}.
In-paper figures were collected either automatically from the HTML pages or manually from the original \TeX\ sources when the HTML was incomplete or corrupted.
When available via an open-access journal version of the same paper, GAs were manually collected.
Although extremely rare, a small number of GAs in video format, recently adopted by some journals, are also included for completeness.
Metadata was extracted via the arXiv API, and author comments were analyzed to identify papers accepted at major international conferences in computer vision, natural language processing, and machine learning, as well as to distinguish between main track and workshop submissions.

\begin{figure}[!t]
    \centering
    \begin{minipage}{0.568\linewidth}
        \centering
        \includegraphics[width=\textwidth]{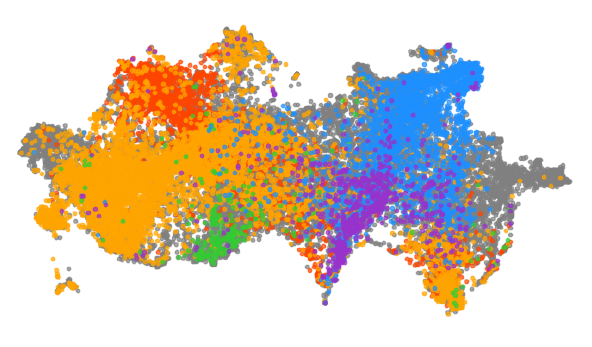}
        \vspace{-5mm}
        \subcaption{GAs / Teasers}
        \label{fig:embed:a}
    \end{minipage}
    \begin{minipage}{0.412\linewidth}
        \centering
        \includegraphics[width=\textwidth]{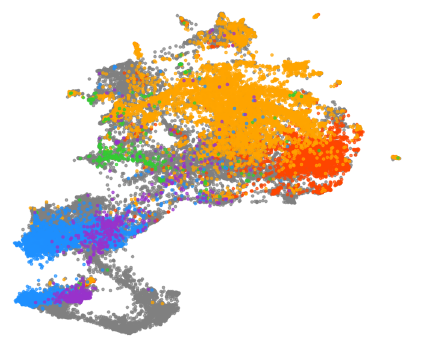}
        \vspace{-5mm}
        \subcaption{Abstracts}
        \label{fig:embed:b}
    \end{minipage}
    \caption{
        Visualization of embeddings colored by research field:
        \textcolor[HTML]{FFA500}{yellow} for Computer Vision,
        \textcolor[HTML]{FF4500}{red} for Computation and Language,
        \textcolor[HTML]{32CD32}{green} for Networking and Internet Architecture,
        \textcolor[HTML]{1E90FF}{blue} for Condensed Matter Physics,
        \textcolor[HTML]{9932CC}{purple} for Mathematics,
        and \textcolor{gray}{gray} for the others.
    }
    \label{fig:embed}
    \vspace{-2mm}
\end{figure}

\nbf{Distribution of Research Domain}
\cref{fig:embed} visualizes the distribution of embedding points, each representing a paper's GA or abstract, mapped using CLIP~\cite{radford2022clip} and projected with UMAP~\cite{mcinnes2018umap}, with colors representing different research fields.
The observed clustering patterns reveal significant field-dependent variations in GA design and abstract writing styles, demonstrating that these design trends exhibit clear distinctions across research domains, a perspective that has not been systematically explored
in prior studies~\cite{jeyaraman2023ga}.
For example, physics GAs often show experimental setups, while computer science ones highlight model architectures.

\subsection{Task Definition for GA Design Support}
\label{sec:proposed:task-definition}

We define two tasks, Intra-GA Recommendation and Inter-GA Recommendation, to support GA design using SciGA-145k.
Let $\smash{\mathcal{D} = \{ d {\scriptstyle ^{(i)}} \mid i \in \{1,2,\dots,N \}\}}$ be the set of $N$ target papers.
Each paper $\smash{d {\scriptstyle ^{(i)}}}$ consists of various components, including body text, $\smash{n {\scriptstyle ^{(i)}}}$ figures $\smash{\{ I {\scriptstyle _j^{(i)}} \mid j \in \{ 1, 2, \dots, n {\scriptstyle ^{(i)}}\} \}}$, and captions, all of which can be utilized for the tasks.
Among them, $\smash{I {\scriptstyle _\mathrm{GA}^{(i)}}}$ denotes either the GA or teaser of $\smash{d {\scriptstyle ^{(i)}}}$.

\nbf{Intra-GA Recommendation}
We define Intra-GA Recommendation as the task of evaluating the appropriateness of each figure $\smash{I {\scriptstyle _j^{(i)}}}$ as a GA within a paper $\smash{d {\scriptstyle ^{(i)}}}$ and recommending the most suitable candidates.  
The candidate set is defined as $\smash{\mathcal{I} {\scriptstyle _\mathrm{Intra}^{(i)}} = \{ I {\scriptstyle _j^{(i)}} \mid j \in \{ \mathrm{GA}, 1, 2, \dots, n {\scriptstyle ^{(i)}}\} \}}$,
where the GT is denoted as $\smash{I {\scriptstyle _\mathrm{GA}^{(i)}}}$.
If $\smash{I {\scriptstyle _\mathrm{GA}^{(i)}}}$ consists of multiple subfigures, its relevance score is determined by the maximum similarity among its subfigures.
In most cases, $\smash{I {\scriptstyle _\mathrm{GA}^{(i)}}}$ correspond to \textit{Figure~1}, meaning that models prioritizing it tend to achieve high scores.  
However, such models fail to effectively assess GA suitability beyond positional bias.  
Therefore, figures must be evaluated as a set, independent of their order of appearance, which serves as a constraint of this task.

\nbf{Inter-GA Recommendation}
We define Inter-GA Recommendation as the task of evaluating the relevance of GAs from other papers as design references for creating a GA for a given paper $\smash{d {\scriptstyle ^{(i)}}}$.
The candidate set is defined as $\smash{\mathcal{I} {\scriptstyle _\mathrm{Inter}^{(i)}} = \{ I {\scriptstyle _\mathrm{GA}^{(i')}} \mid i' \in \{1, 2, \dots, N\}, i' \neq i \}}$.
Since this task supports authors before creating a GA, $\smash{I{\scriptstyle_\mathrm{GA}^{(i)}}}$ must not be used as the query, which is a constraint of this task.
Unlike Intra-GA Recommendation, this task does not have an explicitly defined GT, as the relevance of a figure depends on subjective design preferences and contextual factors.  


\subsection{Confidence Adjusted top-1 GT Ratio (CAR)}
\label{sec:proposed:car}

Evaluating Intra-GA Recommendation is inherently challenging, as multiple figures within a paper may plausibly serve as GAs beyond the GT with explicit labels.
However, each instance in our settings has only binary labels (1 for the GT, 0 for others), despite the presence of semantically plausible alternatives.
This is because predefining graded relevance labels is neither scalable nor reliable.
In such settings, traditional metrics such as R@$k$, mean reciprocal rank (MRR), and even nDCG face critical limitations, where nDCG, despite its potential for fine-grained evaluation, degenerates into a discrete rank-based score.
As a result, they assign the same score to any prediction that ranks the GT at the same position, regardless of how the model ranks or scores the other candidates.
This makes it difficult to distinguish between a model that confidently ranks a clearly wrong candidate first and one that narrowly prefers a plausible alternative over the GT, or between a clearly correct and an uncertain one, where the GT is chosen by chance among similar candidates.
This leads to coarse-grained and potentially misleading performance signals, particularly in tasks with inherently ambiguous or instance-level soft correctness.

To address this limitation, we introduce the Confidence Adjusted top-1 GT Ratio@$k$ (CAR@$k$), a confidence-aware evaluation metric for ambiguous recommendation tasks with binary GT labels.
It penalizes overconfident errors, discounts uncertain successes, and rewards predictions that are both correct and confident.
Formally, CAR@$k$ is defined as:
\begin{equation}
\label{eq:CAR}
\textrm{CAR@}k = \frac{p_\mathrm{GT}}{p_\mathrm{top\text{-}1}} \cdot \mathcal{C}(P, k),
\end{equation}
where $\smash{P \in \mathbb{R}^k}$ is the vector of predicted relevance scores over the top-$k$ candidates, which are z-score normalized and converted into probabilities via the softmax function.
Here, $p_\mathrm{GT}$ and $p_\mathrm{top\text{-}1}$ denote the estimated probabilities of the GT and the top-1 candidate, respectively.

$\mathcal{C}(P, k)$ quantifies the model’s confidence, ranging from 0.5 (low confidence) to 1.0 (high confidence):
\begin{equation}
\label{eq:confidence}
\mathcal{C}(P, k) = 1 - \frac{1}{2} \max \left( 0, \frac{H(P) - h}{H_{\mathrm{max}}(P) - h} \right),
\end{equation}
where $H(P)$ is the entropy of the probability distribution $P$, $H_{\mathrm{max}}(P) = \log k$ is the maximum entropy, and $h = \log k / 2$ is a threshold that separates high and low certainty regimes.
If $H(P) \leq h$, then $\mathcal{C}(P, k) = 1.0$; otherwise, it decreases smoothly toward 0.5 as uncertainty increases.

It provides an interpretable, instance-level continuous score in the $[0, 1]$ that semantically distinguishes model behavior:
1) Clearly Correct ($\textrm{CAR} \sim 1.0$): the model strongly prefers the GT over all other candidates.
2) Uncertain Success / Understandable Error ($\textrm{CAR} \sim 0.5$):  the model either selects the GT with low confidence or ranks a plausible alternative slightly higher.
3) Clearly Incorrect ($\textrm{CAR} \sim 0.0$): the model strongly prefers a non-GT over the GT.
Furthermore, the proportion of instances where $\textrm{CAR} > 0.5$ offers a direct estimate of how often a model makes semantically justifiable predictions.
Note that CAR provides this perspective without relying on graded labels, and assigns instance-level continuous scores, whereas nDCG requires such costly label annotations, while R@$k$ assigns binary scores per instance.

\section{Experiments}
\label{sec:experiments}


\subsection{Experimental Setup}
\label{sec:experiments:experimental-setup}

We conducted experiments using a subset of SciGA-145k: 20,520 papers with GAs or teasers in computer science, the largest field in the dataset.
The dataset is split into training, validation, and test sets (8:1:1).
For implementation and environment details, see the Appendix.

\subsubsection{Intra-GA Recommendation}
\label{sec:experiments:experimental-setup:intra-GA-recommendation}
We compare four different approaches:
(i) an abstract-to-caption lexical matching-based method (Abs2Cap),
(ii) a GA/non-GA binary classification-based method (GA-binCl),
(iii) an abstract-to-figure retrieval-based method (Abs2Fig),
and (iv) an abstract-to-figure retrieval-based method that incorporates figure captions (Abs2Fig w/cap).
The backbone models and details are described in \cref{sec:experiments:benchmark-methods}.
All models are evaluated using R@$k$ ($k = 1, 2, 3$), MRR, nDCG@5, and CAR@5.
CAR@5 is reported as its mean and the proportion of queries exceeding 0.5.

\subsubsection{Inter-GA Recommendation}
\label{sec:experiments:experimental-setup:inter-GA-recommendation}
In this setting, each abstract from the test set serves as a query, while the candidates are the GAs from the training set.
We adopt the same methods as in Intra-GA Recommendation, excluding GA-binCl.  
As a baseline (BL), we adopt a random sampling approach, where $k$ GAs are randomly sampled from the training set per query.

Inter-GA Recommendation aims to retrieve GAs as design references.
Because \textit{usefulness} is subjective and lacks explicit GT, we follow the three-step protocol below:

\nbf{(1) Define interpretable quantitative axes}
Based on empirical observations and prior studies~\cite{jeyaraman2023ga}, we identify following complementary quantitative axes that characterize desirable reference GAs:
\textbf{(a)~Field Match}:
A desirable GA should belong to the same research field as the target paper, reflecting field-specific visual conventions and layout styles (\cref{fig:embed}).
We quantify this by a binary indicator of whether the query and candidate GAs share the same arXiv primary category.
\textbf{(b)~Semantic Coherence}: 
The reference GA should be semantically related to the research topics and core concepts of the query paper.  
We measure this as the cosine similarity between the Sentence-BERT~\cite{reimers2019sentence-bert} embeddings of their abstracts.
\textbf{(c)~Visual Coherence}:
The GA should be visually consistent with the author-created GA, effectively conveying the intended information structure and emphasis.  
We compute this as the CLIPScore~\cite{hessel2020clipscore} between their images.
\textbf{(d)~Aesthetic Quality}:
A good GA should maintain visual balance, readability, and overall clarity—qualities that make it easy to interpret at a glance.  
We estimate it using the LAION Aesthetic Predictor V2~\cite{schuhmann2022aesthetic}.
These axes are not intended as absolute evaluation criteria, but rather as interpretable axes that describe different properties contributing to effective GA recommendations.

\nbf{(2) Human Validation}
To examine how these quantitative axes relate to human judgments,  
we conducted a user study with 15 professional machine learning researchers who have experience in GA creation.
Each participant evaluated 60 query abstracts, each accompanied by six GA pairs retrieved by different models, resulting in a total of $15 \times 60 = 900$ pairwise comparisons.
For each pair, participants were asked to select the GA they preferred as design inspiration without being informed of the method used, and to indicate the key factors that influenced their decision.
The six pairs covered all possible combinations among four methods: Random Sampling, Abs2Cap (ROUGE-L), Abs2Fig (CLIP), and Abs2Fig w/cap (CLIP).
For each comparison, we computed the point-biserial correlation between the difference in each quantitative axis and participants’ binary choices.
This analysis identifies which quantitative aspect best aligns with human preference.

\nbf{(3) Evaluating under Silver Standard Criterion}
All models are evaluated using nDCG@$k$ ($k = 5, 10, 30$), where relevance labels between query and candidate GAs are determined by the quantitative axis that aligns most consistently with human preference.  
If none of the candidate axes exhibit a statistically meaningful correlation with human judgment, we report the results under all axes without privileging a particular one, treating Inter-GA Recommendation as an open, exploratory analysis rather than a definitive ranking task.


\subsection{Benchmark Methods}
\label{sec:experiments:benchmark-methods}

To benchmark different methods for two GA recommendation tasks, we construct models that rank figures based on relevance scores defined according to various criteria.
These models then recommend the top-$k$ candidates based on their computed rankings.
For methods utilizing figure captions, we preprocess captions by removing tags (e.g., \textit{Figure~1}) to eliminate positional bias, as outlined in \cref{sec:proposed:task-definition}.
Please refer to the Appendix for details of the models.

\nbf{(i) Abs2Cap} 
We quantify the relevance of each figure $\smash{I {\scriptstyle _j^{(i)}}}$ by measuring the textual similarity between its caption $\smash{C {\scriptstyle _j^{(i)}}}$ and the abstract $\smash{T {\scriptstyle ^{(i)}}}$, using metrics such as ROUGE-L~\cite{lin2004rouge-l}, METEOR~\cite{banerjee2005meteor}, CIDEr~\cite{vedantam2015cider}, BM25~\cite{robertson1994bm25}, and BERTScore~\cite{zhang2020bertscore}.
This approach follows prior studies~\cite{yang2019identifying, yamamoto2021visual}, and serves as a baseline for Intra-GA Recommendation.

\nbf{(ii) GA-binCl}
We formulate Intra-GA Recommendation as a set of binary classification problems to prevent \textit{Figure~1} from always being selected, as noted above. Each figure $\smash{I {\scriptstyle _j^{(i)}}}$ is independently assessed to estimate its probability of being a GA,
which serves as its relevance score.
Several models, including EfficientNetV2~\cite{tan2021efficientnetv2}, ViT~\cite{dosovitskiy2020vit}, CLIP image encoder, SwinTransformerV2~\cite{liu2022swin-transformer}, and ConvNeXtV2~\cite{woo2023convnext} are fine-tuned using cross-entropy loss to distinguish GA from non-GA figures.
Unlike other methods that leverage the query paper's contextual information, this method relies solely on individual visual features.
Thus, it applies only to Intra-GA Recommendation and is excluded from the Inter-GA Recommendation, which requires cross-paper comparisons.

\nbf{(iii) Abs2Fig} 
We employ a contrastive learning model consisting of a text encoder $f(\cdot)$ and an image encoder $g(\cdot)$.
These encoders project the abstract $\smash{T {\scriptstyle ^{(i)}}}$ and each figure $\smash{I {\scriptstyle _j^{(i)}}}$ into a shared embedding space.
The relevance score of $\smash{I {\scriptstyle _j^{(i)}}}$ is then computed as the cosine similarity between $\smash{f(T {\scriptstyle ^{(i)}})}$ and $\smash{g(I {\scriptstyle _j^{(i)}})}$, denoted as $\rho(f(T {\scriptstyle ^{(i)}}), g(I {\scriptstyle _j^{(i)}}))$.
Models such as CLIP, OpenCLIP~\cite{cherti2023open-clip}, Long-CLIP~\cite{zhang2024long-clip}, SigLIP2~\cite{tschannen2025siglip2}, BLIP-2~\cite{li2023blip-2}, and X$\smash{{}\scriptstyle ^2}$-VLM~\cite{zeng2023x2-vlm} are trained with a contrastive loss based on InfoNCE~\cite{oord2018info-nce}, which maximizes the similarity between a query embedding $\boldsymbol{z}^\mathrm{q}$ and a positive example $\boldsymbol{z}^\text{+}$ while minimizing similarities with a set of negative examples $\boldsymbol{z}_i^\text{-}$:
\begin{equation}
\label{eq:InfoNCE}
\mathcal{L}_\mathrm{C}(
    \boldsymbol{z}^\mathrm{q},
    \boldsymbol{z}^\text{+},
    \{ \boldsymbol{z}_i^\text{-} \}
)
= - \log \frac
{
    e^{\frac{\rho(\boldsymbol{z}^\mathrm{q}, \boldsymbol{z}^\text{+})}{\tau}}
}
{
    e^{\frac{\rho(\boldsymbol{z}^\mathrm{q}, \boldsymbol{z}^\text{+})}{\tau}}
    + \sum_{i} e^{\frac{\rho(\boldsymbol{z}^\mathrm{q}, \boldsymbol{z}^\text{-}_i)}{\tau}}
},
\end{equation}
where $\tau$ is a temperature parameter that controls the scaling of similarity scores.
In mini-batch training, a mini-batch $\mathcal{B} \subset \{1, 2, \dots, N\}$ is randomly sampled from the dataset.
For Intra-GA Recommendation, the model is optimized using the following loss function:
\begin{equation}
\label{eq:intraLoss}
\mathcal{L}_\mathrm{Intra} =
\frac{1}{|\mathcal{B}|}
\sum_{i \in \mathcal{B}}
    \mathcal{L}_\mathrm{C}
    (
        f(T^{(i)}),
        g(I_\mathrm{GA}^{(i)}),
        \{ g(I_\mathrm{j \neq \mathrm{GA}}^{(i)}) \}
    ).
\end{equation}
This strengthens associations between the abstract and GA while pushing apart non-GA figures.
Since the number of figures $\smash{n {\scriptstyle ^{(i)}}}$ varies across papers, we randomly sample $m$ figures during training, applying zero-padding when fewer than $m$ figures are available.
In Inter-GA Recommendation, the model is optimized using the following loss function:
\begin{align}
\label{eq:interLoss}
\mathcal{L}_\mathrm{Inter}
= & \frac{1}{2|\mathcal{B}|}
\sum_{i \in \mathcal{B}}
    \mathcal{L}_\mathrm{C}
    (
        f(T^{(i)}),
        g(I_\mathrm{GA}^{(i)}),
        \{ g(I_\mathrm{GA}^{(i' \neq i)}) \}
    ) \notag \\
+ & \frac{1}{2|\mathcal{B}|}
\sum_{i \in \mathcal{B}}
    \mathcal{L}_\mathrm{C}
    (
        g(I_\mathrm{GA}^{(i)}),
        f(T^{(i)}),
        \{ f(T^{(i' \neq i)}\}
    ),
\end{align}
which strengthens associations between abstracts and their GAs while pushing apart those from other papers.

\nbf{(iv) Abs2Fig w/cap} 
To further enhance the representation of each figure $\smash{I {\scriptstyle _j^{(i)}}}$, we integrate its caption embedding $\smash{f(C {\scriptstyle _j^{(i)}})}$ into the figure embedding $\smash{g(I {\scriptstyle _j^{(i)}})}$ via a Hadamard product.
The relevance score of $\smash{I {\scriptstyle _j^{(i)}}}$ is then computed as the cosine similarity $\smash{\rho(f(T {\scriptstyle ^{(i)}}), g(I {\scriptstyle _j^{(i)}}) \odot f(C {\scriptstyle _j^{(i)}}))}$.
This modified similarity measure is also used during training, replacing $\smash{\rho(f(T {\scriptstyle ^{(i)}}), g(I {\scriptstyle _j^{(i)}}))}$ in the loss functions $\smash{\mathcal{L} {\scriptstyle _\mathrm{Intra}}}$ and $\smash{\mathcal{L} {\scriptstyle _\mathrm{Inter}}}$.
\section{Results and Discussion}
\label{sec:results}


\subsection{Intra-GA Recommendation}
\label{sec:results:intra}

\begin{table*}[!t]
    \caption{
        Quantitative comparison of various approaches for the Intra-GA Recommendation.
        0.5 $\uparrow$ indicates the proportion of semantically justifiable predictions (CAR@5 $>$ 0.5).  
        (iii) Abs2Fig w/cap performed best overall, highlighting the benefit of incorporating abstract and caption context. 
        The best results for each metric are highlighted in \textbf{bold}. 
    }
    \label{tbl:intra_results}
    \centering
    \small
    \resizebox{0.63\textwidth}{!}{
        \begin{tabular}{llrrcccccc}
            \toprule
            
            \multirow{2}{*}{\textbf{Method}} &
            \multirow{2}{*}{\begin{tabular}[c]{@{}c@{}} \textbf{Backbone} \\ ($\smash{ \scriptstyle ^\dagger}$ Max Token Length) \end{tabular}} &
            \multirow{2}{*}{\textbf{R@1}} &
            \multirow{2}{*}{\textbf{R@2}} &
            \multirow{2}{*}{\textbf{R@3}} &
            \multirow{2}{*}{\textbf{MRR}} &
            \multirow{2}{*}{\textbf{nDCG@5}} &
            \multicolumn{2}{c}{\textbf{CAR@5}} \\
            \cmidrule(lr){8-9}
            & & & & & & & Mean & 0.5 $\uparrow$ \\
  
            \midrule
            
            \multirow{5}{*}{(i) Abs2Cap}
            & ROUGE-L~\cite{lin2004rouge-l}                                              & 0.394 & 0.625 & 0.759 & 0.601 & 0.664 & 0.429 & 0.448 \\
            & METEOR~\cite{banerjee2005meteor}                                           & 0.353 & 0.589 & 0.737 & 0.571 & 0.638 & 0.404 & 0.401 \\
            & CIDEr~\cite{vedantam2015cider}                                             & 0.277 & 0.489 & 0.653 & 0.500 & 0.571 & 0.374 & 0.089 \\
            & BM25~\cite{robertson1994bm25}                                              & 0.508 & 0.739 & 0.849 & 0.690 & 0.747 & 0.528 & 0.633 \\
            & BERTScore~\cite{zhang2020bertscore} ($\smash{ \scriptstyle ^\dagger}$ 512) & 0.485 & 0.707 & 0.819 & 0.668 & 0.726 & 0.505 & 0.545 \\
    
            \cmidrule(lr){1-1}
            \cmidrule(lr){2-2}
            \cmidrule(lr){3-3}
            \cmidrule(lr){4-4}
            \cmidrule(lr){5-5}
            \cmidrule(lr){6-6}
            \cmidrule(lr){7-7}
            \cmidrule(lr){8-8}
            \cmidrule(lr){9-9}
     
            \multirow{5}{*}{(ii) GA-binCl}
            & EfficientNetV2~\cite{tan2021efficientnetv2}      & 0.449 & 0.674 & 0.797 & 0.643 & 0.703 & 0.486 & 0.545 \\
            & ViT~\cite{dosovitskiy2020vit}                    & 0.346 & 0.606 & 0.762 & 0.574 & 0.647 & 0.420 & 0.430 \\
            & CLIP image encoder~\cite{radford2022clip}        & 0.493 & 0.708 & 0.826 & 0.675 & 0.734 & 0.518 & 0.602 \\
            & SwinTransformerV2~\cite{liu2022swin-transformer} & 0.494 & 0.712 & 0.823 & 0.675 & 0.730 & 0.516 & 0.584 \\
            & ConvNeXtV2~\cite{woo2023convnext}                & 0.483 & 0.703 & 0.816 & 0.667 & 0.725 & 0.511 & 0.577 \\
    
            \cmidrule(lr){1-1}
            \cmidrule(lr){2-2}
            \cmidrule(lr){3-3}
            \cmidrule(lr){4-4}
            \cmidrule(lr){5-5}
            \cmidrule(lr){6-6}
            \cmidrule(lr){7-7}
            \cmidrule(lr){8-8}
            \cmidrule(lr){9-9}
            
            \multirow{5}{*}{(iii) Abs2Fig}
            & CLIP~\cite{radford2022clip}                            ($\smash{ \scriptstyle ^\dagger}$  77) & 0.573 & 0.791 & 0.877 & 0.735 & 0.786 & 0.573 & 0.647 \\
            & BLIP-2~\cite{li2023blip-2}                             ($\smash{ \scriptstyle ^\dagger}$ 512) & 0.578 & 0.787 & 0.867 & 0.737 & 0.787 & 0.577 & 0.649 \\
            & X$\smash{{}\scriptstyle ^2}$-VLM~\cite{zeng2023x2-vlm} ($\smash{ \scriptstyle ^\dagger}$  40) & 0.489 & 0.711 & 0.825 & 0.672 & 0.730 & 0.514 & 0.571 \\
            & OpenCLIP~\cite{cherti2023open-clip}                    ($\smash{ \scriptstyle ^\dagger}$  77) & 0.566 & 0.780 & 0.870 & 0.730 & 0.781 & 0.567 & 0.641 \\
            & SigLIP2~\cite{tschannen2025siglip2}                    ($\smash{ \scriptstyle ^\dagger}$  64) & 0.558 & 0.772 & 0.869 & 0.724 & 0.776 & 0.588 & 0.636 \\
            & Long-CLIP~\cite{zhang2024long-clip}                    ($\smash{ \scriptstyle ^\dagger}$ 248) & 0.575 & 0.783 & 0.877 & 0.735 & 0.785 & 0.573 & 0.646 \\

            \cmidrule(lr){1-1}
            \cmidrule(lr){2-2}
            \cmidrule(lr){3-3}
            \cmidrule(lr){4-4}
            \cmidrule(lr){5-5}
            \cmidrule(lr){6-6}
            \cmidrule(lr){7-7}
            \cmidrule(lr){8-8}
            \cmidrule(lr){9-9}
            
            \multirow{5}{*}{(iv) Abs2Fig w/cap}
            & CLIP~\cite{radford2022clip}                            ($\smash{ \scriptstyle ^\dagger}$  77) &        0.628 &        0.822 &        0.902 &        0.771 &        0.816 &        0.610 &        0.689 \\
            & BLIP-2~\cite{li2023blip-2}                             ($\smash{ \scriptstyle ^\dagger}$ 512) &        0.557 &        0.767 &        0.863 &        0.721 &        0.773 &        0.557 &        0.626 \\
            & X$\smash{{}\scriptstyle ^2}$-VLM~\cite{zeng2023x2-vlm} ($\smash{ \scriptstyle ^\dagger}$  40) &        0.538 &        0.757 &        0.857 &        0.709 &        0.763 &        0.546 &        0.618 \\
            & OpenCLIP~\cite{cherti2023open-clip}                    ($\smash{ \scriptstyle ^\dagger}$  77) &        0.621 &        0.817 &        0.905 &        0.767 &        0.813 &        0.603 &        0.681 \\
            & SigLIP2~\cite{tschannen2025siglip2}                    ($\smash{ \scriptstyle ^\dagger}$  64) &        0.527 &        0.748 &        0.853 &        0.702 &        0.758 &        0.564 &        0.593 \\
            & Long-CLIP~\cite{zhang2024long-clip}                    ($\smash{ \scriptstyle ^\dagger}$ 248) &\textbf{0.637}&\textbf{0.826}&\textbf{0.914}&\textbf{0.778}&\textbf{0.824}&\textbf{0.615}&\textbf{0.691}\\
    
            \bottomrule
        \end{tabular}
    }

    \vspace{-2mm}

\end{table*}

\begin{figure}[!t]
    \centering
    \includegraphics[width=\linewidth]{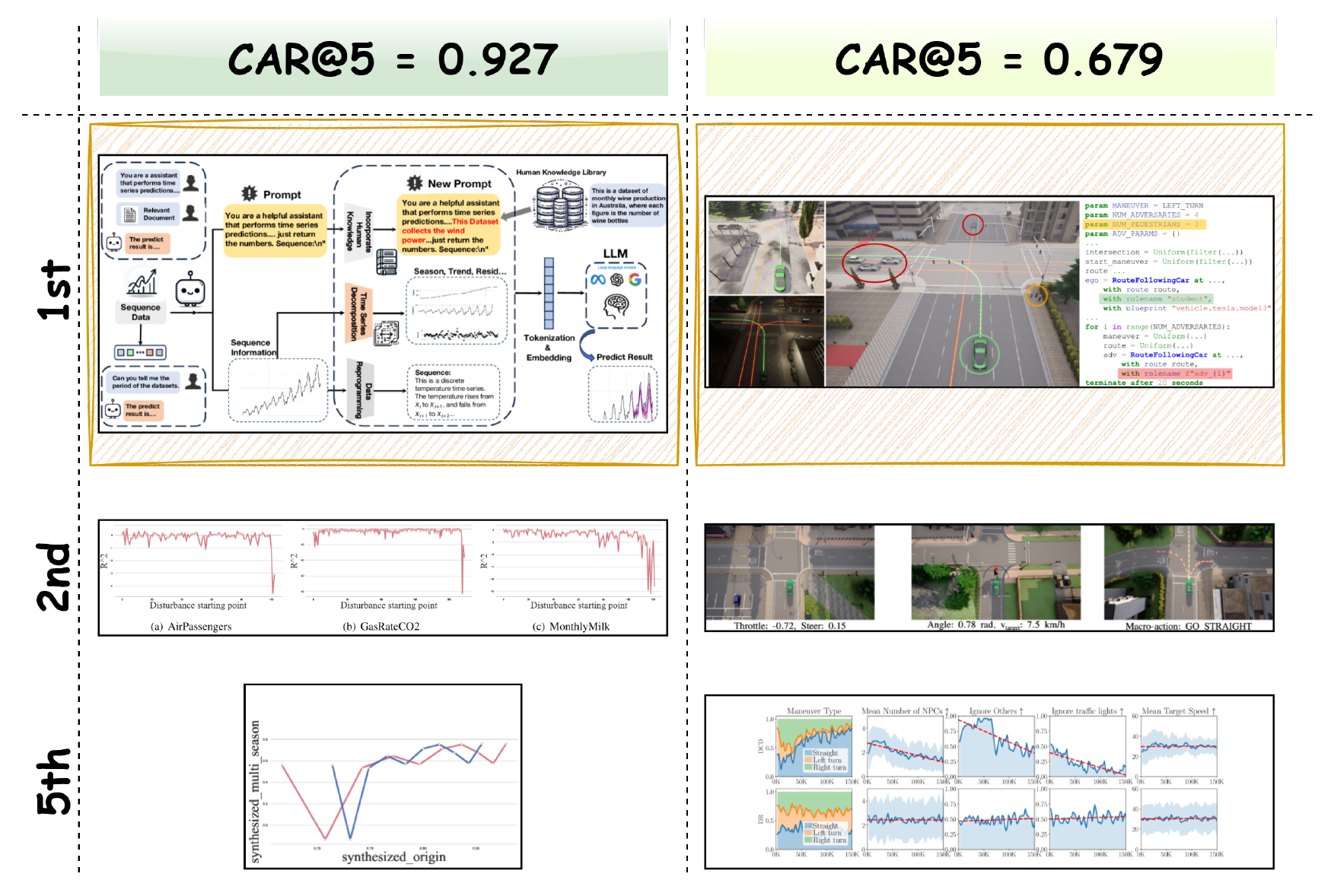}
    \caption{
        Representative Intra-GA Recommendation results obtained by the best-performing model. \protect \footnotemark[4]
        Both rank the paper's GA (yellow box) first, but CAR captures the model’s confidence, dropping when the GA does not clearly stand out from non-GA figures.
    }
    \label{fig:intra_results}
    \vspace{-2mm}
\end{figure}

\nbf{Performance}
\cref{tbl:intra_results} summarizes the quantitative results for Intra-GA Recommendation.  
Methods (iii) Abs2Fig and (iv) Abs2Fig w/cap consistently outperformed (i) Abs2Cap and (ii) GA-binCl.  
Notably, method (iv) further improved upon method (iii), suggesting that captions provide additional context useful for distinguishing fine-grained differences among visually similar candidates.
In particular, within method (iv), Long-CLIP demonstrated the best retrieval performance (R@1: 0.637).
This improvement is likely due to Long-CLIP's extended text encoder input length (248 tokens), allowing it to leverage comprehensive abstracts and longer captions to establish more detailed and accurate alignments.
In contrast, BLIP-2, despite its strong baseline performance in (iii) Abs2Fig, showed a drop in performance in (iv) Abs2Fig w/cap.
This suggests that incorporating captions using BLIP-2’s Q-Former may undermine, rather than improve, the alignment between figures, captions, and abstracts.
\cref{fig:intra_results} shows a representative example where the model correctly ranks the GA among in-paper figures, reflecting its preference for clear and overview-style visuals.

\nbf{Error Analysis with CAR}
Unlike conventional metrics that only reflect the GT's rank, CAR additionally captures the confidence of model predictions.
Interestingly, X$^2$-VLM in (iii) Abs2Fig shows a comparable or even higher nDCG (0.730) than SwinTransformerV2 (0.730) and ConvNeXtV2 (0.725) in (ii) GA-binCl, as shown in \cref{tbl:intra_results}.
However, its proportion of predictions with CAR@5 exceeding 0.5 was lower (0.571), indicating a lower robustness compared to GA-binCl models.
This discrepancy highlights the importance of evaluating CAR in addition to ranking-based metrics.

\footnotetext[4]{
\footnotesize
\begin{tabular}[t]{@{}ll@{}}
arXiv ID: 
\href{https://arxiv.org/abs/2402.10835}{2402.10835},
\href{https://arxiv.org/abs/2403.17805}{2403.17805}
\end{tabular}
}

Beyond aggregate scores, CAR supports instance-level analysis.  
We found that CAR around 0.5 often occurred when GTs appeared alongside other plausible candidates, such as overviews of model architecture or figures highlighting the impact of key contributions, making it difficult for the model to decide. 
CAR near 0.0 often reflected confident failures, where the model favored a common and visually appealing figure like a pipeline or setup diagram, while the GT was a less salient figure used to supplement background context.
These cases reveal that current models, while capturing visual and content-level cues, often fail to reflect the author's intent.
Addressing this gap is future work.


\subsection{Inter-GA Recommendation}
\label{sec:results:inter}

\begin{figure}[!t]
    \centering
    \includegraphics[width=\linewidth]{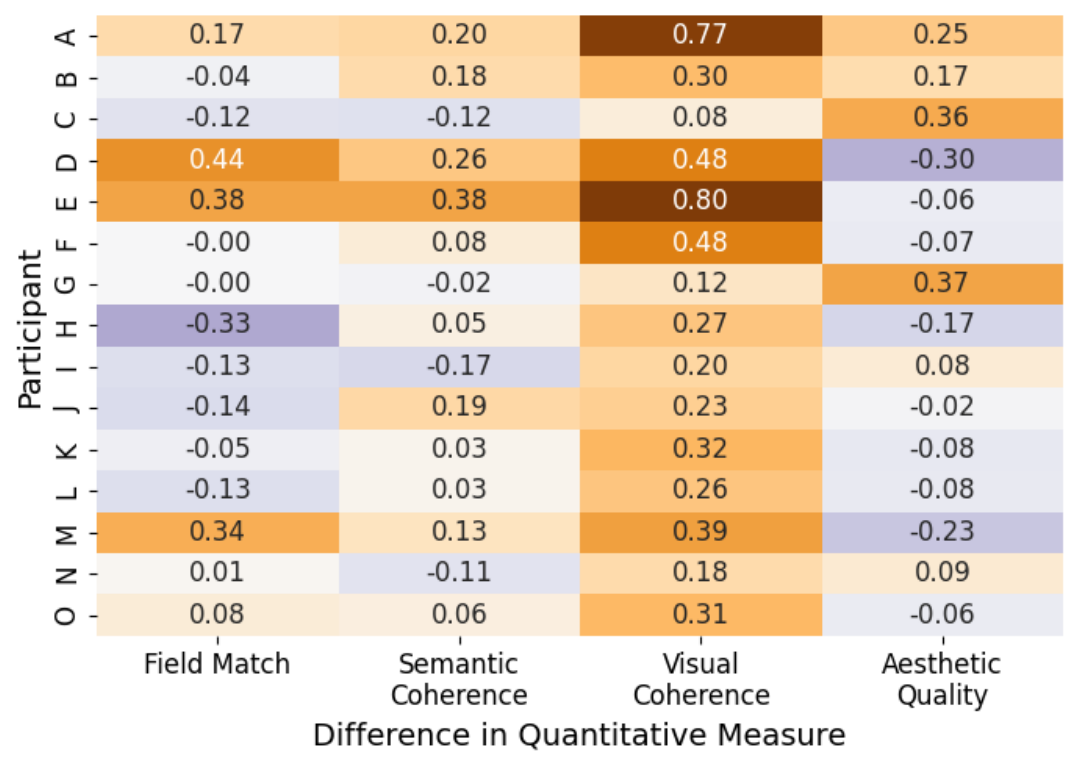}
    \vspace{-6mm}
    \caption{
        Per-participant point-biserial correlation coefficients between human preference and differences in each quantitative axis in Inter-GA Recommendation.
    }
    \label{fig:user_study_correlation}
\end{figure}

\nbf{Human Validation}
\cref{fig:user_study_correlation} shows the per-participant correlation matrix between human preference and the difference in each quantitative axis.
Across all participants, Visual Coherence, which was quantified using the CLIPScore between the author-created and recommended GAs, consistently exhibited positive correlation with human preference.
Overall, the correlation coefficient was $0.421$, and the $p$-value was $1.06\times10^{-14}$, indicating a significant association, with several users showing strong alignment.
Other axes revealed more diverse tendencies:
Correlations with Field Match varied across participants.
Some showed positive alignment, suggesting a preference for same-domain references, while others exhibited weak or even negative correlation, indicating openness to cross-disciplinary inspirations.
Some participants preferred references with higher Aesthetic Quality.
Overall, these results highlight that while human preference is inherently diverse, Visual Coherence captures the most universal tendency and serves as a key determinant of what users regard as a good reference GA.
We also observed a mean agreement score of $0.673 \pm 0.095$, indicating reasonable consistency despite the subjectivity of GA interpretation.

\begin{figure}[!t]
    \centering
    \includegraphics[width=\linewidth]{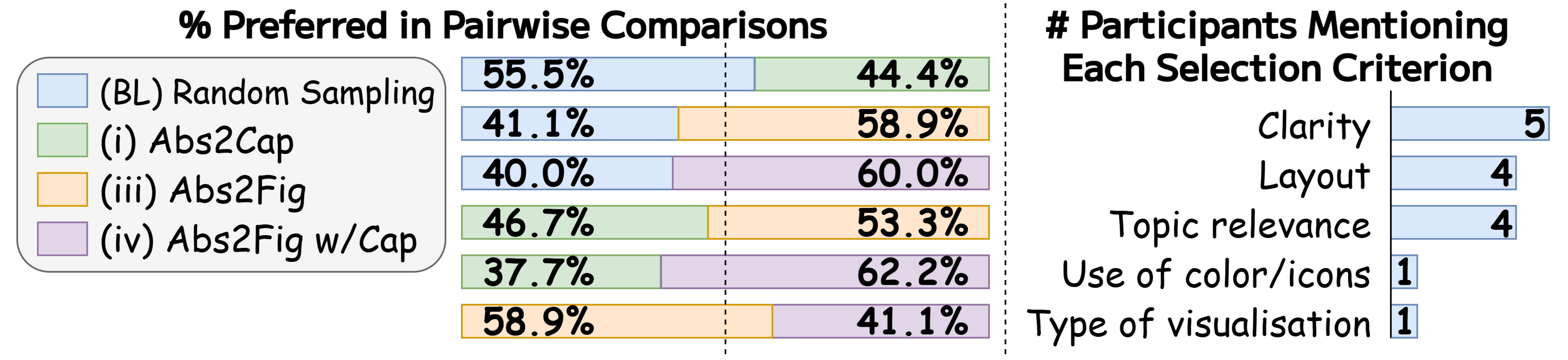}
    \caption{
        Users’ pairwise preferences results.
        Left: User preferences between method pairs.
        Right: Factors that influenced their decision, with the number of participants mentioning each factor.
    }
    \label{fig:user_study_results}
    \vspace{-2mm}
\end{figure}

\begin{figure}[!t]
    \centering
    \includegraphics[width=\linewidth]{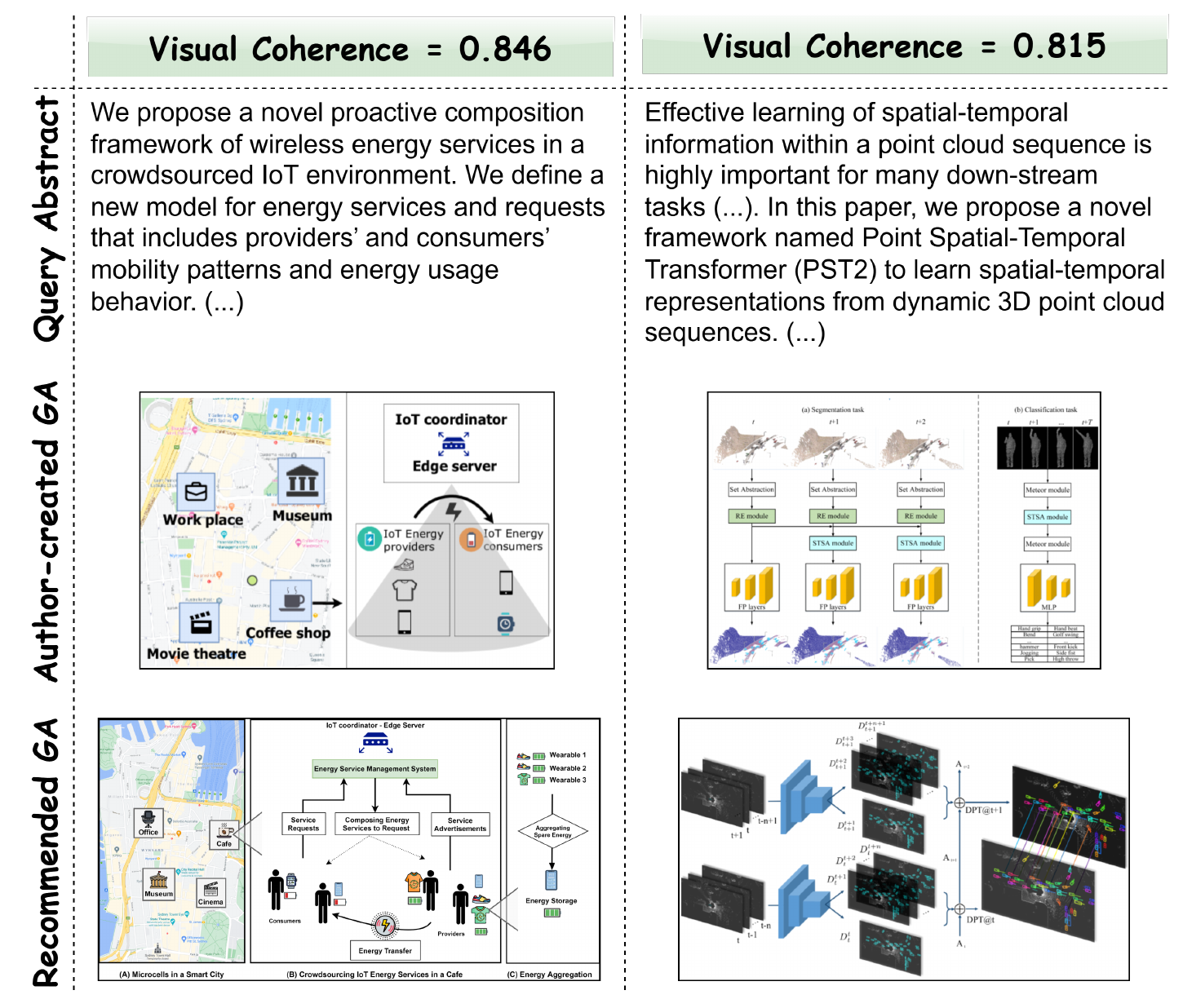}
    \vspace{-2mm}
    \caption{
        Representative Inter-GA Recommendation results obtained by the best-performing model. \protect \footnotemark[5]
        The model presents cross-paper GAs that, despite being retrieved solely from the abstract, exhibit strong Visual Coherence with the author-created GA.
    }
    \label{fig:inter_results}
    \vspace{-2mm}
\end{figure}

\nbf{Silver-Standard Evaluation using Visual Coherence}
Based on the above human validation, we adopt the CLIPScore between the author-created and recommended GAs as pseudo-relevance labels.
These continuous relevance labels allow nDCG to serve as a strong and reliable ranking metric.
\cref{tbl:inter_results_pseudo} summarizes the quantitative results under this criterion.
CLIP in (iii) Abs2Fig achieved the highest ranking performance (nDCG@5: 0.855), which is consistent with the model-preference trends observed in \cref{fig:user_study_results}.
Notably, these results also mirror the visual factors that participants most frequently mentioned as influencing their choices—such as \textit{Clarity} and \textit{Layout}.
Representative recommendation results are shown in \cref{fig:inter_results}, where Abs2Fig recommends visually coherent GAs that align with human perception.

\begin{table}[!t]
    \caption{
        Quantitative comparison of various approaches for the Inter-GA Recommendation using pseudo-relevance labels based on Visual Coherence (CLIPScore between the author-created and recommended GAs).
        Abs2Fig performed best overall.
        The best results for each metric are highlighted in \textbf{bold}. 
    }
    \label{tbl:inter_results_pseudo}
    \centering
    \small
    \resizebox{\linewidth}{!}{
        \begin{tabular}{llccc}
            \toprule
            \textbf{Method} &
            \textbf{Backbone} &
            \textbf{nDCG@5} &
            \textbf{nDCG@10} & 
            \textbf{nDCG@30} \\

            \midrule
            
            (BL) Random Sampling & -- & 0.770 & 0.794 & 0.862 \\
            
            \cmidrule(lr){1-1}
            \cmidrule(lr){2-2}
            \cmidrule(lr){3-3}
            \cmidrule(lr){4-4}
            \cmidrule(lr){5-5}

            \multirow{5}{*}{(i) Abs2Cap}
            & ROUGE-L~\cite{lin2004rouge-l}       & 0.808 & 0.830 & 0.887 \\
            & METEOR~\cite{banerjee2005meteor}    & 0.812 & 0.831 & 0.889 \\
            & CIDEr~\cite{vedantam2015cider}      & 0.820 & 0.839 & 0.892 \\
            & BM25~\cite{robertson1994bm25}       & 0.817 & 0.836 & 0.889 \\
            & BERTScore~\cite{zhang2020bertscore} & 0.807 & 0.828 & 0.887 \\
            
            \cmidrule(lr){1-1}
            \cmidrule(lr){2-2}
            \cmidrule(lr){3-3}
            \cmidrule(lr){4-4}
            \cmidrule(lr){5-5}
            
            \multirow{6}{*}{(iii) Abs2Fig} 
            & CLIP~\cite{radford2022clip}                            &\textbf{0.855}&\textbf{0.870}&\textbf{0.913} \\
            & BLIP-2~\cite{li2023blip-2}                             & 0.831        & 0.849        & 0.899 \\
            & X$\smash{{}\scriptstyle ^2}$-VLM~\cite{zeng2023x2-vlm} & 0.634        & 0.640        & 0.780 \\
            & OpenCLIP~\cite{cherti2023open-clip}                    & 0.841        & 0.858        & 0.906 \\
            & SigLIP 2~\cite{tschannen2025siglip2}                   & 0.811        & 0.831        & 0.888 \\
            & Long-CLIP~\cite{zhang2024long-clip}                    & 0.853        & 0.868        & 0.911 \\

            \cmidrule(lr){1-1}
            \cmidrule(lr){2-2}
            \cmidrule(lr){3-3}
            \cmidrule(lr){4-4}
            \cmidrule(lr){5-5}
            
            \multirow{6}{*}{(iv) Abs2Fig w/cap} 
            & CLIP~\cite{radford2022clip}                            & 0.822 & 0.840 & 0.893 \\
            & BLIP-2~\cite{li2023blip-2}                             & 0.815 & 0.835 & 0.890 \\
            & X$\smash{{}\scriptstyle ^2}$-VLM~\cite{zeng2023x2-vlm} & 0.786 & 0.808 & 0.871 \\
            & OpenCLIP~\cite{cherti2023open-clip}                    & 0.824 & 0.843 & 0.894 \\
            & SigLIP 2~\cite{tschannen2025siglip2}                   & 0.647 & 0.682 & 0.801 \\
            & Long-CLIP~\cite{zhang2024long-clip}                    & 0.819 & 0.838 & 0.891 \\

            \bottomrule
        \end{tabular}
    }
\end{table}

\section{Conclusion}
\label{sec:conclusion}

We introduced SciGA-145k, the first large-scale dataset explicitly designed to support GA design.  
We also defined two tasks, Intra-GA and Inter-GA Recommendation, along with a new recommendation metric, CAR.
Benchmark results demonstrated the effectiveness of CAR and contrastive learning with caption integration for Intra-GA Recommendation, while the user study highlighted that Visual Coherence best aligns with human judgment in Inter-GA Recommendation.
In future work, we aim to incorporate video-format GAs to better convey complex temporal processes and multi-dimensional findings beyond the limitations of static visuals.

\footnotetext[5]{
\footnotesize
\begin{tabular}[t]{@{}ll@{}}

arXiv ID (Query): &
\href{https://arxiv.org/abs/2107.12519}{2107.12519},
\href{https://arxiv.org/abs/2110.09783}{2110.09783} \\
arXiv ID (Recommended): &
\href{https://arxiv.org/abs/2308.09886}{2308.09886},
\href{https://arxiv.org/abs/2012.12395}{2012.12395}

\end{tabular}
}

\nbf{Limitation}
Our Inter-GA Recommendation benchmarks use only visual and semantic cues, without  incorporating strategies for novelty, serendipity, or researchers' latent preferences or intent.
Future enhancements could incorporate measures of personalization into frameworks, leveraging online metrics like user feedback or engagement tracking.

{
    \small
    \bibliographystyle{ieeenat_fullname}
    \bibliography{_references}
}

\clearpage
\appendix

\begin{figure*}[t]
\centering
\begin{minipage}{0.95\textwidth}
\begin{lstlisting}[
    language=json,
    caption=Example data in SciGA-145k.\protect \myfootnote,
    label=lst:json-sample,
    breaklines=true,
    captionpos=b,
    frame=single,
    showlines=true,
]
{
    "ID": "2401.13641",
    "title": "How Good is ChatGPT at Face Biometrics? ...",
    "authors": ["...", "..."],
    "published": "...",
    "subjects": {
        "arXiv": ["...", "..."], "ACM": ["...", "..."], "MSC": ["...", "..."]
    },
    "comment": "...",
    "journal_ref": "IEEE Access, February 2024",
    "conference": "...",
    "DOI": ["https://doi.org/10.48550/arXiv.2401.13641", "..."],
    "abstract": "Large Language Models (LLMs) such as GPT developed ..."
    "graphical_abstract": {
        "ID": "2401.13641_GA",
        "type": "Reused",
        "path": ["..."],
        "components": ["2401.13641_F1"],
        "caption": "...",
    },
    "teaser": ["2401.13641_F1"],
    "sections": {
        "ID": "2401.13641_S1",
        "title": "<TAG> 1 </TAG> Introduction",
        "body": "...",
        "subsections": {...},
        "figures": ["2401.13641_F1"],
    }, {...}
    "figures": {
        "ID": "2401.13641_F1",
        "caption": "<TAG> Fig. 1 </TAG> Graphical representation of ..."
        "path": ["..."],
        "subfigures": {...}
    }, {...}
}
\end{lstlisting}
\end{minipage}
\end{figure*}

\FloatBarrier

\section{Dataset Structure}
\label{app:dataset_structure}

Below, we describe the structure of SciGA-145k, focusing on how metadata, textual content, and visual content are represented.

\nbf{Metadata and Textual Data}  
SciGA-145k includes metadata and textual content extracted from scientific papers.  
Each entry is formatted in JSON and contains standard metadata fields (e.g., title, authors, abstract, research fields, DOI) as well as structured content such as section hierarchy and figure composition, including subfigures and captions.  
This rich structural representation supports fine-grained figure-level retrieval and content analysis.  
A partial example of the JSON structure is shown in \cref{lst:json-sample}.

\footnotetext[6]{
\begin{tabular}[t]{@{}ll@{}}
arXiv ID: 
\href{https://arxiv.org/abs/2401.13641}{2401.13641}
\end{tabular}
}

\nbf{Visual Data}
SciGA-145k includes figure data extracted from scientific papers. 
These visual assets are provided in PNG format for static figures and MP4 format for video-based GAs.
While most in-paper figures were successfully extracted from the HTML-rendered versions of arXiv papers, approximately 50k instances ($\sim$~4--5\%) exhibited rendering failures, either appearing as blank placeholders or visibly corrupted images.  
These failures were typically caused by complex \TeX{} figure commands (e.g., \texttt{\textbackslash includegraphics} with \texttt{page}, \texttt{trim}, or \texttt{clip}) that reference partial PDF content.  
To ensure dataset completeness, such figures were manually extracted from the original \TeX{} source files.
In contrast, GAs were manually collected from open-access journal versions identified via arXiv metadata (e.g., related DOI, journal reference, or author comments).  
However, only a small subset of papers include GAs, as most journals do not require them or restrict their reuse due to license limitations.  
Moreover, available GAs are often limited to specific journals and disciplines, making them unsuitable for constructing large-scale, balanced benchmarks, which is a limitation also observed in prior studies that typically rely on fewer than 100 samples.
To enable broader domain coverage and consistent empirical evaluation, we adopt in-paper teaser figures as scalable proxies for GAs, automatically labeled based on their position and layout.
Specifically, we heuristically identify teasers as figures that are cited as \textit{Figure~1} in the \textit{Introduction} and appear: 1) on the first page in full-column width or in the upper-right position, or 2) on the second page in full-column width.

\begin{figure*}[!t]
    \centering
    \includegraphics[width=0.95\textwidth]{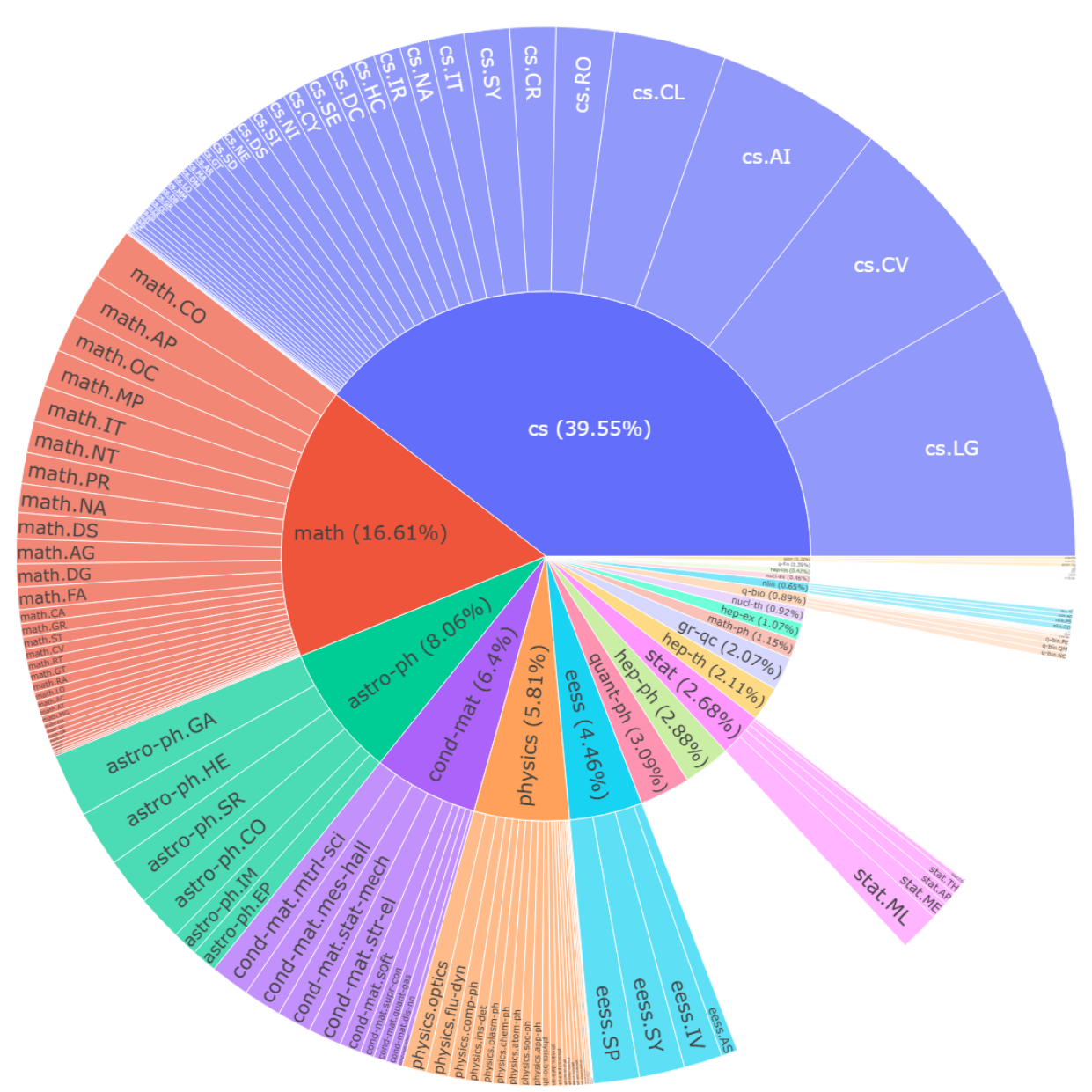}
    \caption{Distribution of research fields of papers included in SciGA-145k.}
    \label{fig:research_fields_chart}
\end{figure*}

\begin{figure*}[t]
    \centering
    \begin{minipage}{0.48\textwidth}
        \centering
        \includegraphics[width=\textwidth]{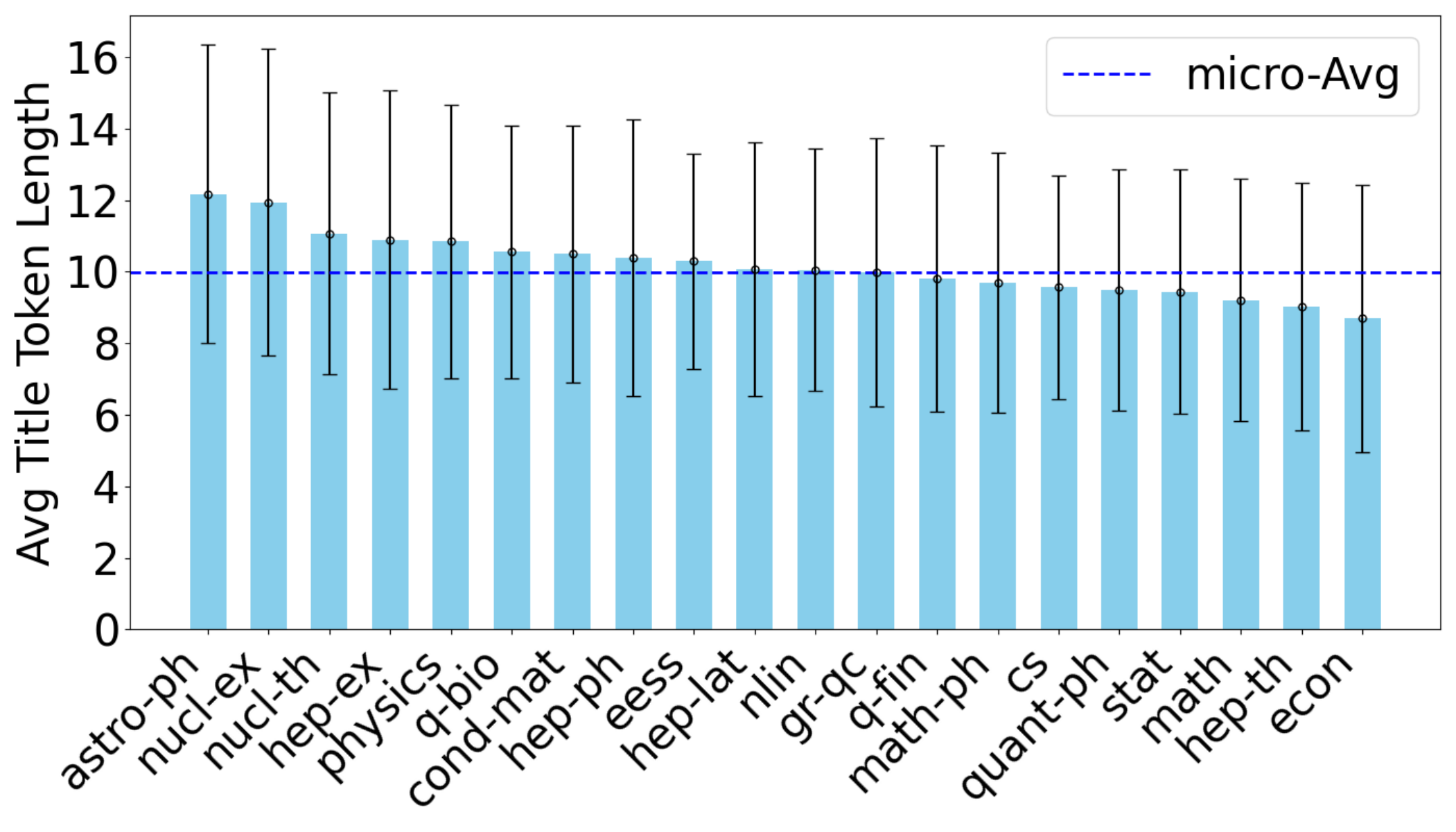}
        \subcaption{Average token length of title}
        \label{fig:statistics:a}
    \end{minipage}
    \begin{minipage}{0.48\textwidth}
        \centering
        \includegraphics[width=\textwidth]{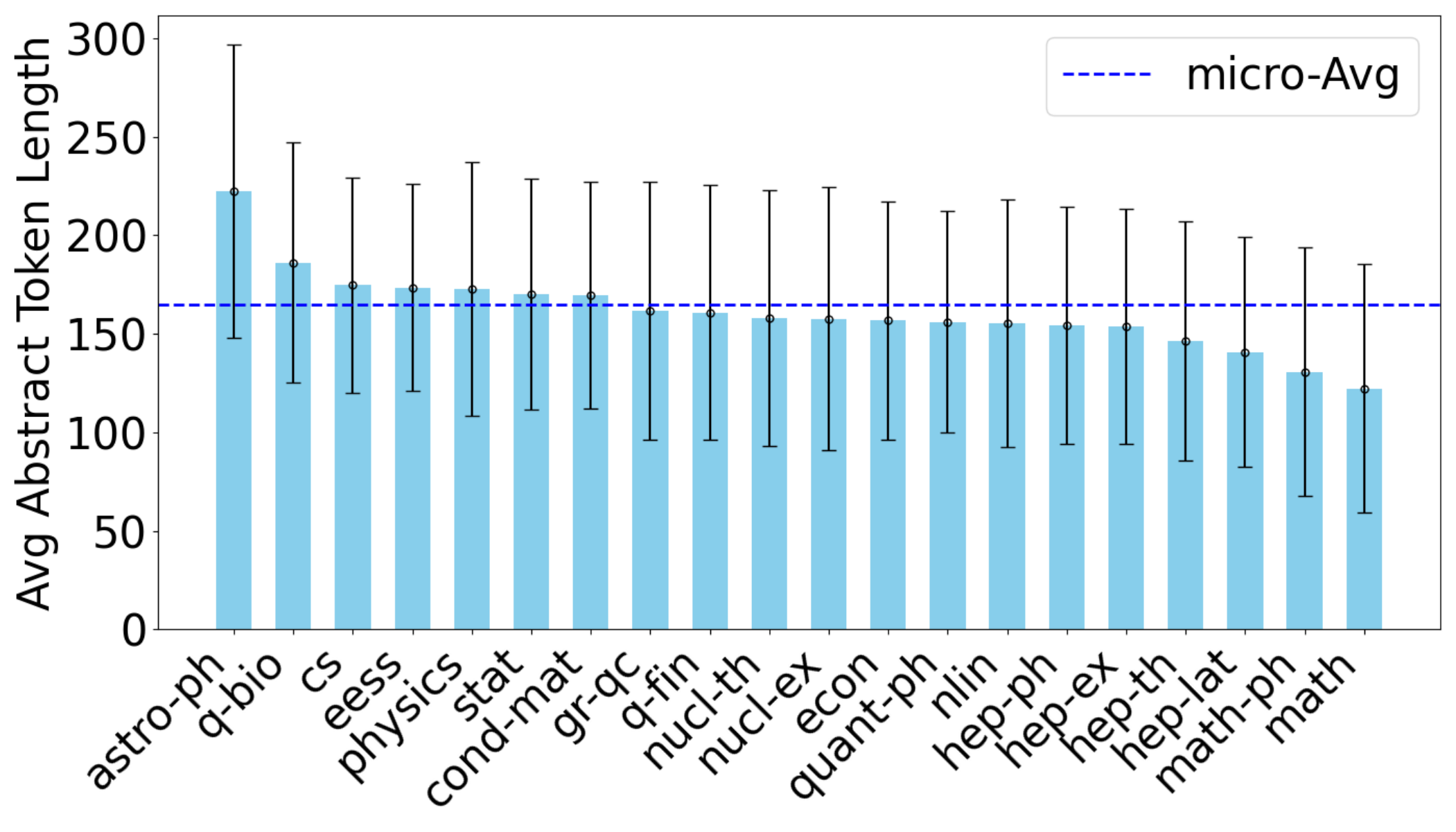}
        \subcaption{Average token length of abstract}
        \label{fig:statistics:b}
    \end{minipage}
    \begin{minipage}{0.48\textwidth}
        \centering
        \includegraphics[width=\textwidth]{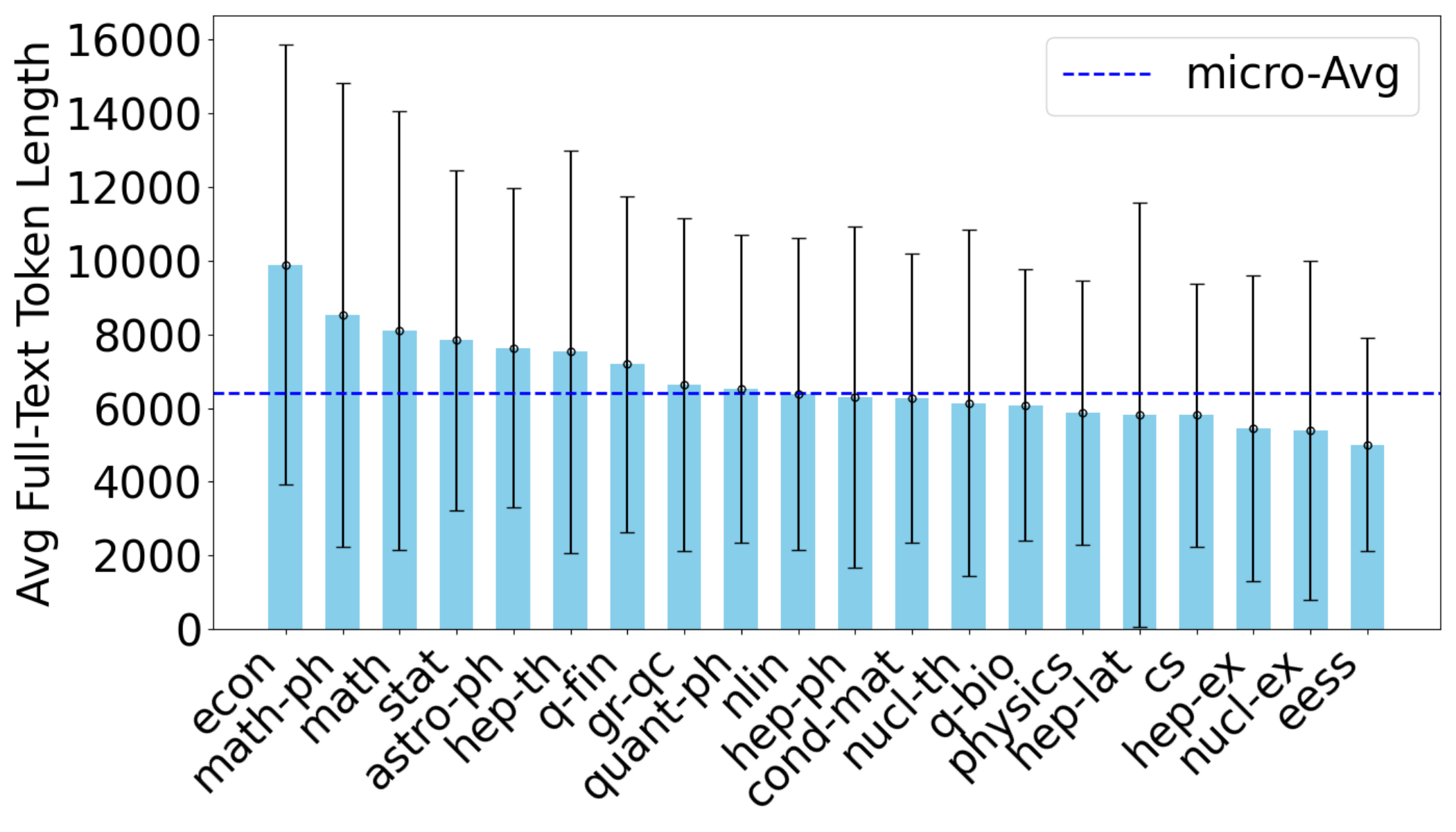}
        \subcaption{Average token length of full-text}
        \label{fig:statistics:c}
    \end{minipage}
    \begin{minipage}{0.48\textwidth}
        \centering
        \includegraphics[width=\textwidth]{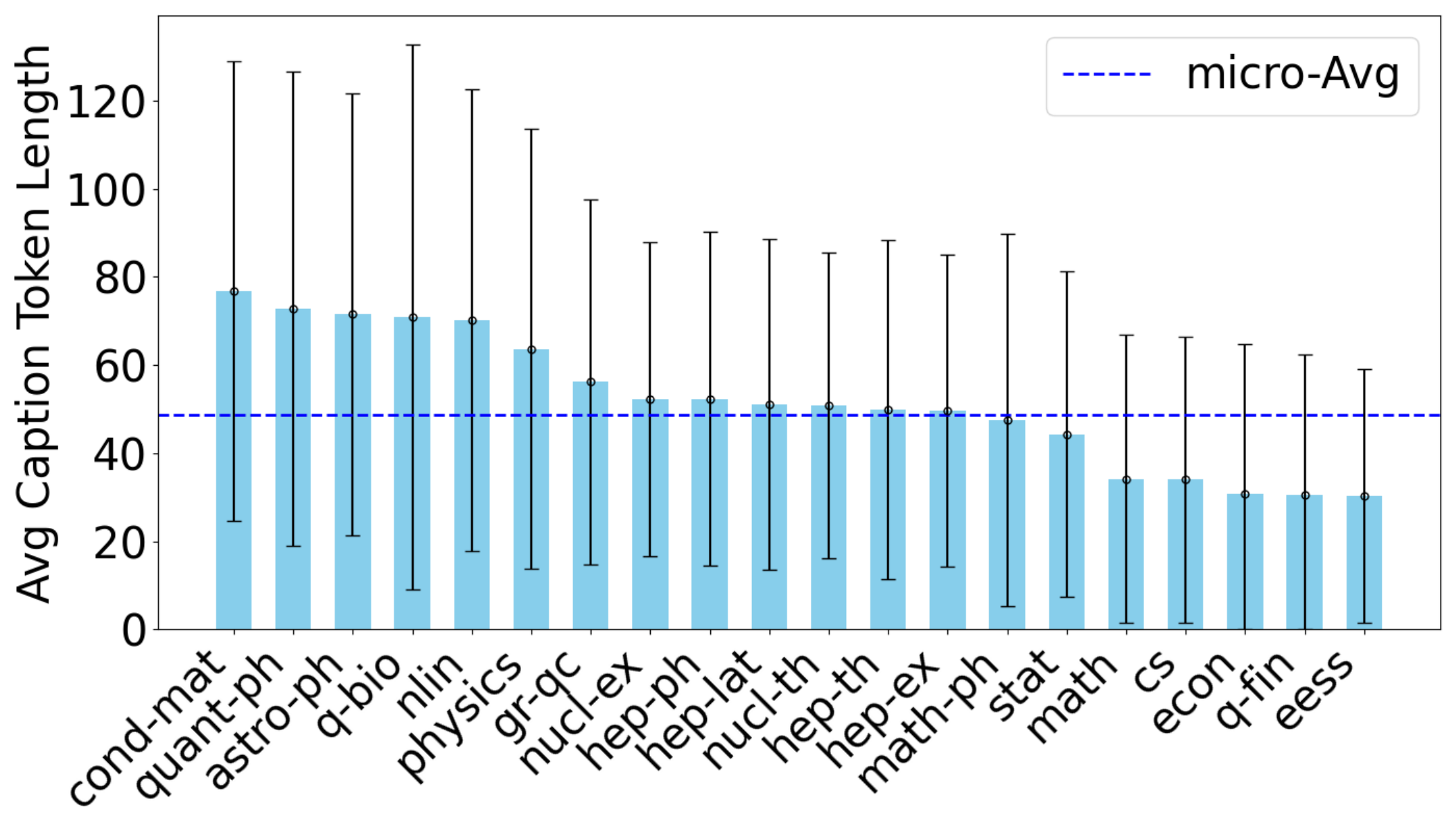}
        \subcaption{Average token length of caption}
        \label{fig:statistics:d}
    \end{minipage}   
    \begin{minipage}{0.48\textwidth}
        \centering
        \includegraphics[width=\textwidth]{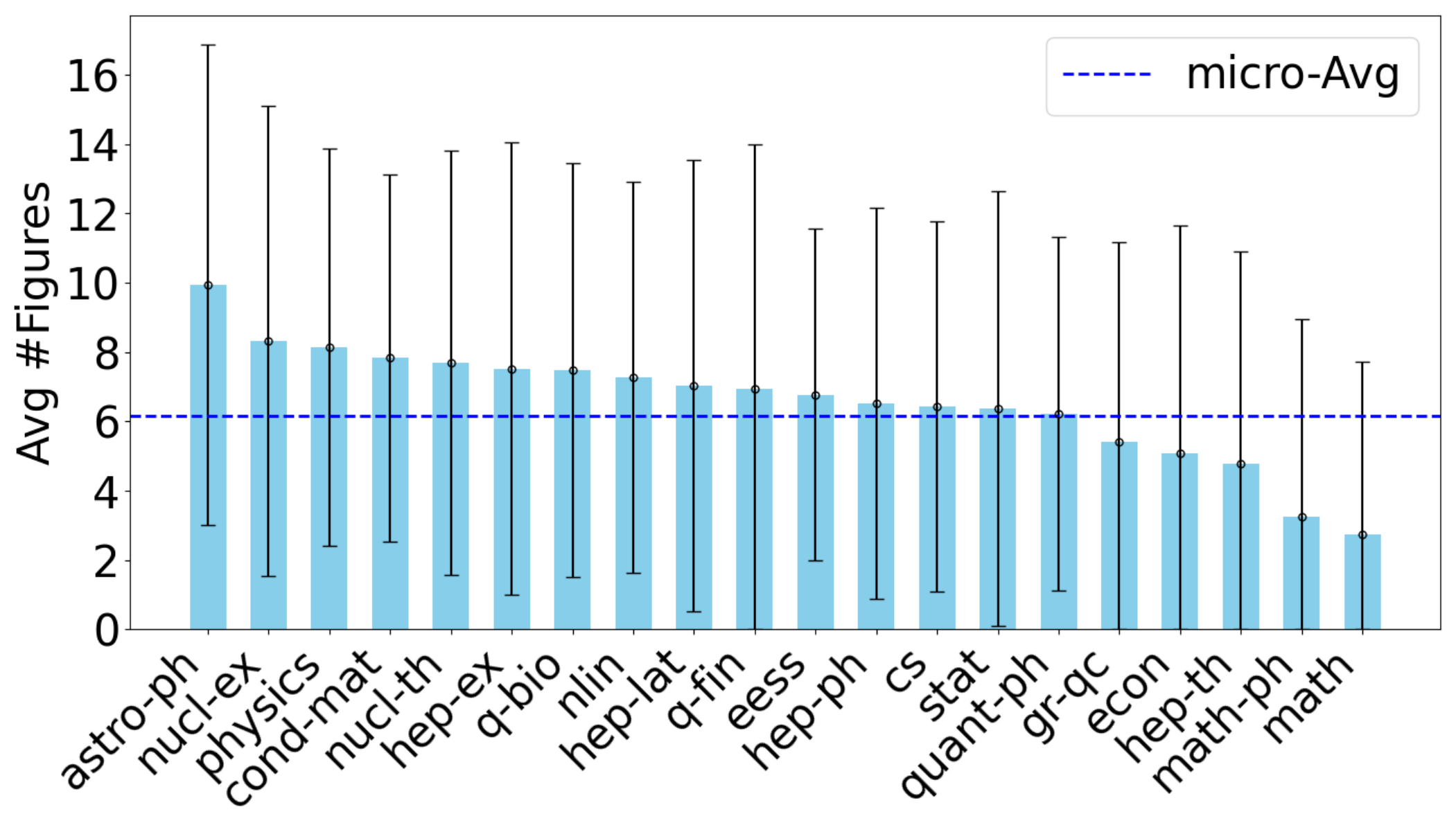}
        \subcaption{Number of figures per paper}
        \label{fig:statistics:e}
    \end{minipage}
    \begin{minipage}{0.48\textwidth}
        \centering
        \includegraphics[width=\textwidth]{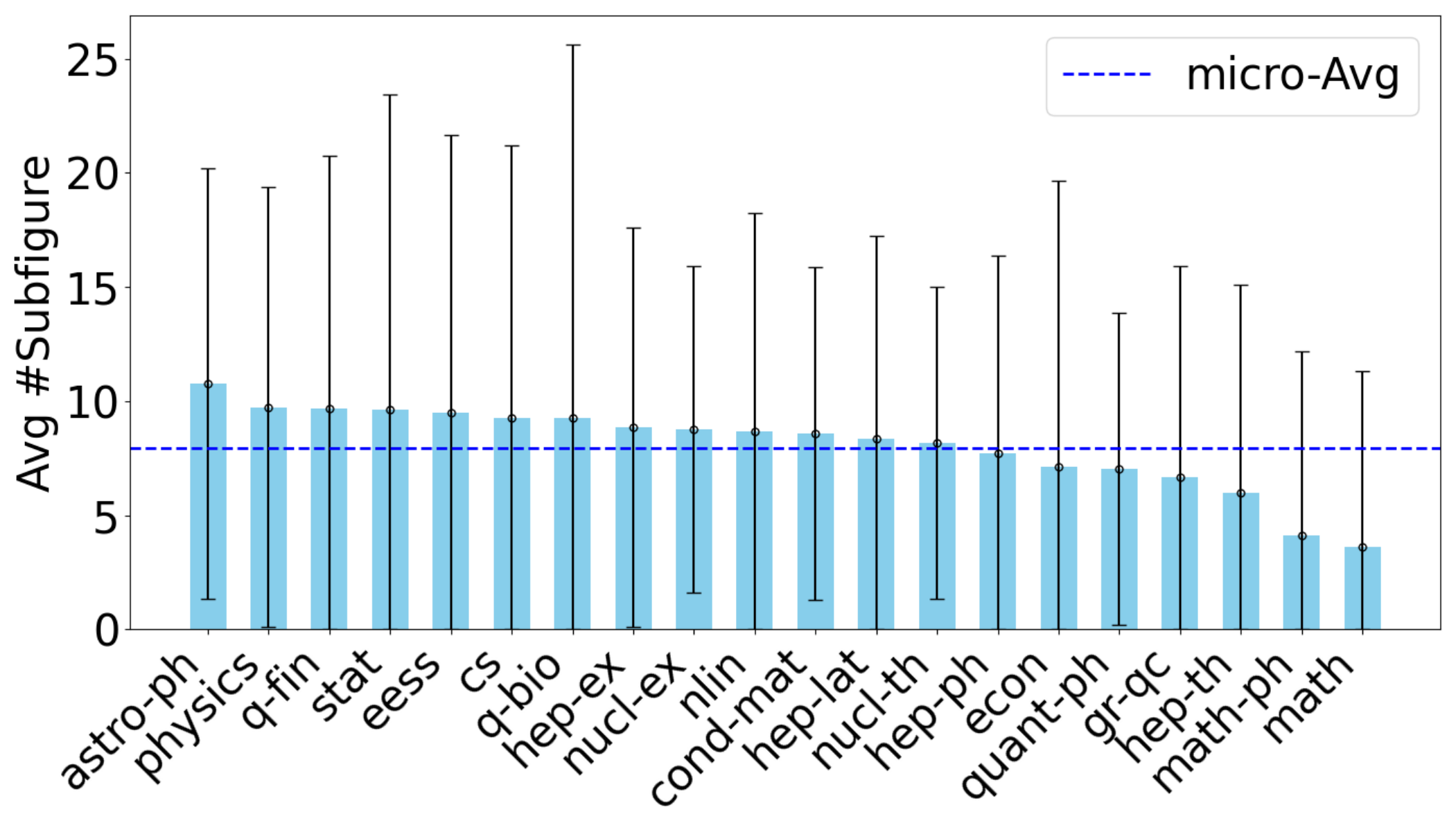}
        \subcaption{Number of subfigures per paper}
        \label{fig:statistics:f}
    \end{minipage}
    \caption{
        Statistical overview of SciGA-145k across top-level arXiv categories.  
        (a) Average token length of titles,
        (b) average token length of abstracts,
        (c) average token length of full-texts,
        (d) average token length of captions,
        (e) number of figures per paper,
        and (f) number of subfigures per paper.  
        Each graph presents the mean and standard deviation for each top-level arXiv category, alongside the overall micro-average.  
        These statistics highlight category-specific variations and overall distribution trends in the dataset.
    }
    \label{fig:statistics}
\end{figure*}

\section{Content Statistics}
\label{app:content_statistics}

This section presents summary statistics of SciGA-145k, covering research field distribution, content structure, and GA creation trends. These analyses complement the main paper and highlight the dataset’s diversity.

\nbf{Research Field Coverage}
Research fields in SciGA-145k are primarily classified using arXiv's hierarchical category system.
In addition, some papers include author-supplied labels from the ACM Computing Classification System~(ACM-CCS)~\footnotemark[7] and the Mathematics Subject Classification 2022~(MSC2022),\footnotemark[8] extracted from arXiv metadata or full-text content.
These entries were often unstructured or inconsistent, so we curated and normalized them using a combination of rule-based preprocessing and manual refinement (e.g., resolving casing and separator inconsistencies, removing label prefixes, assigning primary/secondary roles based on position or formatting, and identifying the taxonomy).

\footnotetext[7]{
\href{https://dl.acm.org/ccs}{https://dl.acm.org/ccs}
}

\footnotetext[8]{
\href{https://doi.org/10.4171/news/115/2}{https://doi.org/10.4171/news/115/2}
}

\cref{fig:research_fields_chart} shows the distribution of primary arXiv categories, with computer science (\texttt{cs}, 39.6\%) and mathematics (\texttt{math}, 16.6\%) as the largest categories, followed by astrophysics (\texttt{astro-ph}, 8.1\%) and condensed matter physics (\texttt{cond-mat}, 6.4\%).
This distribution demonstrates the dataset's broad domain coverage and suitability for cross-disciplinary analysis.

\nbf{Textual and Visual Statistics}
\cref{fig:statistics} summarizes token counts for titles, abstracts, full texts, captions, and the number of figures per paper across research fields.
Titles show little variation across domains, while abstracts tend to be longer in \texttt{astro-ph}, reflecting the field’s more descriptive writing style.
Full texts are notably lengthy in economics (\texttt{econ}), whereas \texttt{cs} and electrical engineering and systems science (\texttt{eess}) papers are generally shorter, following concise conference-oriented formats compared to journal-style publications.
Each paper includes on average $6.16 \pm 5.86$ figures ($7.92 \pm 10.45$ including subfigures), with astrophysics papers often exceeding 10 figures, compared to about 4 for mathematics, highlighting disciplinary differences in visual representation.
Captions also vary widely, with experimental sciences frequently featuring more detailed and informative figure descriptions

\begin{table}[!t]
    \caption{
        Distributional distances between GAs, teasers, and regular in-paper figures in CLIP space.
        GA--Teaser distances are extremely small under both CMMD and FD, indicating that the two figure types share nearly identical embedding distributions.
    }
    \label{tbl:GA_teaser_distance}
    \centering
    \small
    \begin{tabular}{lcc}
        \toprule
        
        & CMMD $\downarrow$ & FD $\downarrow$ \\

        \midrule
        
        GA--Teaser                               & 0.078 & 0.486 \\
        GA--\textit{Regular In-paper Figure}     & 0.284 & 3.888 \\
        Teaser--\textit{Regular In-paper Figure} & 0.128 & 1.974 \\

        \bottomrule
    \end{tabular}
\end{table}

\nbf{Examining GA Origins and Their Relation to Teasers}
Although GAs and teasers originate from different sources (journal-submitted vs.\ in-paper), they serve the same communicative role and are functionally indistinguishable.
To substantiate this claim, we examine whether any cultural or distributional distinctions between GAs and teasers exist.
First, we analyze how authors create their GAs by categorizing each GA into the following three types:
1) \textit{Original}: newly created GAs without reusing any in-paper figures;
2) \textit{Reuse}: GAs directly copied from figures in the paper without modifications; and
3) \textit{Modified}: GAs created by combining or modifying in-paper figures.
Annotations were manually performed by one author using deterministic criteria based on layout similarity, content overlap, and evidence of cropping or compositional edits, and all labels were reviewed with two co-authors. 
Among the 309 GAs, 20.9\% were categorized as \textit{Original}, 64.5\% as \textit{Reused}, and 14.5\% as \textit{Modified}, consistent with prior reports on GA creation patterns~\cite{yuanyuan2023ga}. 
These statistics indicate that authors themselves do not meaningfully distinguish between GAs and in-paper teasers.
Second, we analyze whether this cultural pattern is reflected in their embedding distributions.
\cref{tbl:GA_teaser_distance} summarizes the CMMD~\cite{jayasumana2024cmmd} and Fréchet distances (FD)~\cite{unterthiner2018fid} between GAs, teasers, and regular in-paper figures in CLIP space.
These metrics quantify the distributional discrepancy among figure types, and both clearly show that the GA--Teaser distance is extremely small, indicating that their embedding distributions are nearly identical.
When distances are measured within each research field, the GA--teaser distance becomes even smaller (average CMMD: 0.072), whereas GA--teaser distances across different fields are substantially larger (0.428).
These results suggest that the communicative role of a figure varies along two factors:
1) whether it is a summary-style figure (irrespective of being GA or teaser) or an ordinary figure; and 
2) which research field it belongs to.
Overall, our observations indicate that teasers and GAs can be regarded as fundamentally equivalent summary-style figures.

\section{Threshold Sensitivity of CAR}
\label{app:confidence_threshold}

\begin{figure*}[!t]
    \centering
    \begin{minipage}{0.32\linewidth}
        \centering
        \includegraphics[width=\textwidth]{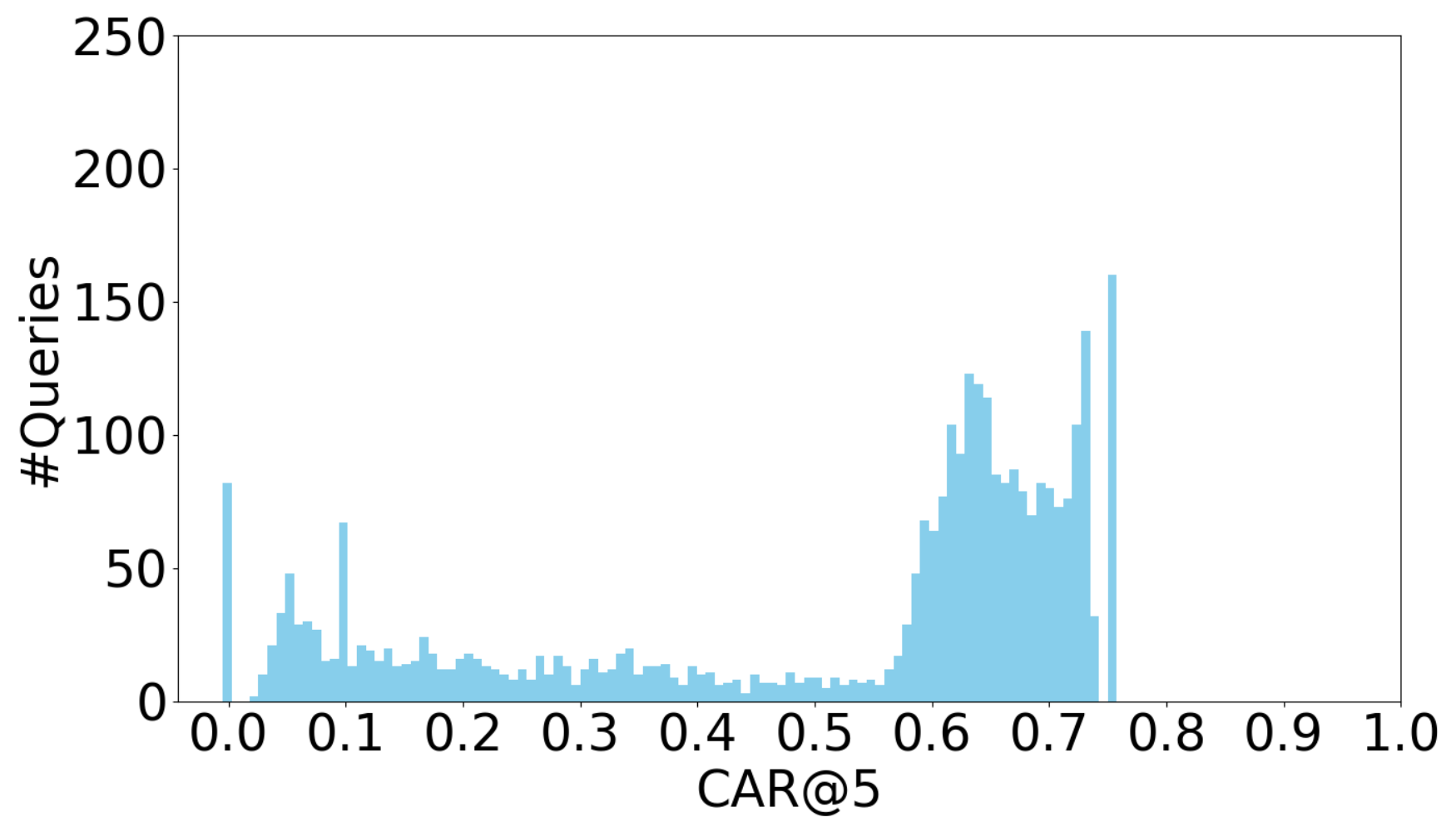}
        \subcaption{$\alpha = 0.1$}
        \label{fig:CAR_sensitivity_dist:a}
    \end{minipage}
    \begin{minipage}{0.32\linewidth}
        \centering
        \includegraphics[width=\textwidth]{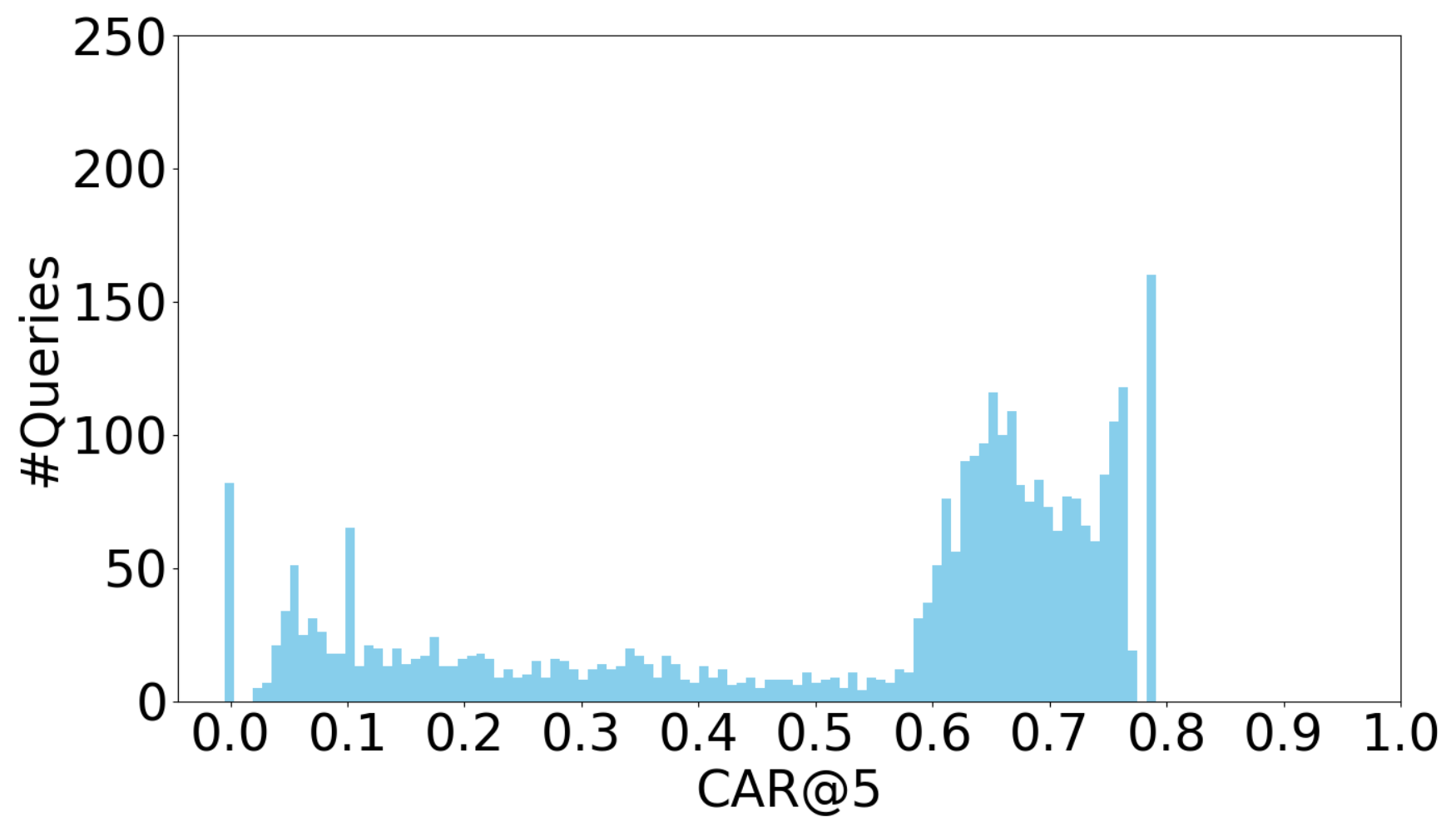}
        \subcaption{$\alpha = 0.2$}
        \label{fig:CAR_sensitivity_dist:b}
    \end{minipage}   
    \begin{minipage}{0.32\linewidth}
        \centering
        \includegraphics[width=\textwidth]{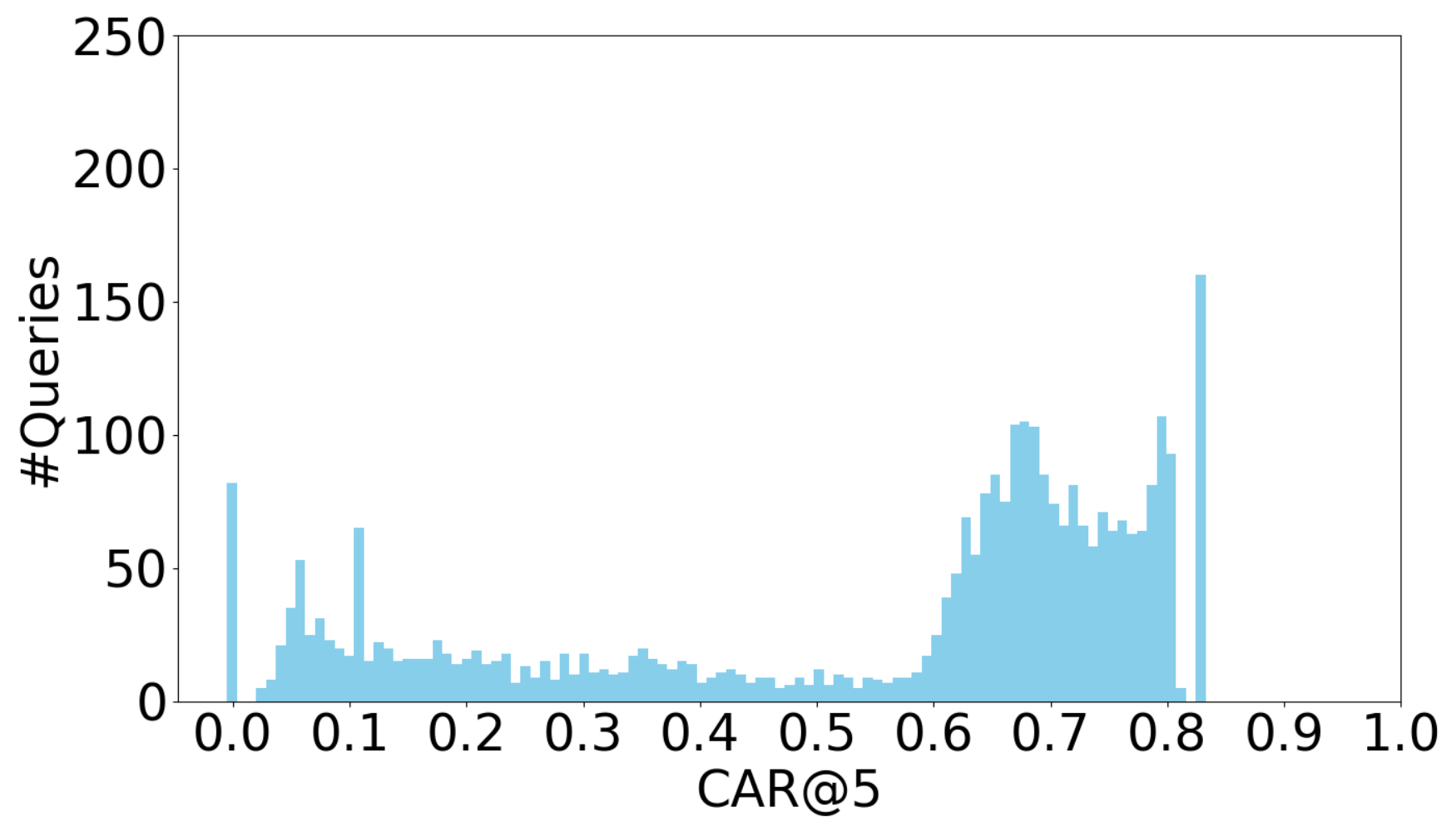}
        \subcaption{$\alpha = 0.3$}
        \label{fig:CAR_sensitivity_dist:c}
    \end{minipage}   
    \begin{minipage}{0.32\linewidth}
        \centering
        \includegraphics[width=\textwidth]{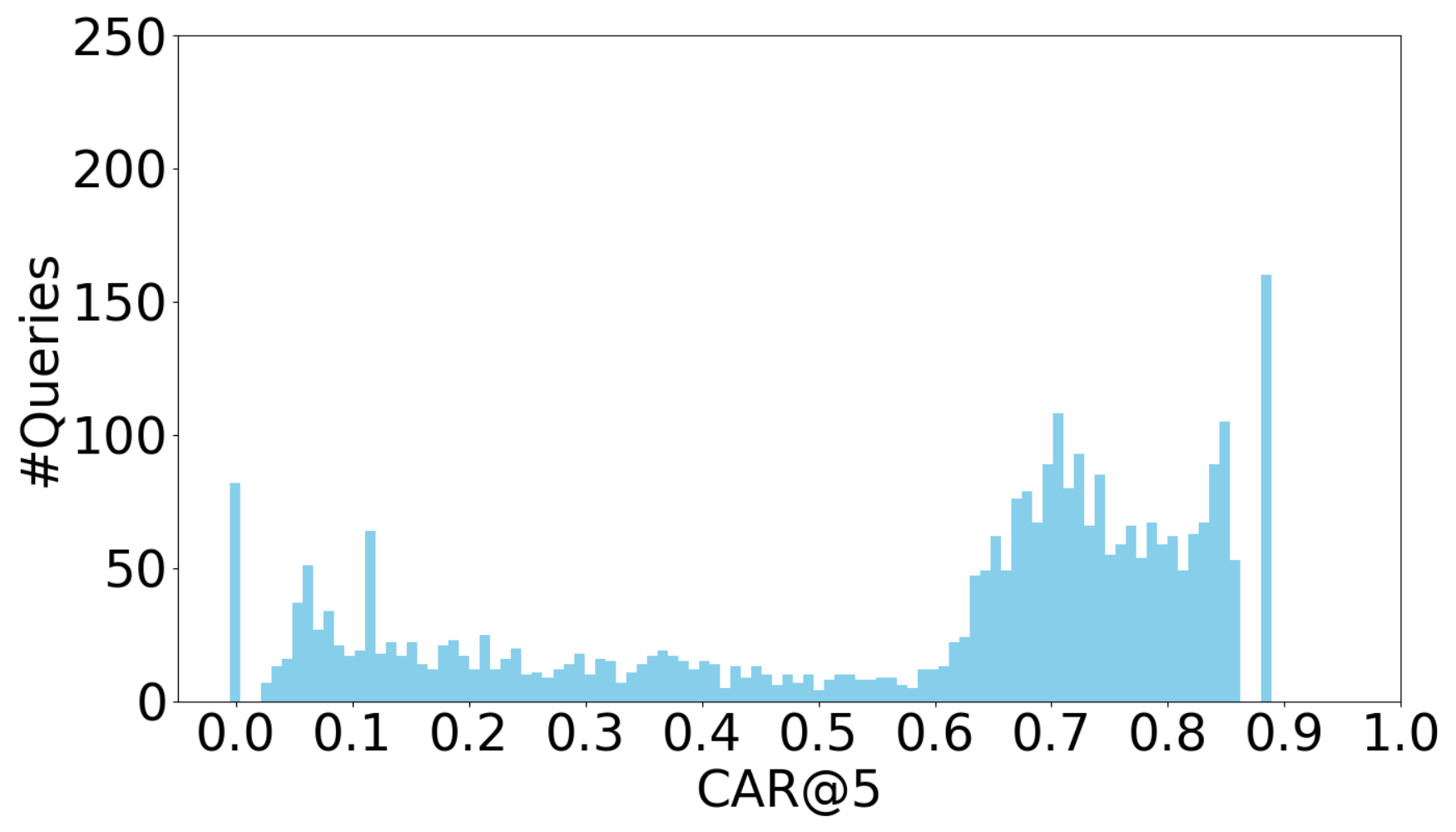}
        \subcaption{$\alpha = 0.4$}
        \label{fig:CAR_sensitivity_dist:d}
    \end{minipage}
    \begin{minipage}{0.32\linewidth}
        \centering
        \includegraphics[width=\textwidth]{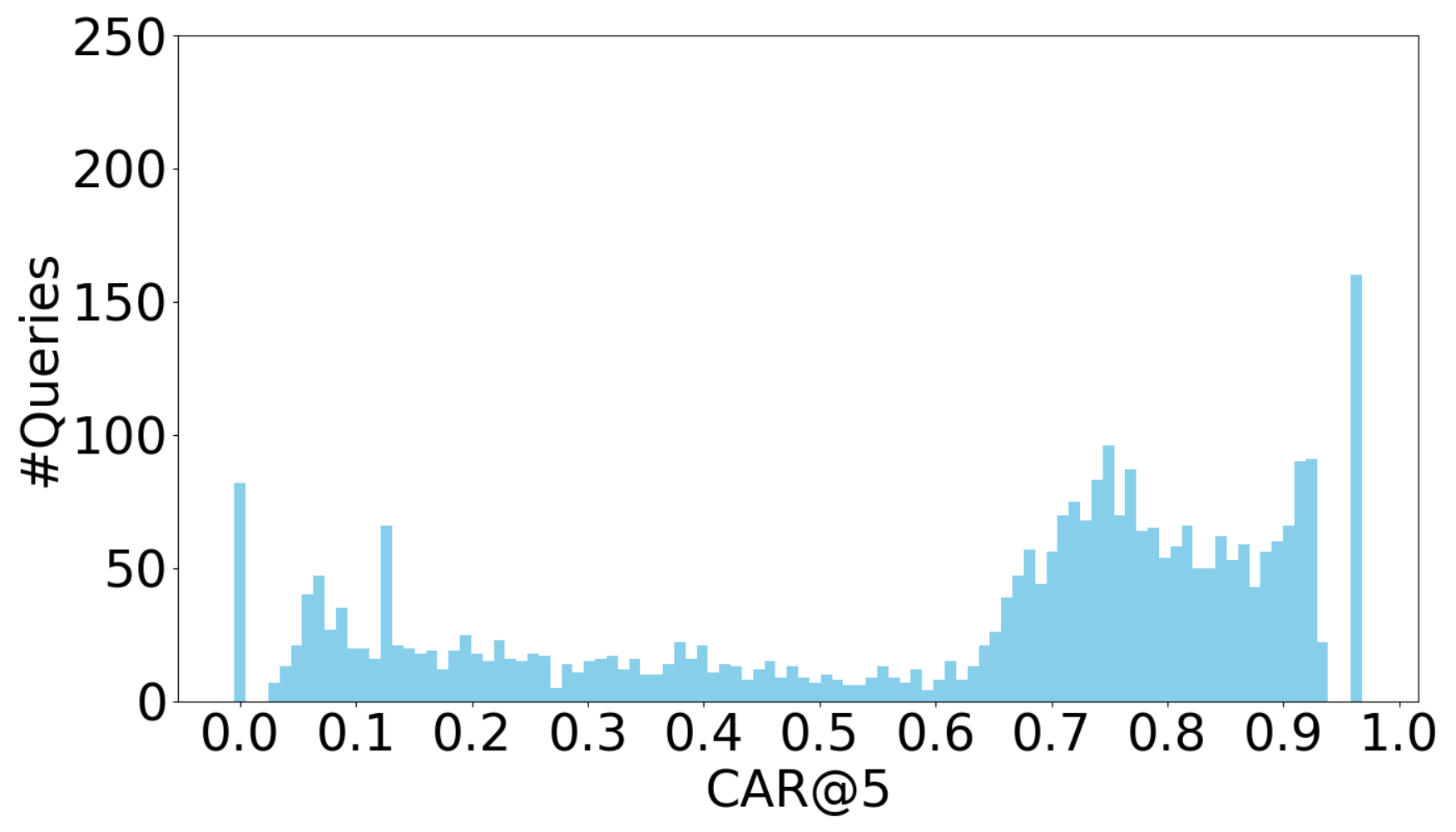}
        \subcaption{$\alpha = 0.5$}
        \label{fig:CAR_sensitivity_dist:e}
    \end{minipage}   
    \begin{minipage}{0.32\linewidth}
        \centering
        \includegraphics[width=\textwidth]{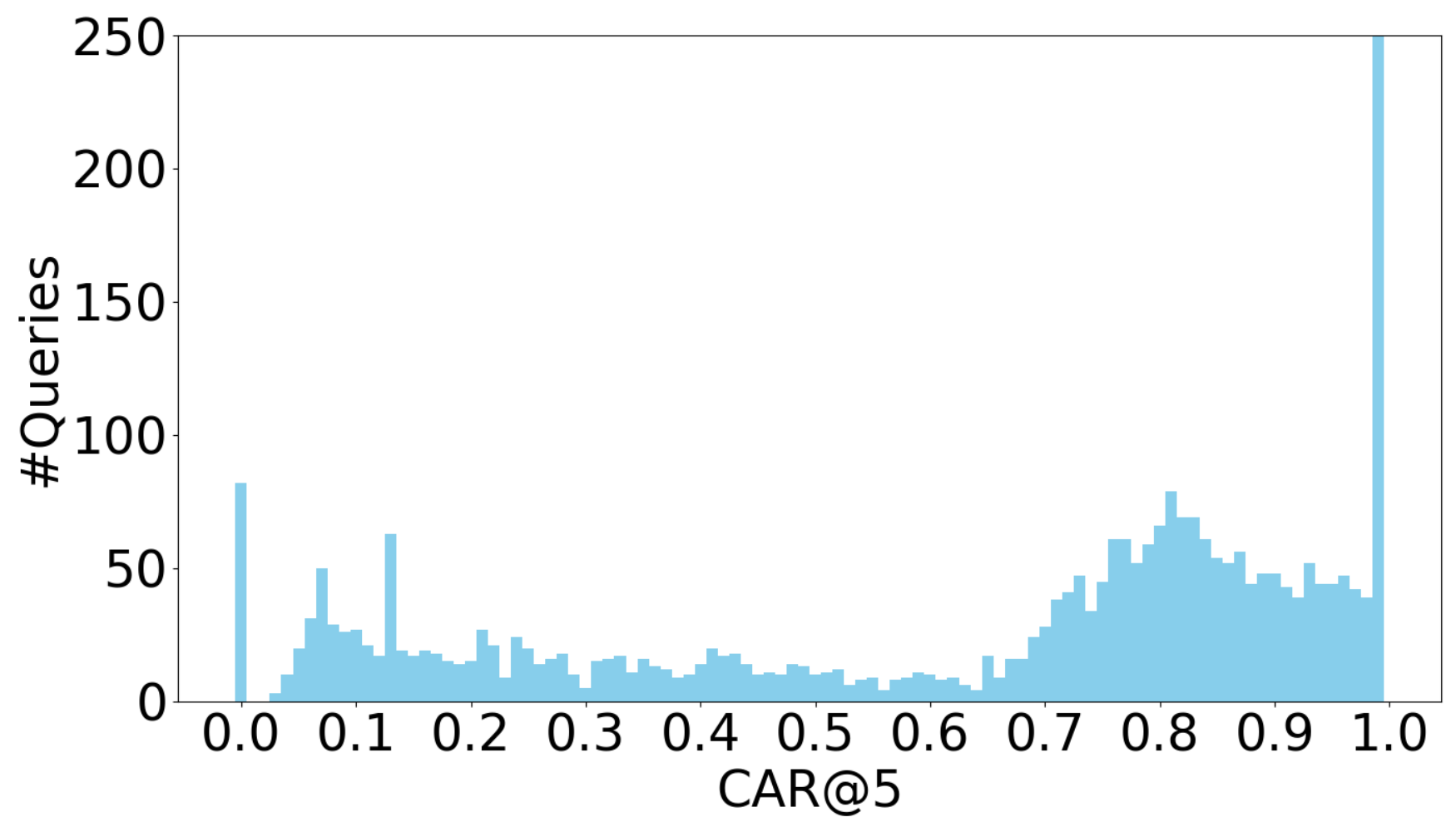}
        \subcaption{$\alpha = 0.6$}
        \label{fig:CAR_sensitivity_dist:f}
    \end{minipage}   
    \begin{minipage}{0.32\linewidth}
        \centering
        \includegraphics[width=\textwidth]{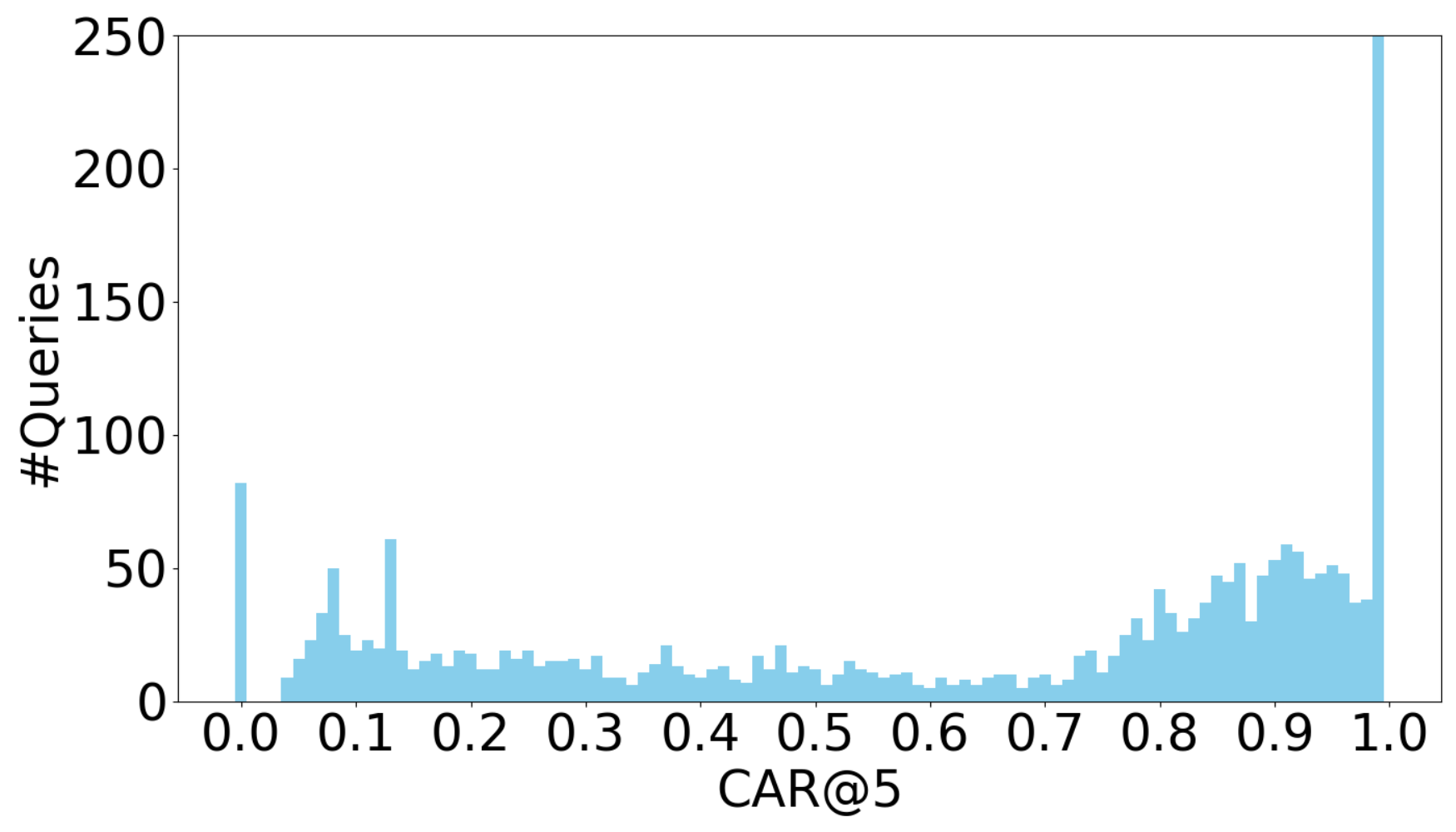}
        \subcaption{$\alpha = 0.7$}
        \label{fig:CAR_sensitivity_dist:g}
    \end{minipage}   
    \begin{minipage}{0.32\linewidth}
        \centering
        \includegraphics[width=\textwidth]{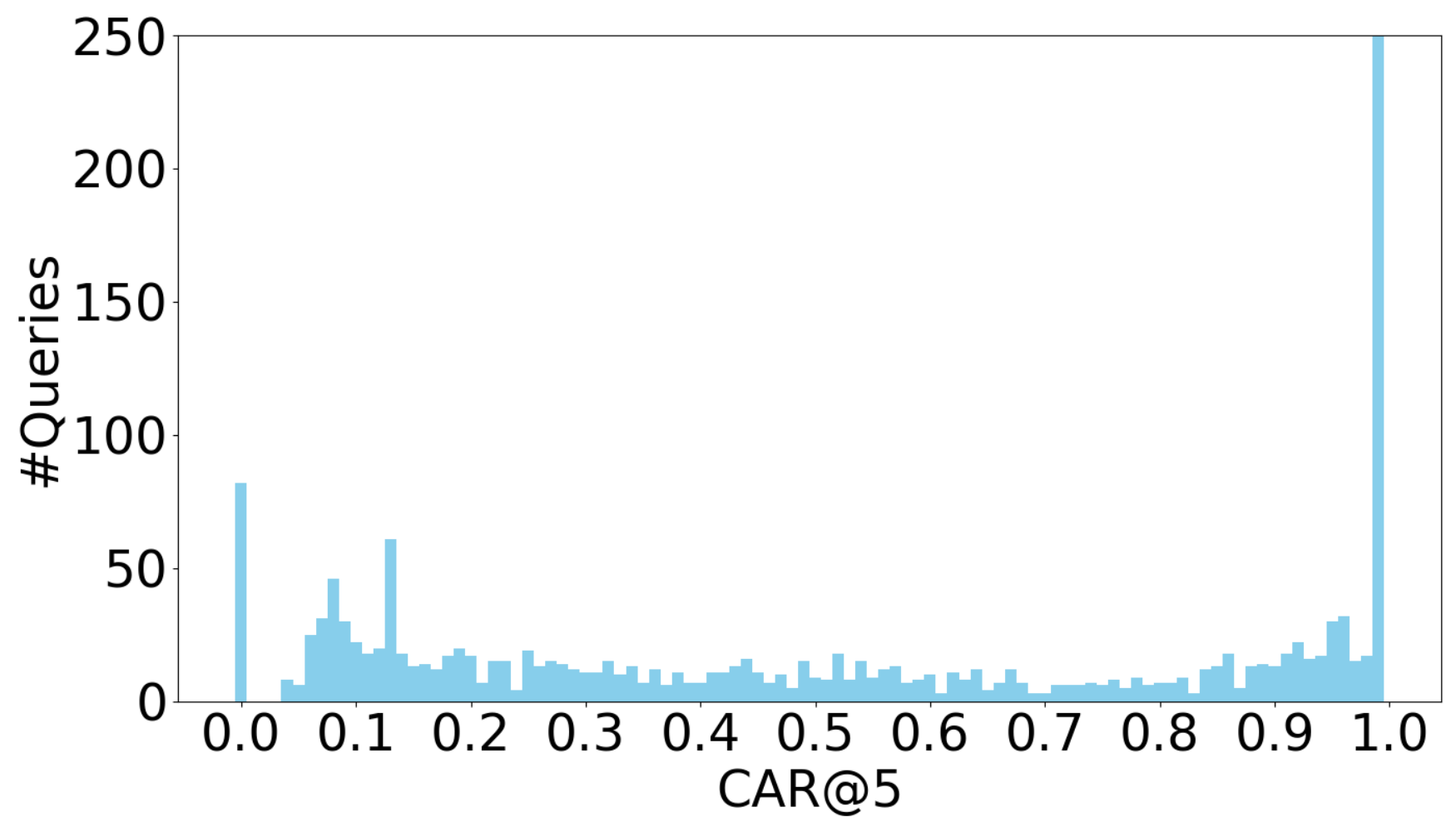}
        \subcaption{$\alpha = 0.8$}
        \label{fig:CAR_sensitivity_dist:h}
    \end{minipage}
    \begin{minipage}{0.32\linewidth}
        \centering
        \includegraphics[width=\textwidth]{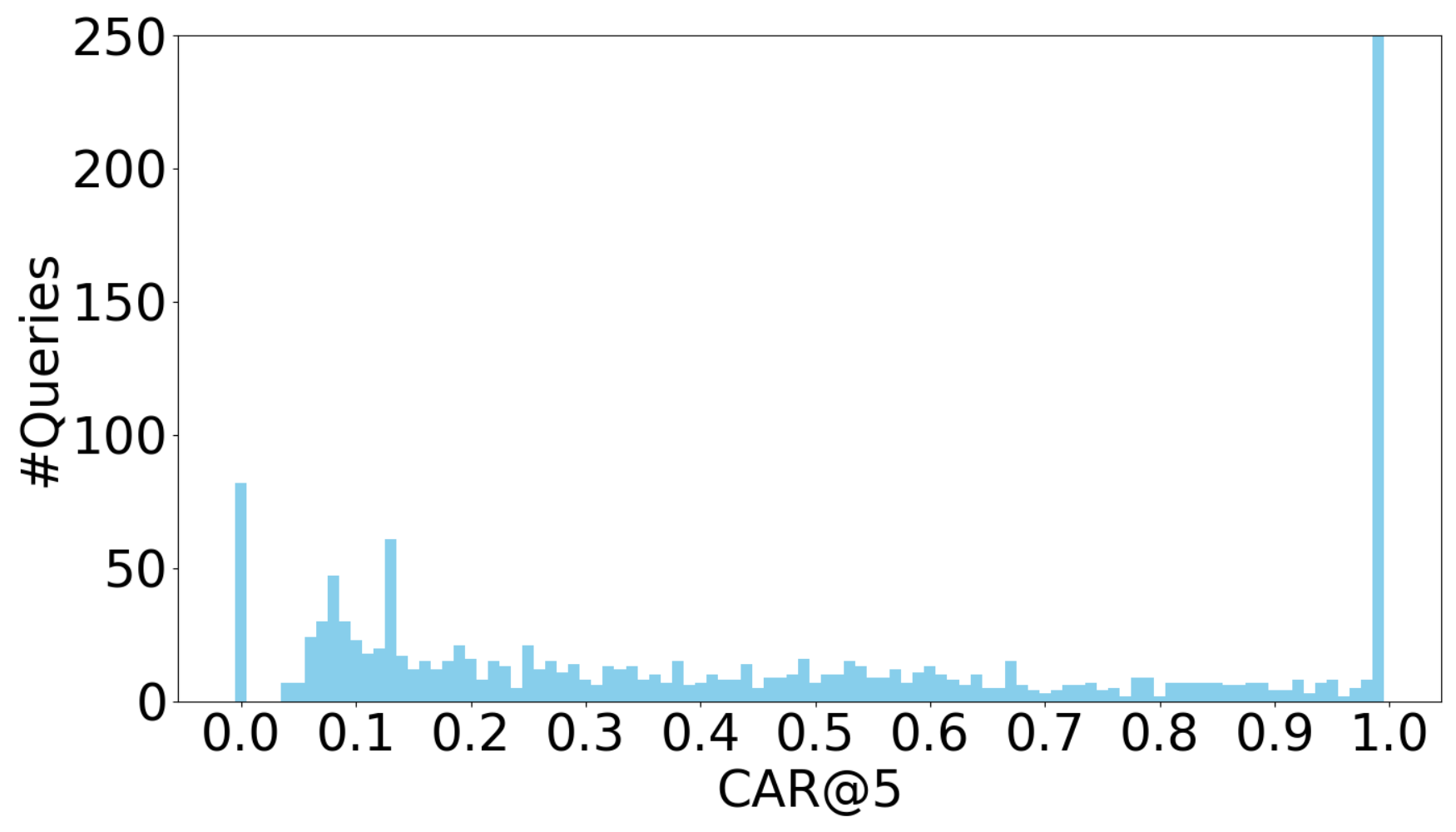}
        \subcaption{$\alpha = 0.9$}
        \label{fig:CAR_sensitivity_dist:i}
    \end{minipage}   
    \caption{
        Distribution of CAR@5 scores across test queries for different values of $\alpha$.
        Each histogram represents the number of queries (\#Queries) for a given CAR@5 score, illustrating how the score distribution shifts as $\alpha$ increases.
        At lower $\alpha$ values, CAR@5 scores are more compressed, while higher $\alpha$ values lead to a broader spread with an increasing concentration near 1.0.
    }
    \label{fig:CAR_sensitivity_dist}
\end{figure*}

\begin{figure*}[!t]
    \centering
    \includegraphics[width=0.5\textwidth]{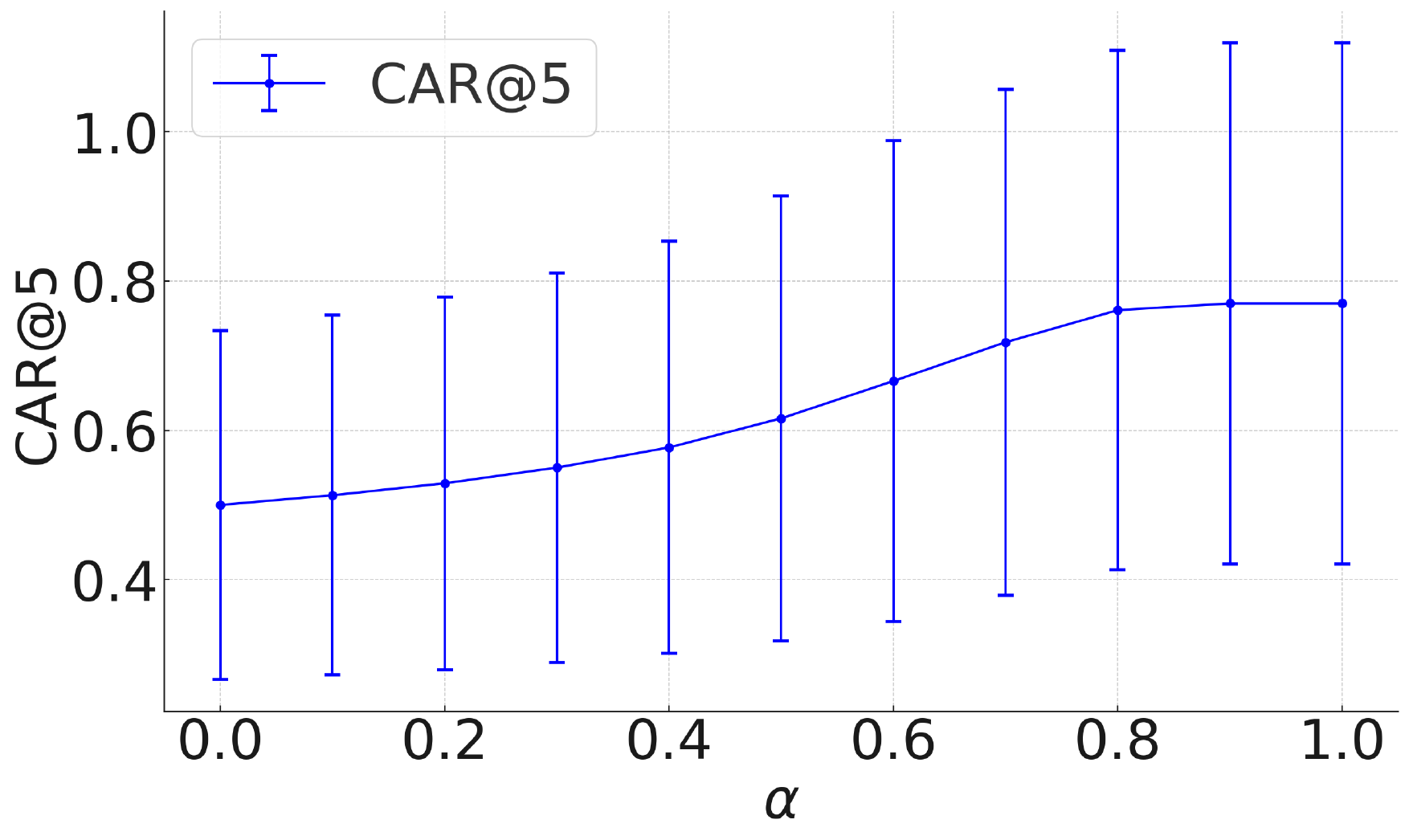}
    \caption{
        Mean and standard deviation of CAR@5 scores across test queries for different values of $\alpha$.
        As $\alpha$ increases, the average CAR@5 score gradually rises, indicating reduced penalization effects on model's confidence.
    }
    \label{fig:CAR_sensitivity_mean}
\end{figure*}

We analyze the sensitivity of the confidence adjustment term $\mathcal{C}(P,k)$ in CAR@$k$ with respect to the threshold parameter $h$.
For this analysis, we reformulate it as a fraction of the maximum entropy, with $h = \alpha H_{\mathrm{max}}(P)$, where $\alpha \in [0,1]$ adjusts the sensitivity of confidence evaluation.
To evaluate the effect of $\alpha$, we measured the distribution of CAR@5 scores across test queries under different threshold values (see \cref{fig:CAR_sensitivity_dist} for score distributions and \cref{fig:CAR_sensitivity_mean} for their means and standard deviations).
When $\alpha$ is too low, $\mathcal{C}(P,k)$ tends to underestimate confidence even when the relevance distribution is sharply peaked, leading to compressed CAR@$k$ scores near the lower end of the $[0,1]$ range.  
Conversely, excessively high $\alpha$ values yield overconfident $\mathcal{C}(P,k)$ estimates even for flat or ambiguous predictions, inflating CAR@$k$ and reducing its discriminative power.
We adopt $\alpha = 0.5$ as a balanced setting, offering interpretable and discriminative scores.
Notably, Changing $\alpha$ does not alter query rankings, ensuring CAR remains valid for comparative evaluation.

\section{Equipment Details}
\label{app:equipment_details}

\begin{figure*}[!t]
    \centering
    \begin{minipage}{0.9\textwidth}
        \centering
        \includegraphics[width=\textwidth]{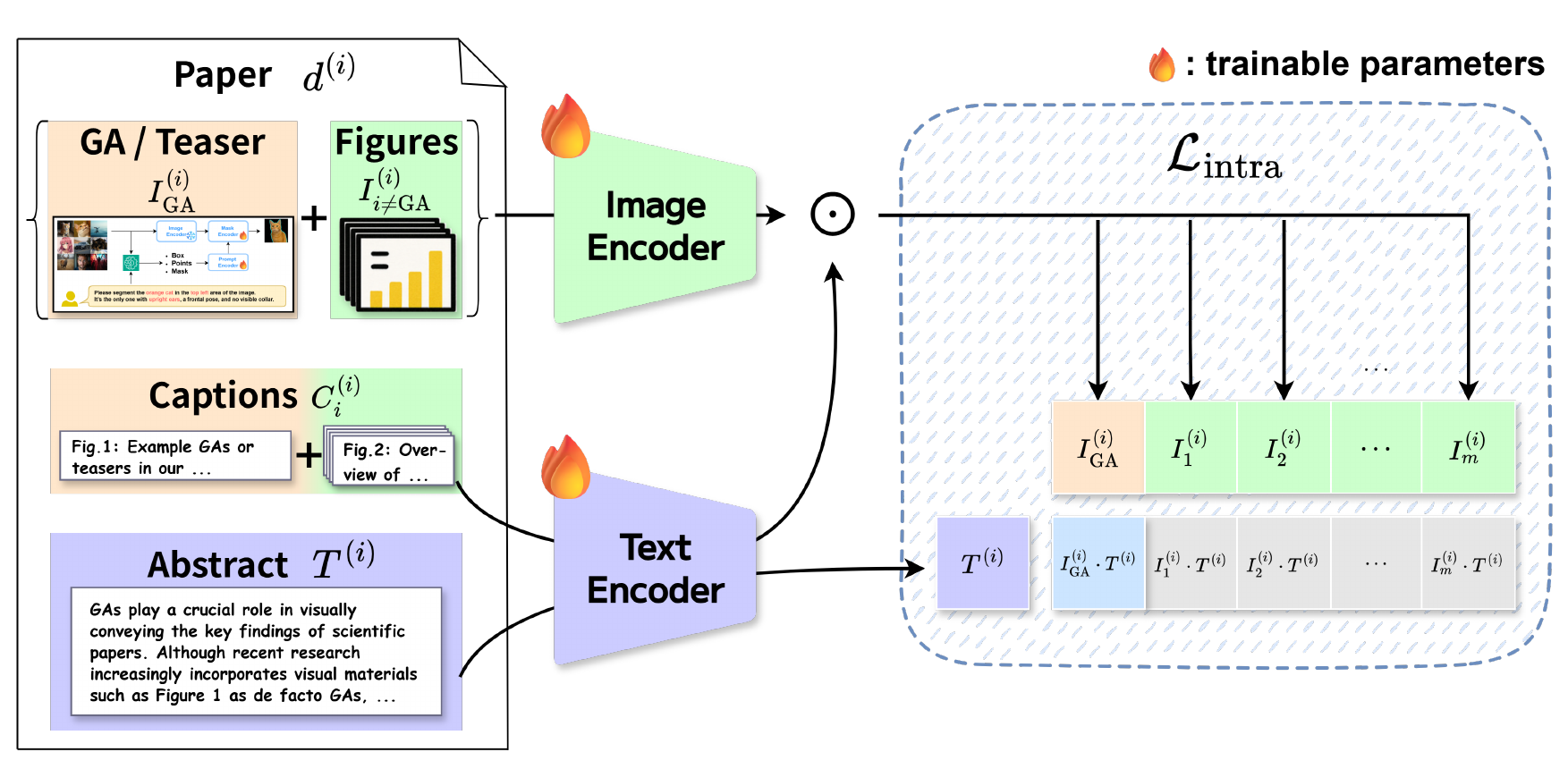}
        \subcaption{Intra-GA Recommendation}
        \label{fig:model_overview:a}
    \end{minipage}
    \begin{minipage}{0.9\textwidth}
        \centering
        \includegraphics[width=\textwidth]{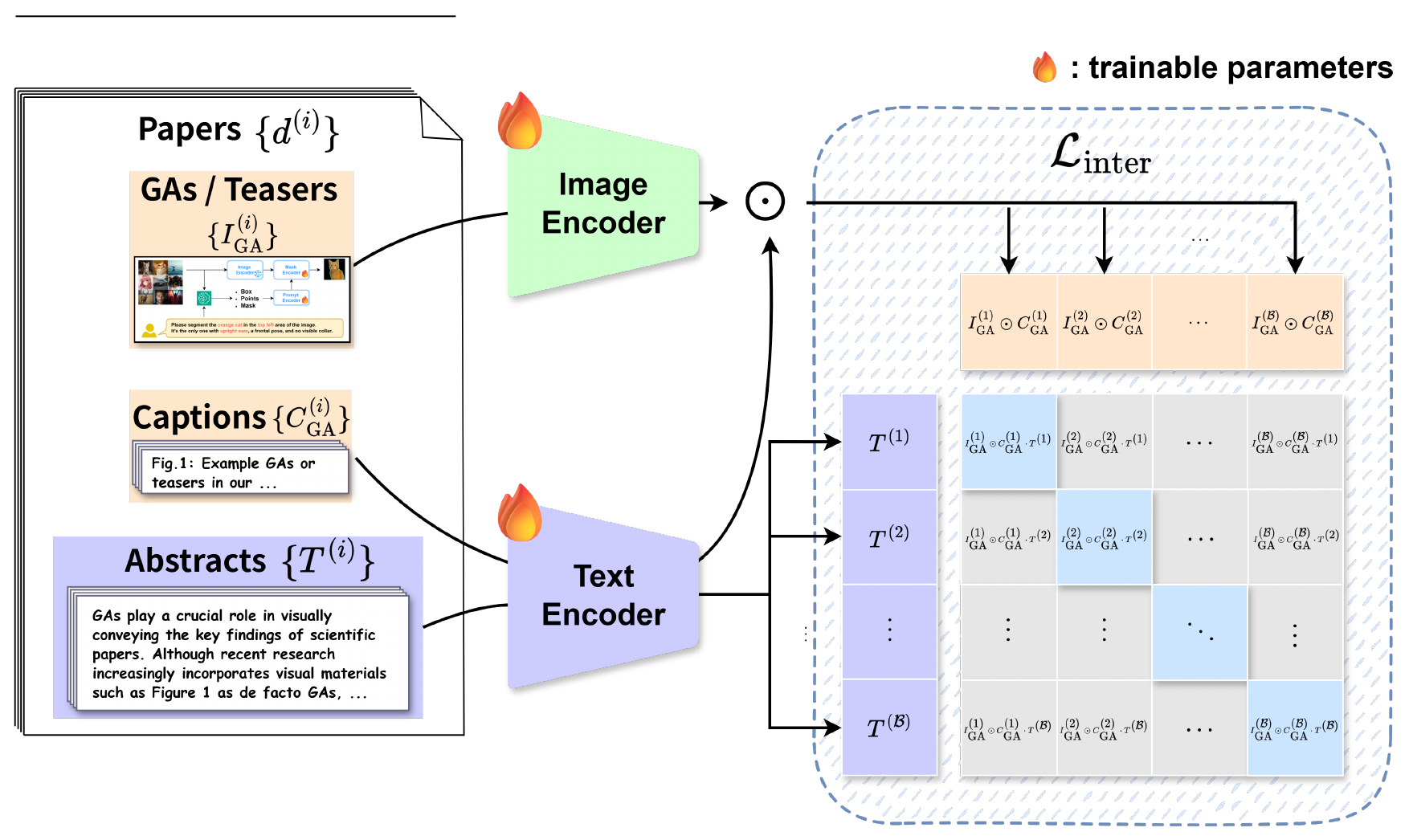}
        \subcaption{Inter-GA Recommendation}
        \label{fig:model_overview:b}
    \end{minipage}   
    \caption{
        Overview of the contrastive learning framework for method (iv) Abs2Fig w/cap applied to (a) Intra-GA Recommendation and (b) Inter-GA Recommendation. Both frameworks encode figures and texts (abstracts and captions) separately into embeddings, optimizing contrastive losses ($\mathcal{L}_{\mathrm{Intra}}$, $\mathcal{L}_{\mathrm{Inter}}$) to align semantically or visually related pairs. The flame icon indicates trainable model components.
    }
    \label{fig:model_overview}
\end{figure*}

This section outlines the experimental setup used in our study, including backbone models, hyper-parameter settings, and computational resources.

\nbf{Model Architectures}
Our experiments utilized representative pretrained backbone models listed in \cref{tbl:weights}.
Among the tested methods, (iv) Abs2Fig w/cap involved a lightweight architectural modification that integrates figure and caption embeddings via the Hadamard product.   
The overall structure is illustrated in \cref{fig:model_overview}.

\begin{table*}[!t]
    \caption{
        Pretrained Weights for Backbone Models.
    }
    \label{tbl:weights}
    \centering
    \small
    \begin{tabular}{lll}
        \toprule
        \textbf{Method} & \textbf{Backbone Model} & \textbf{Pre-trained Weight} \\
        
        \midrule
        (i) Abs2Cap
        & BERTScore~\cite{zhang2020bertscore}
            & \href{https://huggingface.co/allenai/scibert_scivocab_uncased}{\texttt{allenai/scibert\_scivocab\_uncased}} \\

        \cmidrule(r){1-1} \cmidrule(lr){2-2} \cmidrule(l){3-3}
 
        \multirow{5}{*}{(ii) GA-binCl} 
        & EfficientNetV2~\cite{tan2021efficientnetv2}   
            & \href{https://pytorch.org/vision/main/models/generated/torchvision.models.efficientnet_v2_l.html}{\texttt{EfficientNet\_V2\_L\_Weights.IMAGENET1K\_V1}} \\
        & ViT~\cite{dosovitskiy2020vit}                  
            & \href{https://huggingface.co/google/vit-large-patch16-224-in21k}{\texttt{google/vit-large-patch16-224-in21k}} \\
        & CLIP image encoder~\cite{radford2022clip}                
            & \href{https://huggingface.co/openai/clip-vit-large-patch14}{\texttt{openai/clip-vit-large-patch14}} \\
        & SwinTransformerV2~\cite{liu2022swin-transformer}
            & \href{https://huggingface.co/microsoft/swin-large-patch4-window7-224-in22k}{\texttt{microsoft/swin-large-patch4-window7-224-in22k}} \\
        & ConvNeXtV2~\cite{woo2023convnext}               
            & \href{https://huggingface.co/facebook/convnextv2-large-22k-224}{\texttt{facebook/convnextv2-large-22k-224}} \\

        \cmidrule(r){1-1} \cmidrule(lr){2-2} \cmidrule(l){3-3}
        
        & CLIP~\cite{radford2022clip}         
            & \href{https://huggingface.co/openai/clip-vit-large-patch14}{\texttt{openai/clip-vit-large-patch14}} \\
        & BLIP-2~\cite{li2023blip-2}       
            & \href{https://huggingface.co/Salesforce/blip2-itm-vit-g}{\texttt{Salesforce/blip2-itm-vit-g}} \\
        (iii) Abs2Fig
        & X${}^2$-VLM~\cite{zeng2023x2-vlm}
            & \href{https://github.com/zengyan-97/X2-VLM}{\texttt{X2VLM-large (4M)}} \\
        (iv) Abs2Fig w/cap
        & OpenCLIP~\cite{cherti2023open-clip}
            & \href{https://huggingface.co/laion/CLIP-ViT-L-14-laion2B-s32B-b82K}{\texttt{laion/CLIP-ViT-L-14-laion2B-s32B-b82K}} \\
        & SigLIP2~\cite{tschannen2025siglip2}
            & \href{https://huggingface.co/google/siglip2-large-patch16-256}{\texttt{google/siglip2-large-patch16-256}} \\
        & Long-CLIP~\cite{zhang2024long-clip}
            & \href{https://huggingface.co/BeichenZhang/LongCLIP-L}{\texttt{BeichenZhang/LongCLIP-L}} \\

        \bottomrule
    \end{tabular}
\end{table*}

\vspace{-1mm}

\nbf{Hyper-parameter Settings}
We summarize the final hyper-parameter settings for all models in \cref{tbl:hyperparams}.
All models were trained and evaluated under a hold-out validation protocol, with learning rate, batch size, and the number of sampled figures per paper $m$ in Intra-GA Recommendation explored over
\{$1~\times~10^{-7}$, $5~\times~10^{-7}$, $1~\times~10^{-6}$, $5~\times~10^{-6}$, $1~\times~10^{-5}$\},
\{64, 128, 256, 512, 1024, 2048\}, and
\{5, 6, 7, 8, 9\}, respectively.
Other hyper-parameters followed prior literature or official implementations without further tuning.
All reported scores are averaged over five independent runs with different random seeds to account for training variability.  

\begin{table}[!t]
    \caption{
        Detailed training hyper-parameters.
    }
    \label{tbl:hyperparams}
    \centering
    \small
    \begin{tabular}{lr}
        \toprule
        \textbf{Hyper-Parameter} & \textbf{Value} \\
        \midrule
        \rowcolor[HTML]{E0E0E0}
        \multicolumn{2}{c}{For (ii) GA-binCl} \\
        \midrule
        Epochs                   &                30 \\
        Batch Size               &               128 \\
        Learning Rate            &              5e-7 \\
        LR Scheduler             &  Cosine Annealing \\
        Mixed Precision Training &     Enabled (AMP) \\
        Weight Decay             &              1e-3 \\
        AdamW $\beta_1$          &               0.9 \\
        AdamW $\beta_2$          &             0.999 \\
        AdamW $\epsilon$         &              1e-8 \\
        Loss Weighting           & Inverse Frequency \\
        \midrule
        \rowcolor[HTML]{E0E0E0}
        \multicolumn{2}{c}{For (iii) Abs2Fig / (iv) Abs2Fig w/cap} \\
        \midrule
        Epochs                        &               15 \\
        Batch Size                    &             1024 \\
        Sampled Figures per Paper $m$ &                7 \\
        Learning Rate                 &             1e-6 \\
        LR Scheduler                  & Cosine Annealing \\
        Mixed Precision Training      &    Enabled (AMP) \\
        Weight Decay                  &             1e-3 \\
        AdamW $\beta_1$               &              0.9 \\
        AdamW $\beta_2$               &            0.999 \\
        AdamW $\epsilon$              &             1e-8 \\
        Temperature $\tau$            &             0.07 \\
        \bottomrule
    \end{tabular}
\end{table}

\nbf{Computational Environment}
Experiments were conducted using an NVIDIA RTX A6000 GPU (48 GB).
Training and evaluation took approximately 12 hours for Intra-GA Recommendation and 8 hours for Inter-GA Recommendation.

\section{Additional Results}
\label{app:additional_results}


\subsection{Intra-GA Recommendation}
\begin{figure*}[!t]
    \centering
    \includegraphics[width=0.83\textwidth]{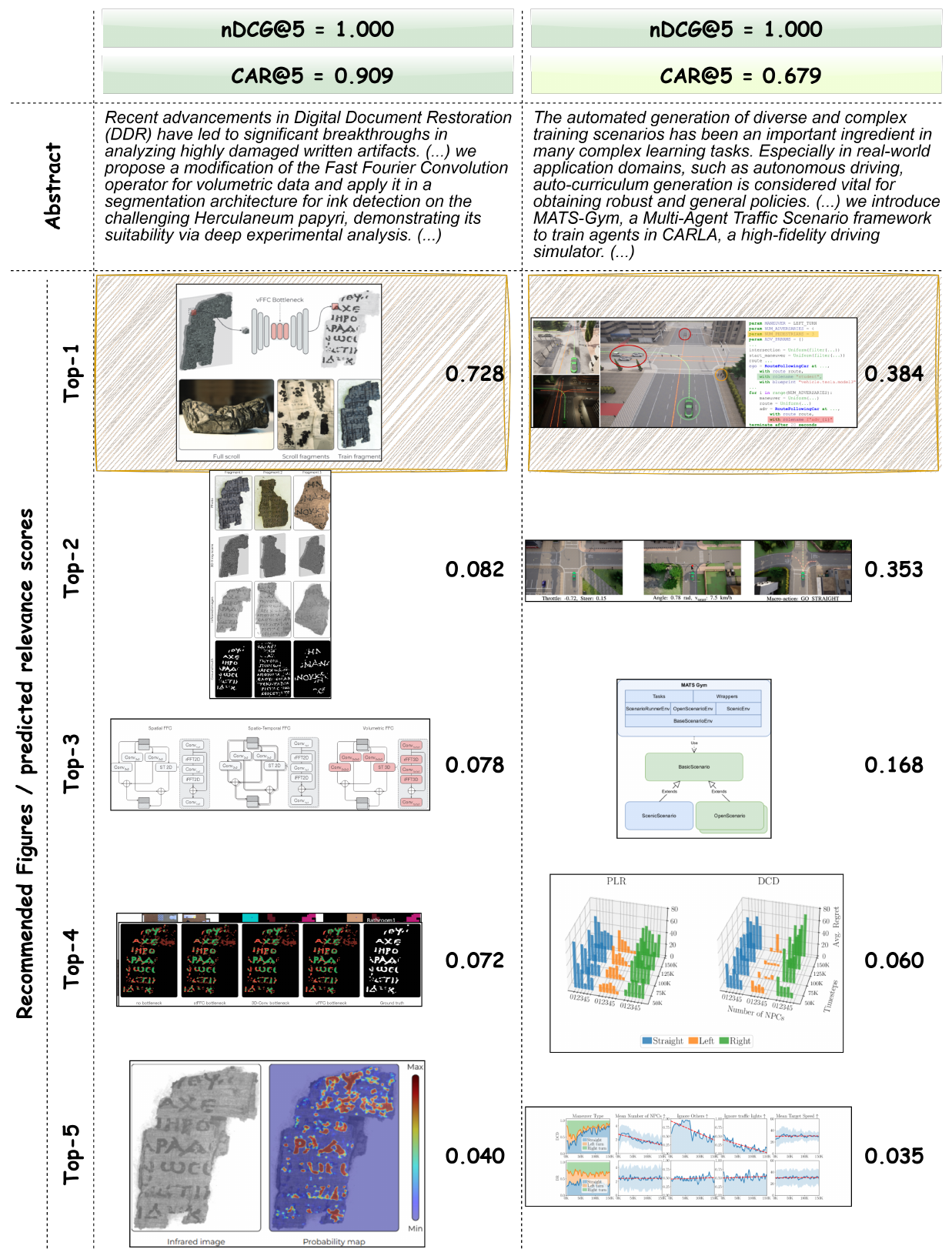}
    \caption{
        Qualitative examples of Intra-GA Recommendation results obtained by the best-performing model (Long-CLIP within method (iv) Abs2Fig w/cap). \protect \footnotemark[9]
        The yellow-highlighted figures represent GTs.
    }
    \label{fig:intra_example_1}
\end{figure*}

\begin{figure*}[!t]
    \centering
    \includegraphics[width=0.83\textwidth]{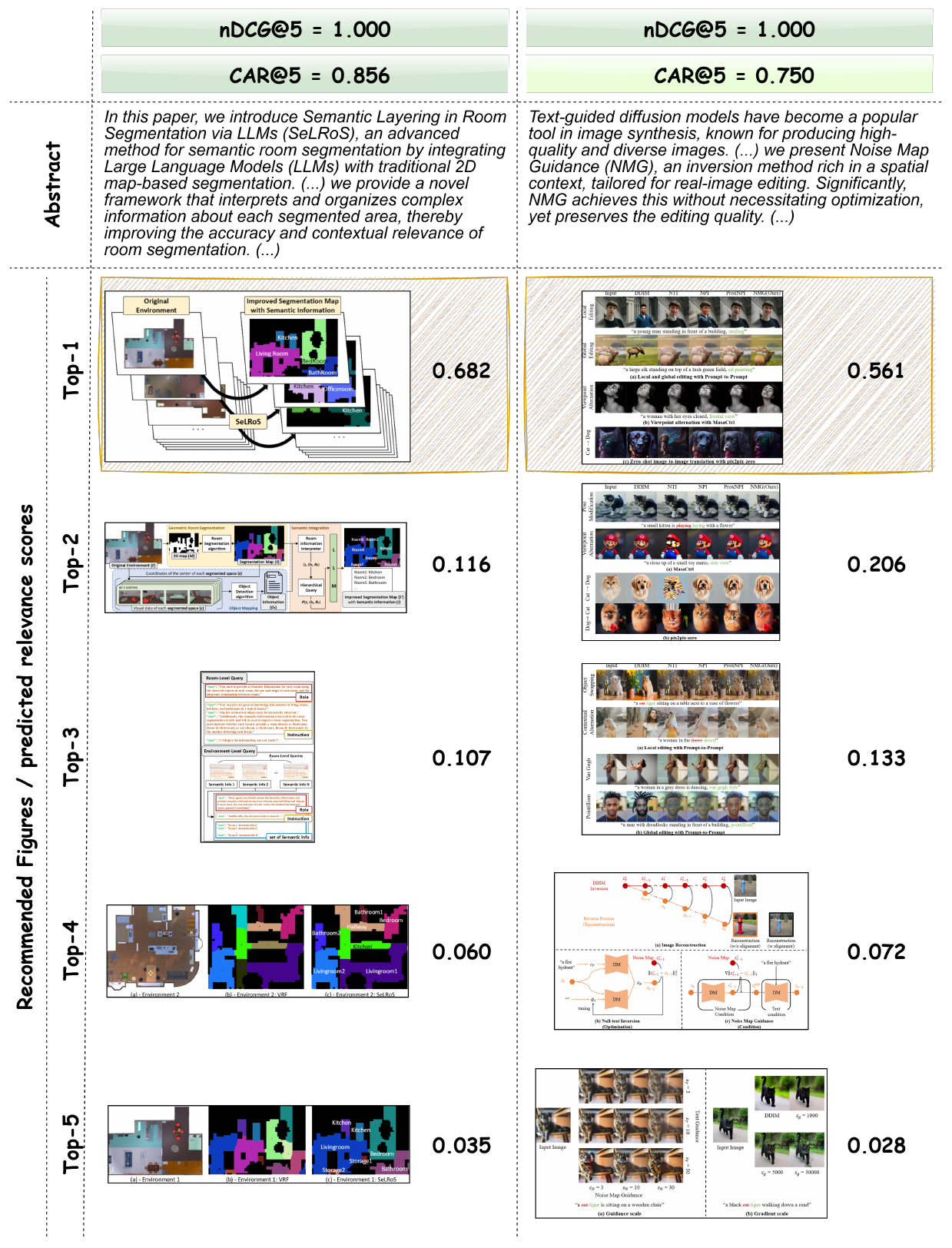}
    \caption{
        Qualitative examples of Intra-GA Recommendation results obtained by the best-performing model (Long-CLIP within method (iv) Abs2Fig w/cap). \protect \footnotemark[10]
        The yellow-highlighted figures represent GTs.
    }
    \label{fig:intra_example_2}
\end{figure*}

\begin{figure*}[!t]
    \centering
    \includegraphics[width=0.83\textwidth]{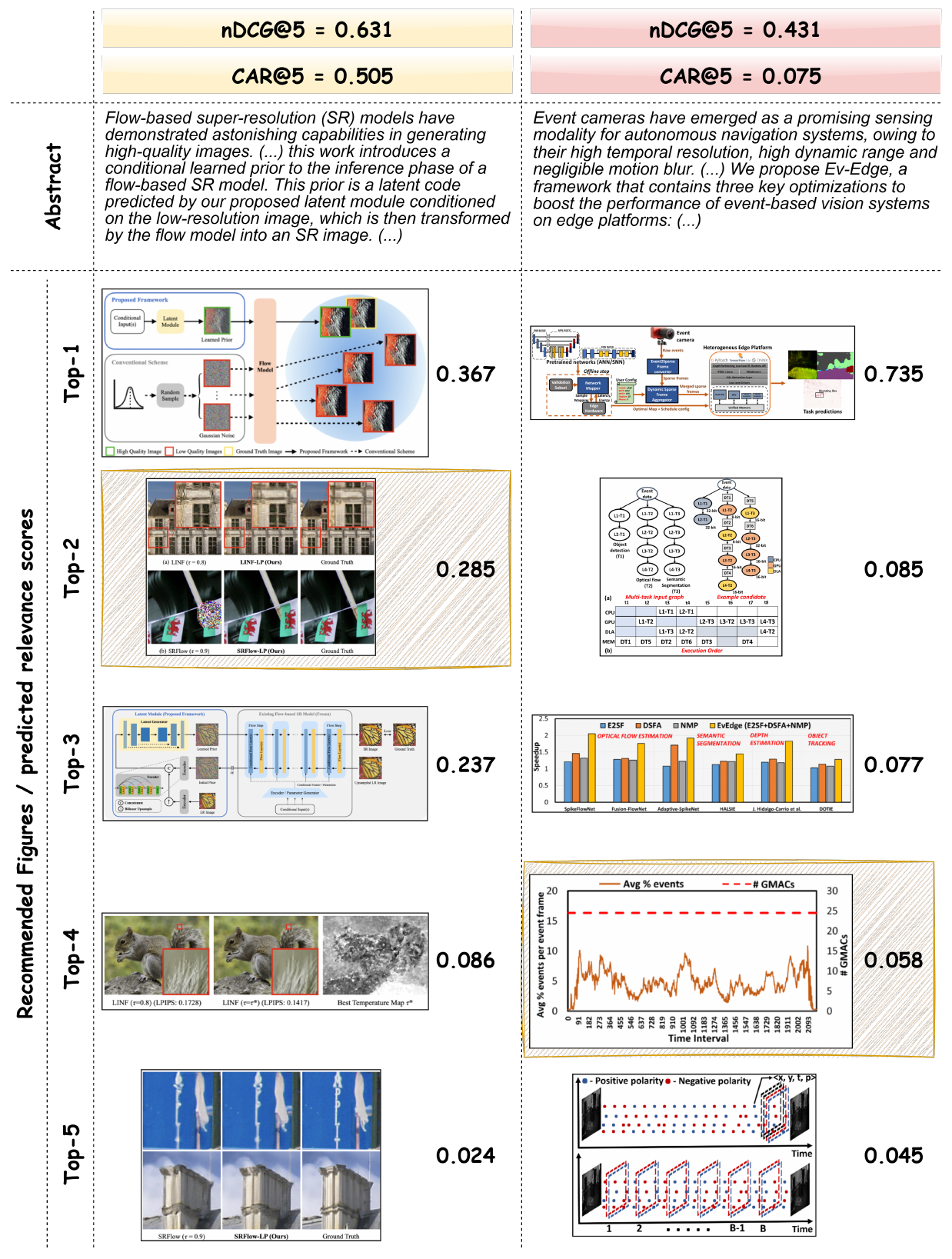}
    \caption{
        Qualitative examples of Intra-GA Recommendation results obtained by the best-performing model (Long-CLIP within method (iv) Abs2Fig w/cap). \protect \footnotemark[11]
        The yellow-highlighted figures represent GTs.
    }
    \label{fig:intra_example_3}
\end{figure*}

\begin{figure*}[!t]
    \centering
    \includegraphics[width=0.83\textwidth]{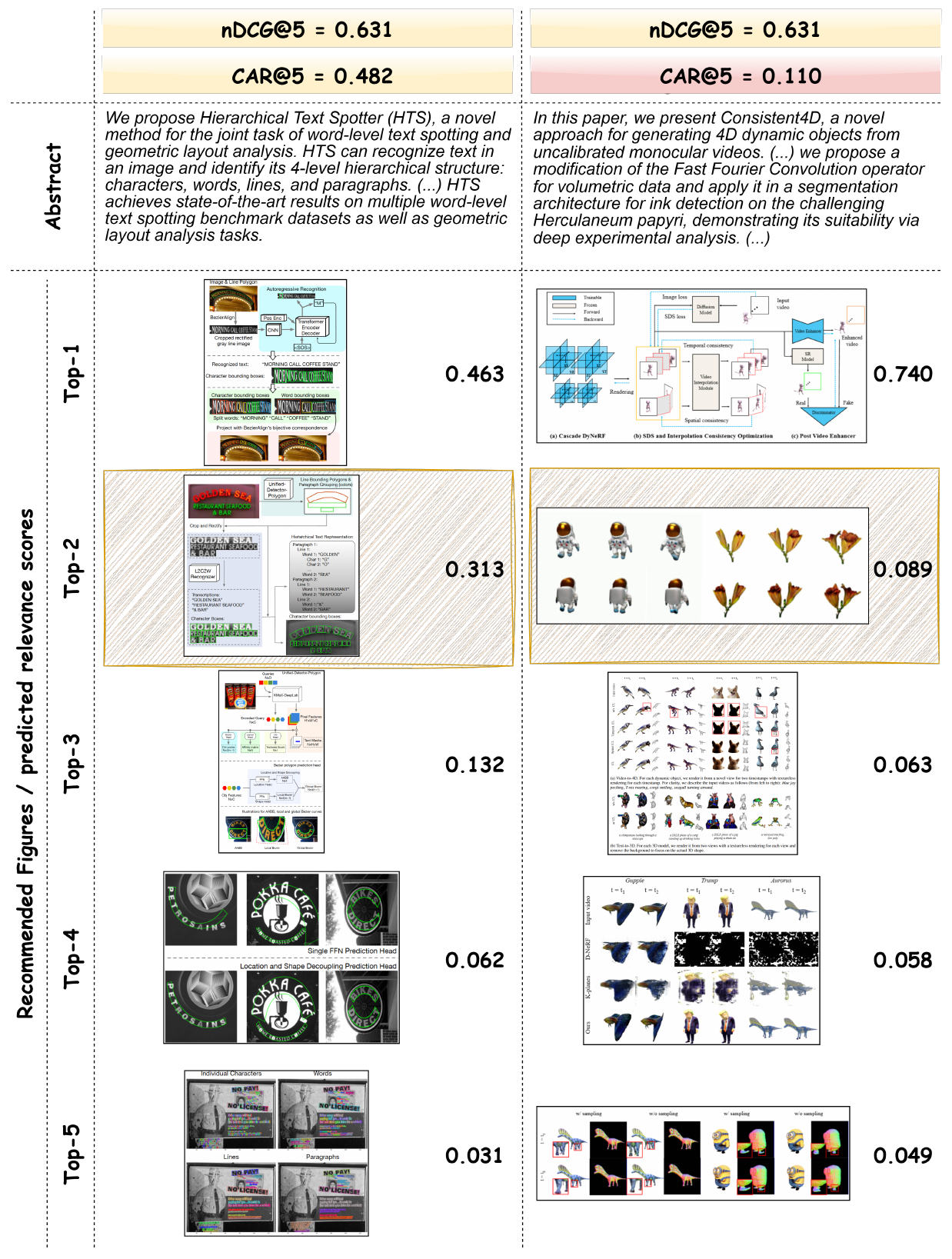}
    \caption{
        Qualitative examples of Intra-GA Recommendation results obtained by the best-performing model (Long-CLIP within method (iv) Abs2Fig w/cap). \protect \footnotemark[12]
        The yellow-highlighted figures represent GTs.
    }
    \label{fig:intra_example_4}
\end{figure*}

\nbf{Qualitative Evaluation}
We analyze representative cases from the best-performing model (Long-CLIP~\cite{zhang2024long-clip} within method (iv) Abs2Fig w/cap) to understand its behavior in Intra-GA Recommendation.
Figs.~\ref{fig:intra_example_1}, \ref{fig:intra_example_2}, \ref{fig:intra_example_3} and \ref{fig:intra_example_4} illustrate typical outputs, and corresponding CAR and nDCG~\cite{burges2005ndcg}.
The model tends to favor figures with architectural overviews or grid-style layouts, which frequently preferred as GAs or teasers by authors.
When multiple candidates share such designs, scores converge and CAR appropriately reflects this ambiguity with moderate values.
This nuance is often missed by conventional metrics.
In contrast, charts are generally scored lower, suggesting a learned preference for structured, concept-focused visuals.
These results indicate that the model captures not only text-figure relevance but also topic-specific visual conventions linked to effective GA design.
CAR quantifies this instance-level behavior, unifying relevance and confidence into a single interpretable score.

\begin{figure*}[!t]
    \centering
    \begin{minipage}{0.49\linewidth}
        \centering
        \includegraphics[width=\textwidth]{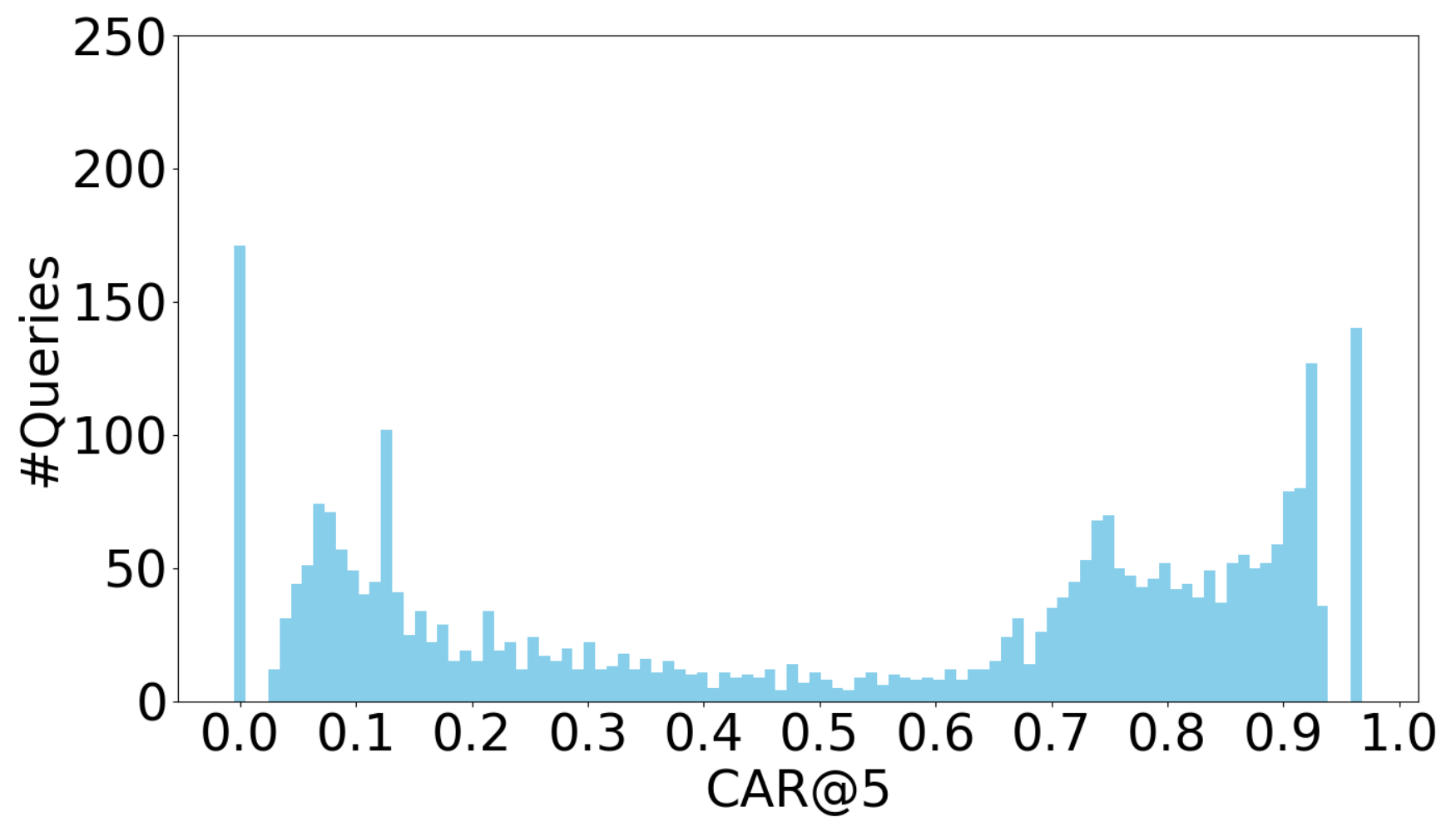}
        \subcaption{(i) Abs2Cap (BM25~\cite{robertson1994bm25})}
        \label{fig:CAR_dist:a}
    \end{minipage}
    \begin{minipage}{0.49\linewidth}
        \centering
        \includegraphics[width=\textwidth]{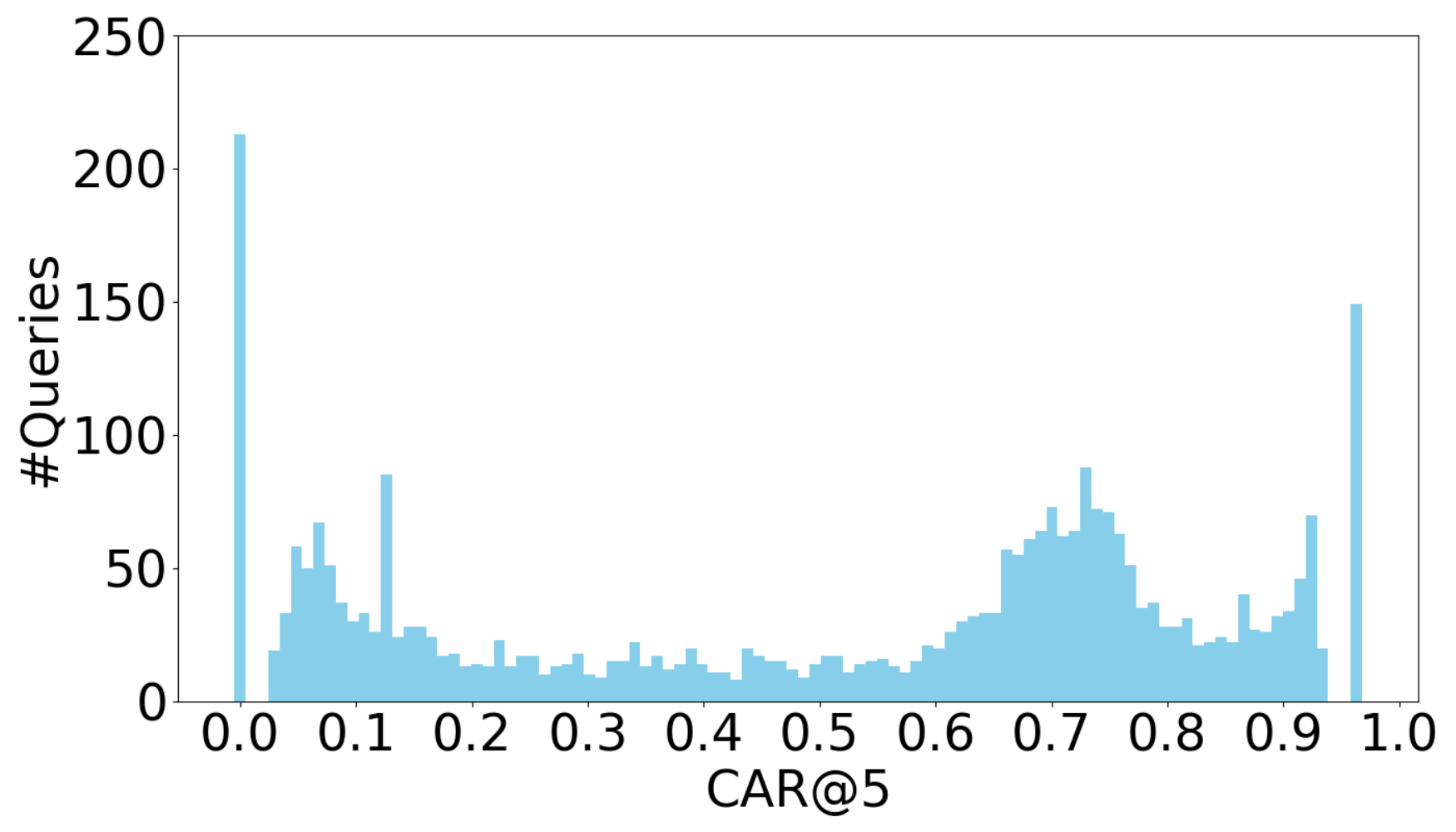}
        \subcaption{(ii) GA-binCl (SwinTransformerV2)}
        \label{fig:CAR_dist:b}
    \end{minipage}   
    \begin{minipage}{0.49\linewidth}
        \centering
        \includegraphics[width=\textwidth]{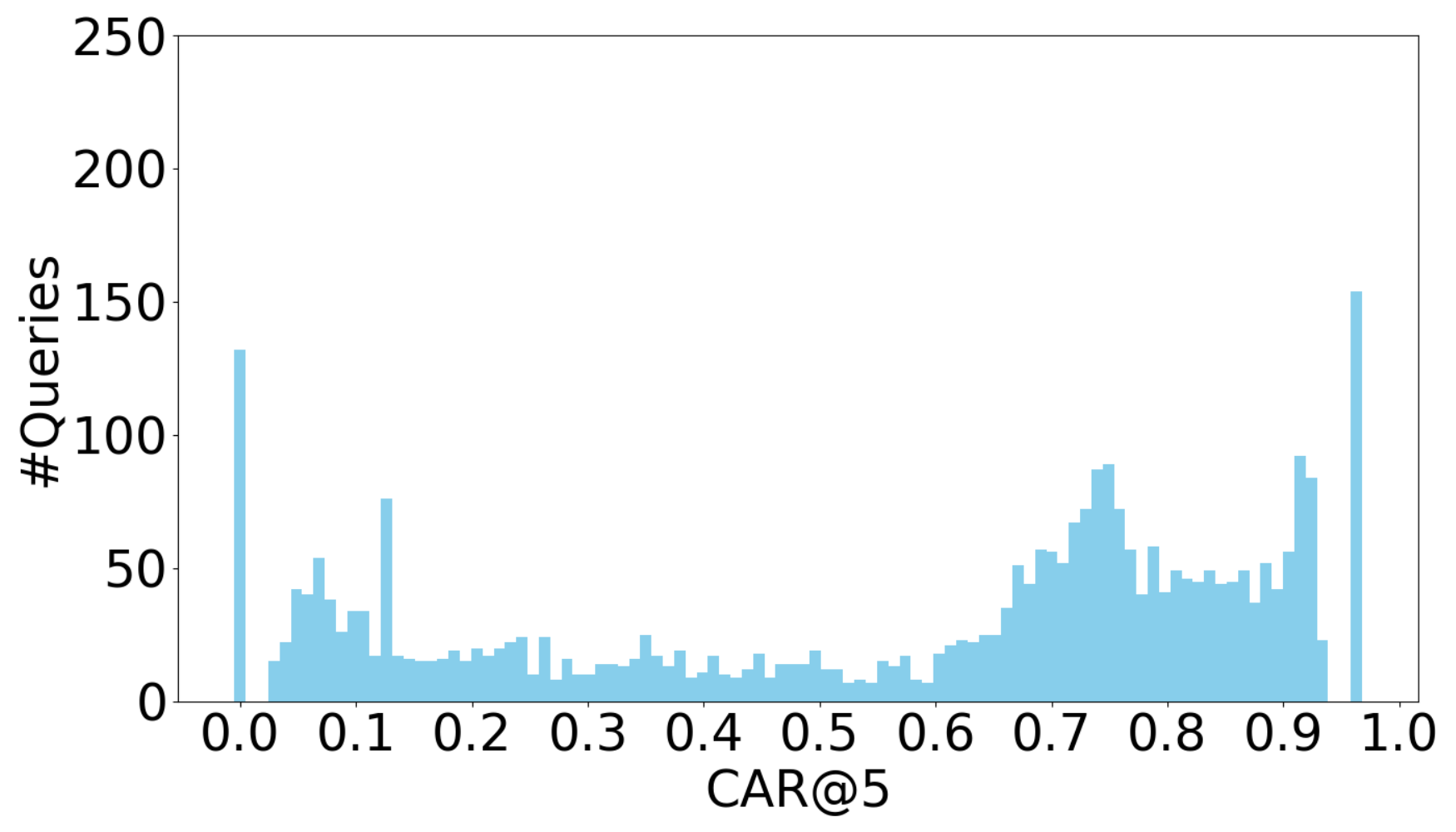}
        \subcaption{(iii) Abs2Fig (BLIP-2)}
        \label{fig:CAR_dist:c}
    \end{minipage}   
    \begin{minipage}{0.49\linewidth}
        \centering
        \includegraphics[width=\textwidth]{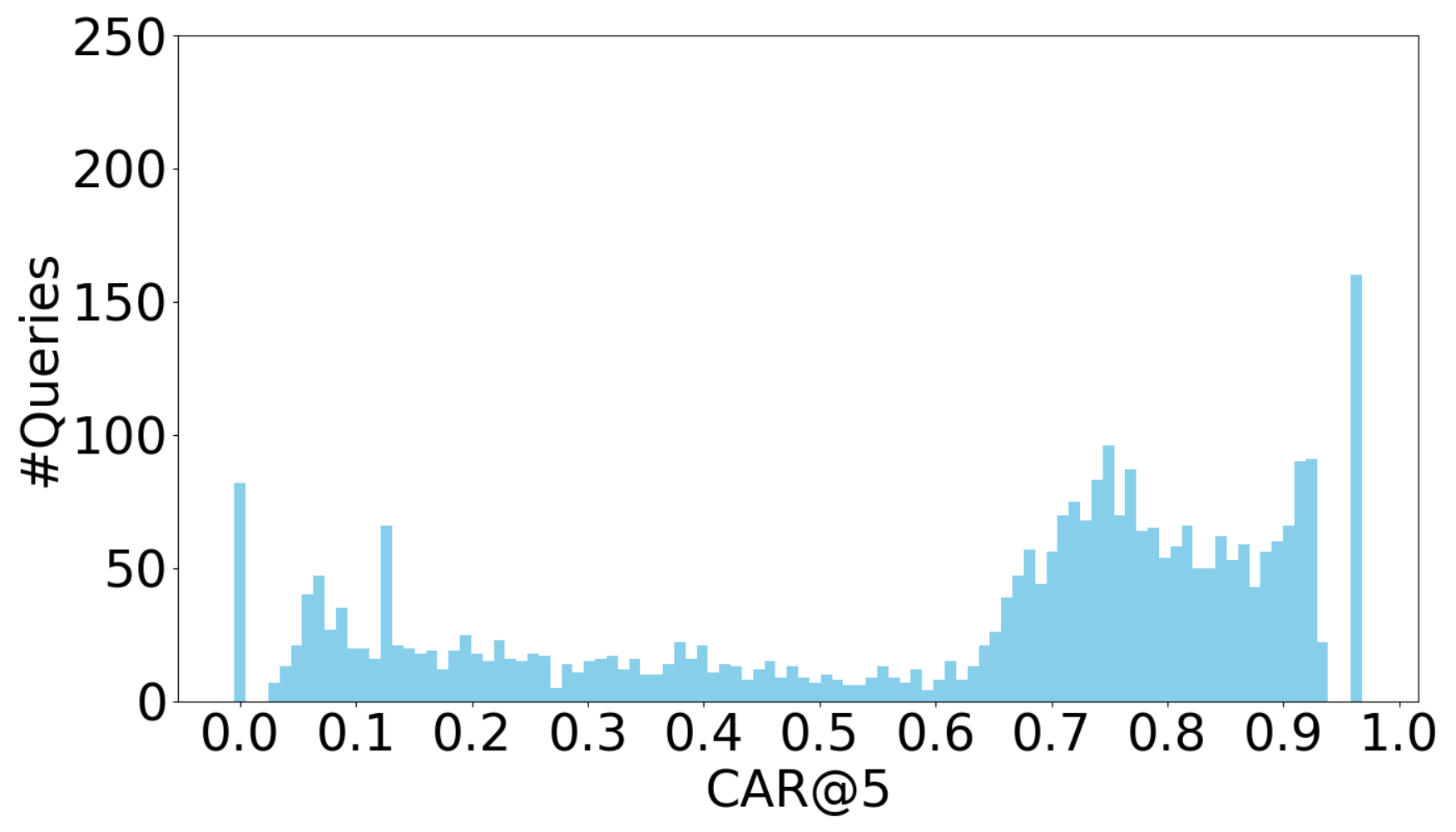}
        \subcaption{(iv) Abs2Fig w/cap (Long-CLIP)}
        \label{fig:CAR_dist:d}
    \end{minipage}   
    \caption{
        Distribution of CAR@5 scores across individual queries for the best-performing models in each Intra-GA Recommendation method.
        Higher CAR@5 values indicate higher model's confidence and more reliable top-ranked GA recommendations.
        Methods with distributions skewed toward higher values reflect stronger model confidence and more effective recommendation performance.
    }
    \label{fig:CAR_dist}
\end{figure*}

\nbf{Distributional Analysis of CAR}
\cref{fig:CAR_dist} shows CAR@5 distributions for each method.
Abs2Cap displays a polarized pattern, working well only with strong lexical cues.
GA-binCl yields more moderate scores ($0.6$--$0.8$) but lacks high-performing cases.
Abs2Fig achieves stronger performance overall ($0.7$--$0.9$), and adding captions (Abs2Fig w/cap) shifts scores above $0.9$ with fewer failures.
These results show that combining captions with visual features improves recommendation quality and consistency.

\begin{figure*}[!t]
    \centering
    \begin{minipage}{0.32\linewidth}
        \centering
        \includegraphics[width=\textwidth]{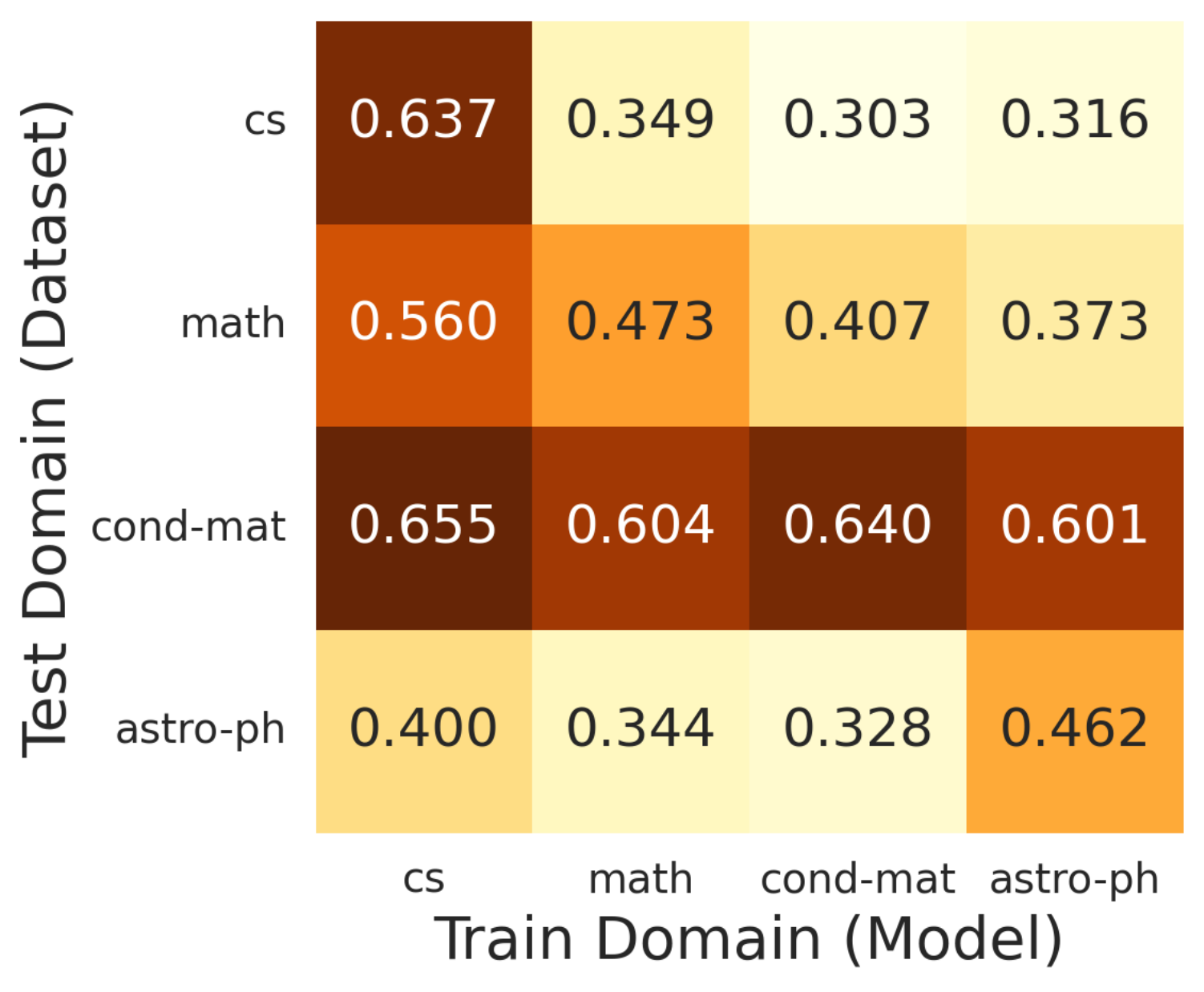}
        \subcaption{R@1}
        \label{fig:cross_domain:a}
    \end{minipage}
    \begin{minipage}{0.32\linewidth}
        \centering
        \includegraphics[width=\textwidth]{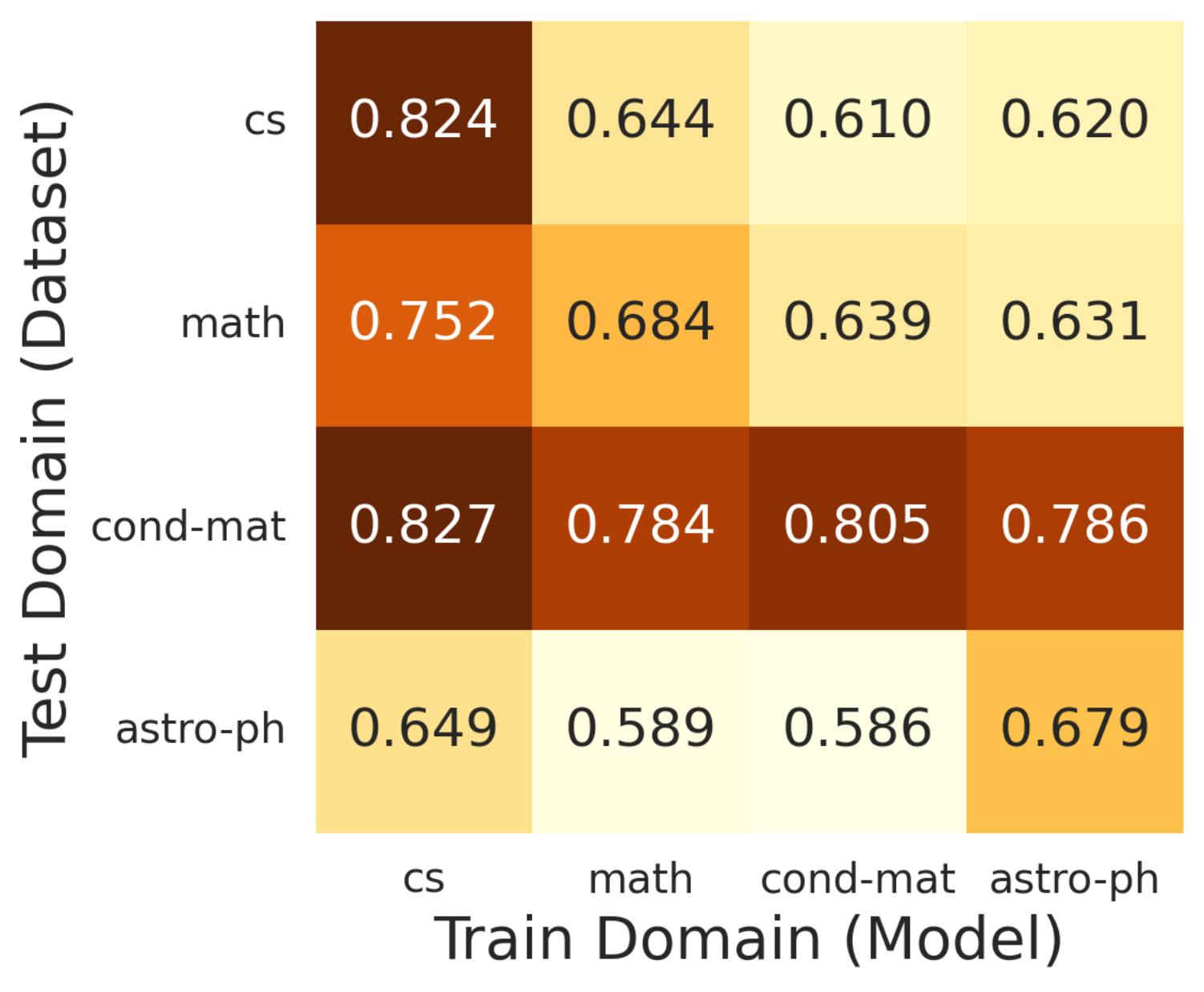}
        \subcaption{nDCG@5}
        \label{fig:cross_domain:b}
    \end{minipage}   
    \begin{minipage}{0.32\linewidth}
        \centering
        \includegraphics[width=\textwidth]{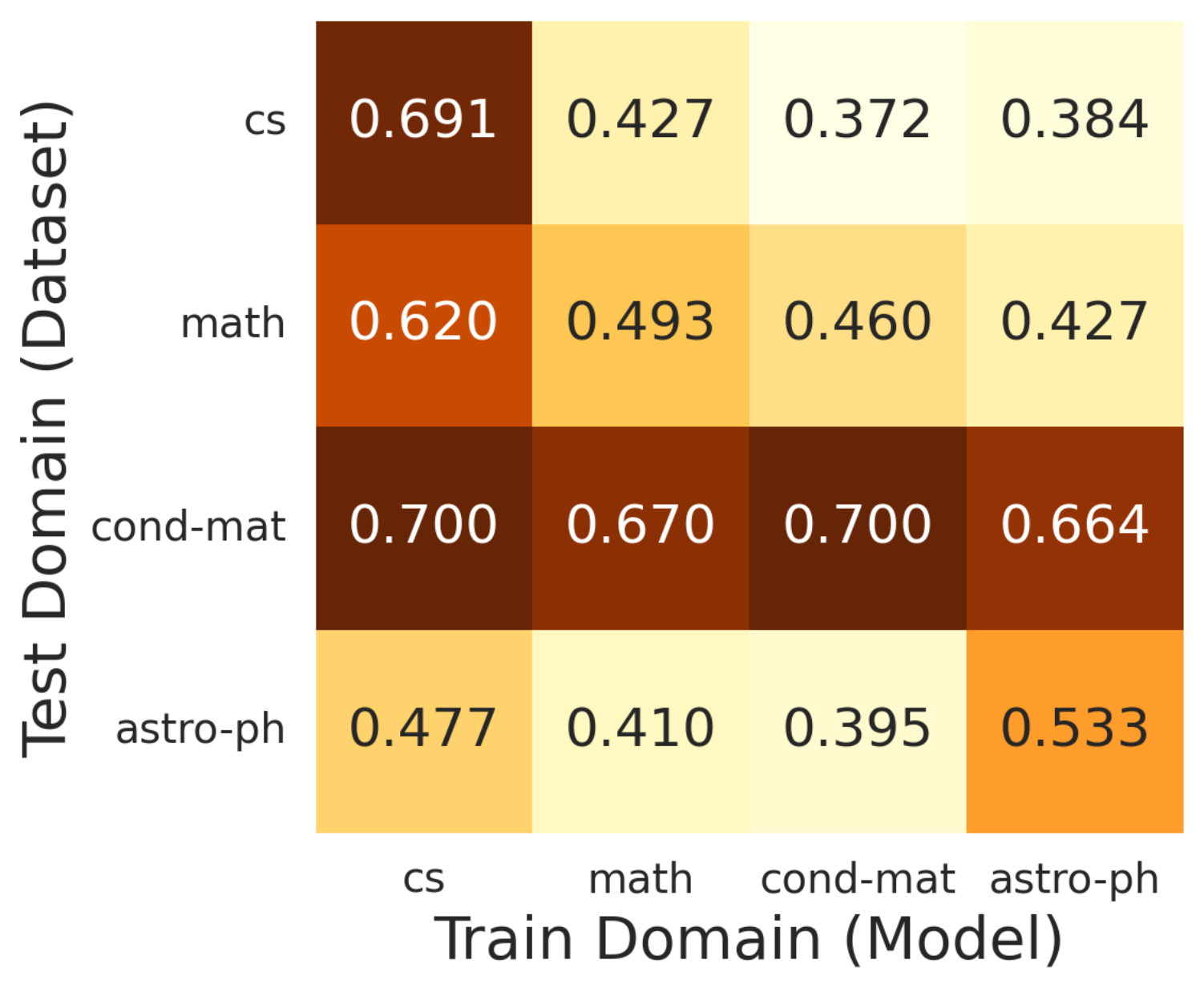}
        \subcaption{CAR@5: 0.5 $\uparrow$}
        \label{fig:cross_domain:c}
    \end{minipage}
    \caption{
        Cross-domain evaluation of Intra-GA Recommendation models trained on different scientific fields.
        We report R@1, nDCG@5, and proportion of predictions with CAR@5 exceeding 0.5.
        Models trained on any domain performed well on the \texttt{cond-mat} test set, indicating that GAs in this field are relatively easy to identify.
        \texttt{cs}-trained models generalize well, possibly due to the widespread use of teaser figures in that field.
    }
    \label{fig:cross_domain}
\end{figure*}

\nbf{Domain-specific and Cross-domain Analysis}
This experiment evaluates how well our GA recommendation model generalizes across scientific domains.  
We trained the Abs2Fig w/cap model using Long-CLIP independently on three domains: \textit{math}, \textit{cond-mat}, and \textit{astro-ph}.  
For each domain, the model was trained, validated, and tested on papers from that domain only, using the same retrieval setting as in the main paper: given an abstract, rank all candidate figures from the same paper and compute R@1, DCG@5, and CAR@5.
\cref{tbl:cross_domain} shows in-domain performance.  
The \textit{cond-mat} model achieves the highest scores across all metrics, while \textit{math} and \textit{astro-ph} perform notably worse, suggesting that GA identification is harder in these fields, possibly due to smaller datasets or less standardized figure styles.
We then evaluated each model on the test sets of the other domains without fine-tuning to assess cross-domain transferability.  
\cref{fig:cross_domain} summarizes the results. Models trained on the \texttt{cs} domain generalize best to all other domains, likely because of the larger training scale and more consistent GA formats (e.g., model diagrams, algorithmic flows).  
In contrast, models trained on \textit{math} or \textit{astro-ph} transfer poorly, reflecting narrower visual semantics and limited training data.  
These findings demonstrate that domain coverage and visual diversity in training data strongly affect GA recommendation performance and that cross-domain modeling or adaptive approaches could further improve generalization.


\subsection{Inter-GA Recommendation}
\begin{figure*}[!t]
    \centering
    \begin{minipage}{0.86\textwidth}
        \centering
        \includegraphics[width=\textwidth]{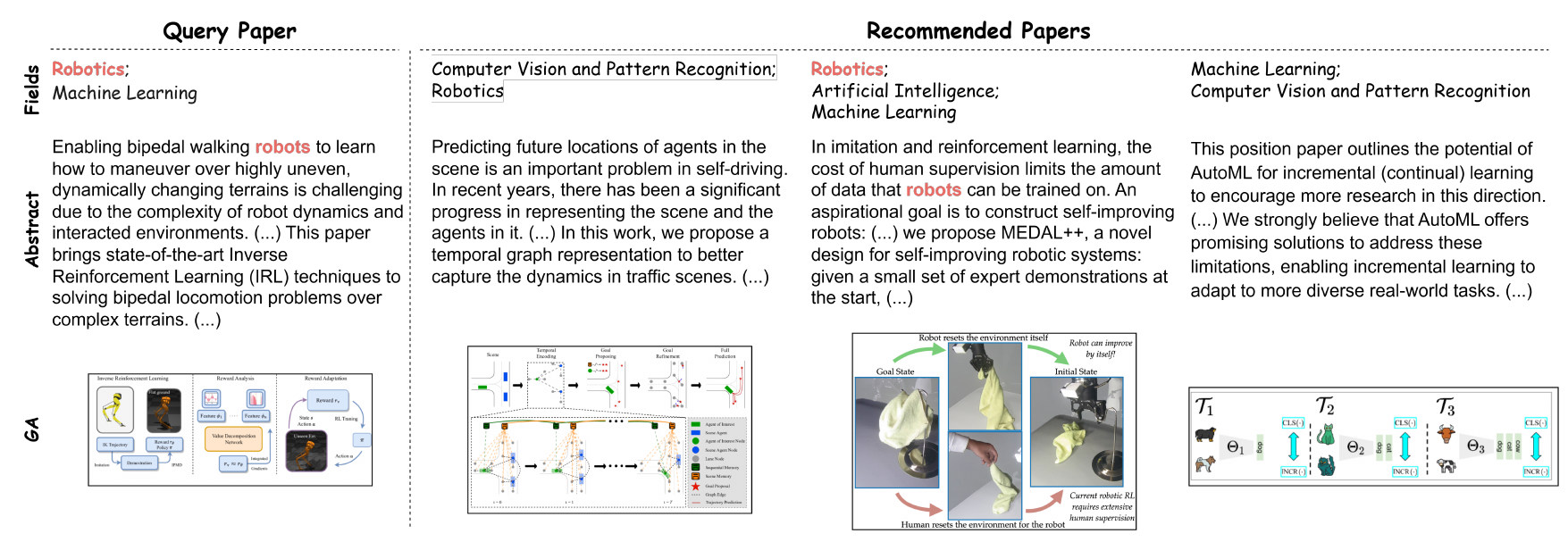}
        \subcaption{(i) Abs2Cap (ROUGE-L) \protect \footnotemark[13]}
        \label{fig:inter_example:a}
    \end{minipage}   
    \begin{minipage}{0.86\textwidth}
        \centering
        \includegraphics[width=\textwidth]{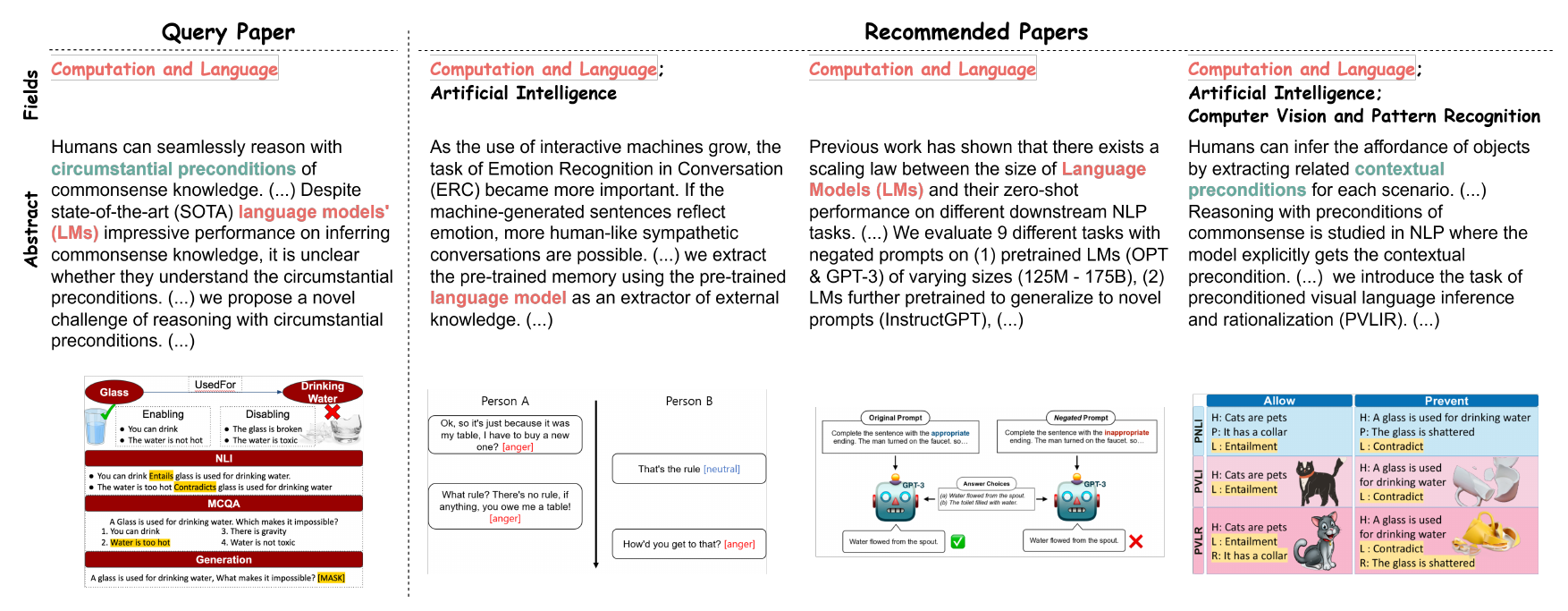}
        \subcaption{(iii) Abs2Fig (CLIP) \protect \footnotemark[14]}
        \label{fig:inter_example:b}
    \end{minipage}   
    \begin{minipage}{0.86\textwidth}
        \centering
        \includegraphics[width=\textwidth]{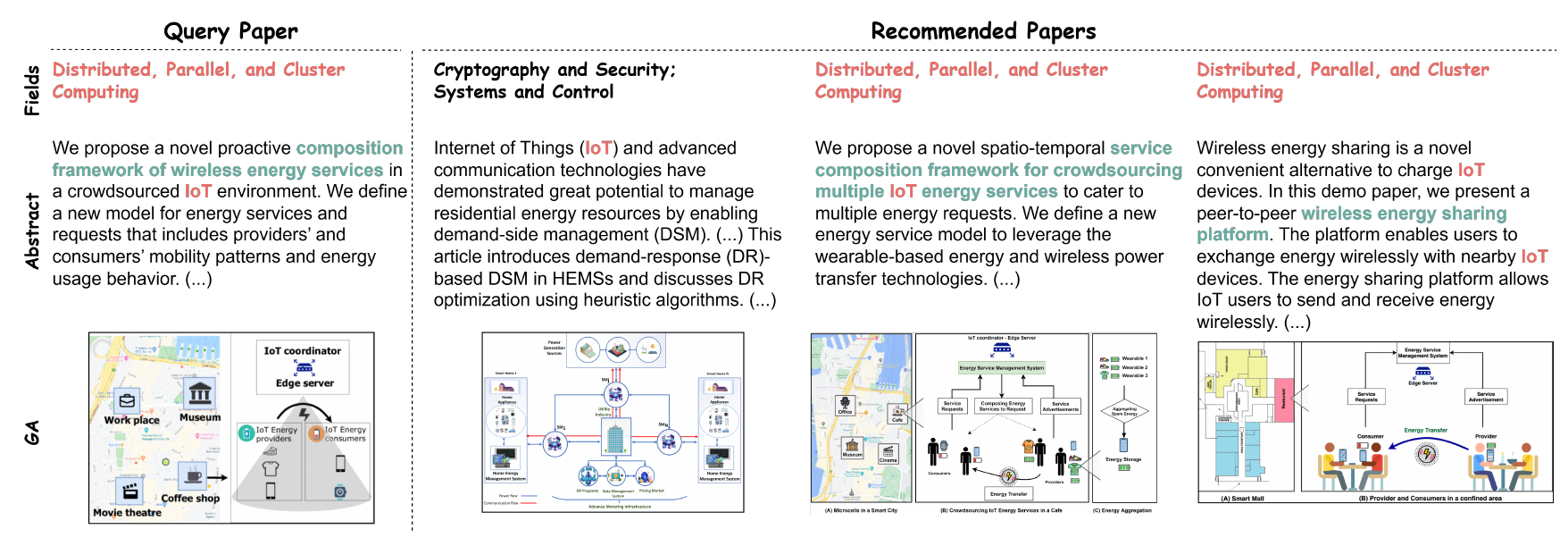}
        \subcaption{(iv) Abs2Fig w/cap (CLIP) \protect \footnotemark[15]}
        \label{fig:inter_example:c}
    \end{minipage}
    \caption{
        Examples of Inter-GA Recommendation results obtained by different methods.
        Pink-highlighted research fields or keywords within abstracts indicate matching primary research categories.
        Green-highlighted phrases denote topic-level relevance.
        These results highlight the different characteristics of the recommendation methods. (a) Abs2Cap (ROUGE-L~\cite{lin2004rouge-l}) produces diverse recommendations, retrieving papers from a broad range of topics. In contrast, (b) Abs2Fig (CLIP) and (c) Abs2Fig w/cap (CLIP) focus on recommending GAs from papers that share similar topics with the query paper, emphasizing strong semantic alignment within the same research domain.
    }
    \label{fig:inter_example}
\end{figure*}

\nbf{Qualitative Evaluation}
\cref{fig:inter_example} presents examples of Inter-GA Recommendation results obtained using different methods.
We observe that methods based on CLIP-like models, such as Abs2Fig and Abs2Fig w/cap, are capable of retrieving GAs from papers that share similar topics with the query.
Notably, this is achieved even though the retrieval process relies solely on the abstract of the query and the GAs of candidate papers, without explicitly incorporating the abstracts or full-texts of the recommended papers into the search context.
While these methods effectively capture topic-level alignment, they may lack the capacity to suggest GAs that introduce surprising or serendipitous ideas from different research areas. Enabling such cross-domain inspiration may require additional mechanisms to intentionally diversify the recommendation results beyond semantic similarity.

\nbf{Quantitative Analysis}
\cref{tbl:inter_results} summarizes the top-$k$ mean and standard deviation along the four axes introduced in \cref{sec:experiments:experimental-setup:inter-GA-recommendation}. 
CLIP-based models exhibit consistently higher Semantic and Visual Coherence, indicating that contrastive encoders effectively retrieve GAs that are both 
topically and visually aligned with the query. 
Long-CLIP achieves the strongest Visual Coherence on average, while caption-based lexical methods (Abs2Cap) remain weak across all axes. 
The variance patterns further distinguish model behaviors: 
random sampling naturally yields the largest diversity, whereas contrastive models produce tighter, more coherent top-$k$ sets. 
These observations confirm that the four axes provide complementary perspectives for characterizing Inter-GA retrieval quality.

\nbf{Robustness Check with DreamSim}
To evaluate the robustness of the Visual Coherence axis, we additionally computed nDCG@$k$ and top-$k$ mean similarity using DreamSim~\cite{fu2023dreamsim}, an alternative perceptual similarity metric.
As shown in \cref{tbl:dreamsim}, results under DreamSim exhibit trends consistent with those obtained from CLIPScore~\cite{hessel2020clipscore}, indicating that the visual consistency of recommended GAs is similarly captured across different perceptual similarity measures.

\section{User Study Interface and Protocol}
\label{sec:appendix:user-study}

\begin{figure*}[!t]
    \centering
    \begin{minipage}{0.49\linewidth}
        \centering
        \includegraphics[width=\textwidth]{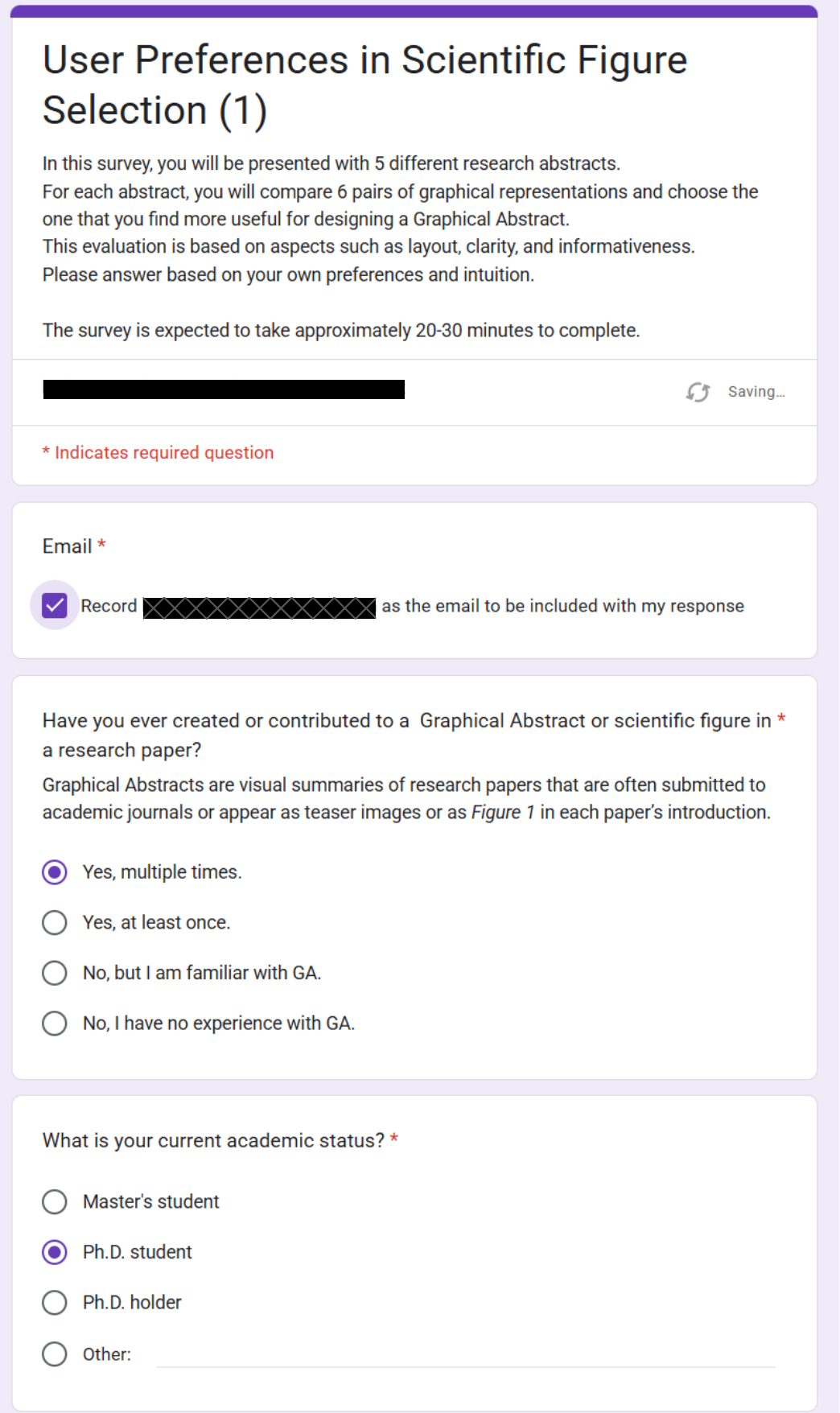}
        \subcaption{}
        \label{fig:user_study_form:a}
    \end{minipage}
    \begin{minipage}{0.43\linewidth}
        \centering
        \includegraphics[width=\textwidth]{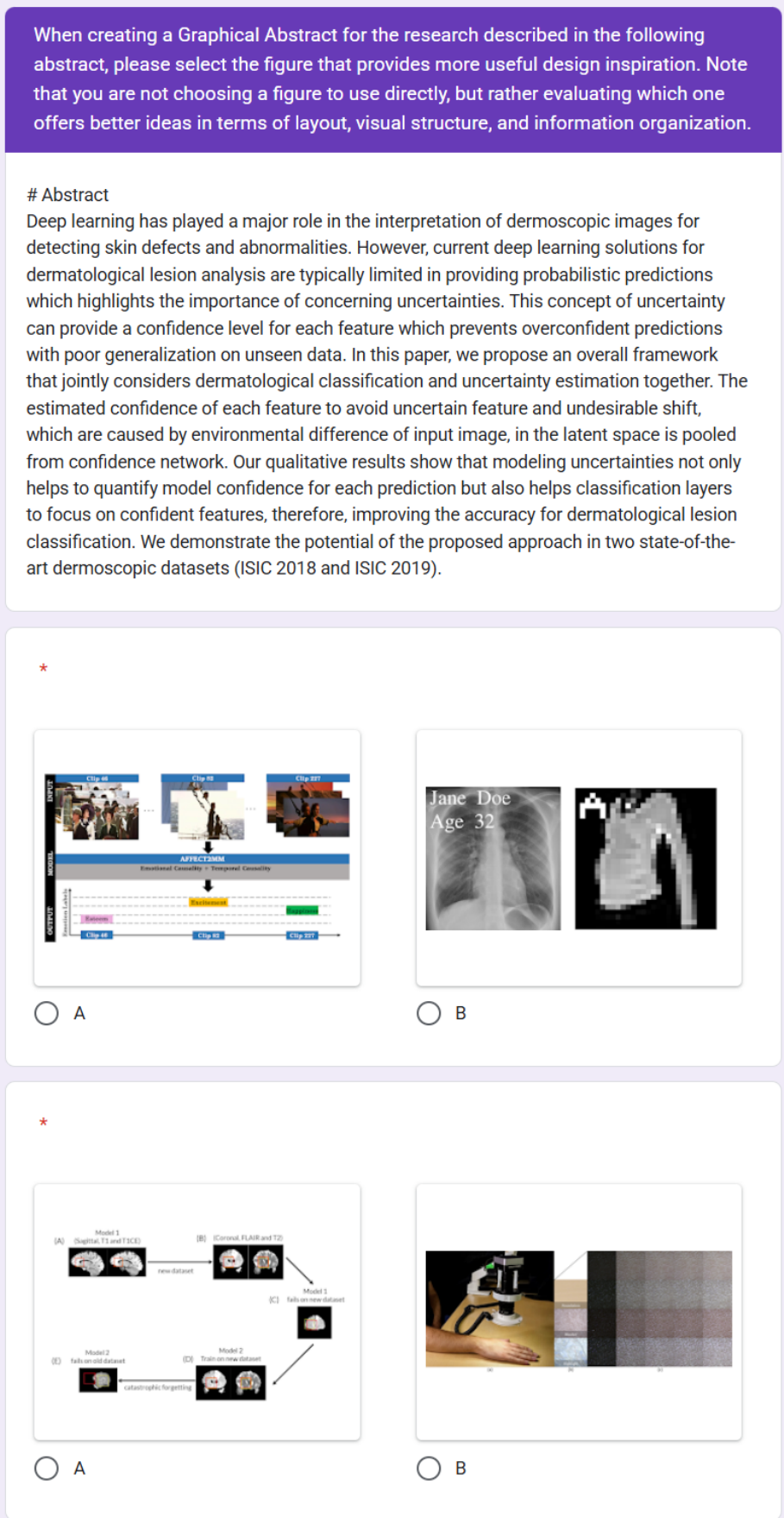}
        \subcaption{}
        \label{fig:user_study_form:b}
    \end{minipage}  
    \caption{
        Screenshot of the questionnaire used in the user study. 
        (a) The introductory section of the questionnaire, asking participants about their prior experience with GAs and their current academic status. 
        (b) Example of the comparative evaluation task. After reading an abstract, participants were presented with pairs of figures recommended by different methods and asked to select the one they found more useful as a design reference when creating a new GA.
    }
    \label{fig:user_study_form}
\end{figure*}

We present the format and participant demographics of our user study, which was conducted via an online questionnaire (Google Form).
\cref{fig:user_study_form} shows the questionnaire format and questions.
The study involved 15 participants: 3 master's students, 6 industry researchers with master’s degrees, 2 Ph.D. students, and 4 Ph.D. holders. 
All participants had prior experience designing GAs or teasers and peer-reviewed publication.

\clearpage
\onecolumn

\footnotetext[9]{
\begin{tabular}[t]{@{}ll@{}}
arXiv ID:
\href{https://arxiv.org/abs/2308.05070}{2308.05070},
\href{https://arxiv.org/abs/2403.17805}{2403.17805} 
\end{tabular}
}
 \clearpage

\footnotetext[10]{
\begin{tabular}[t]{@{}ll@{}}
arXiv ID:
\href{https://arxiv.org/abs/2403.12920}{2403.12920},
\href{https://arxiv.org/abs/2402.04625}{2402.04625} 
\end{tabular}
} \clearpage

\footnotetext[11]{
\begin{tabular}[t]{@{}ll@{}}
arXiv ID:
\href{https://arxiv.org/abs/2403.10988}{2403.10988},
\href{https://arxiv.org/abs/2403.15717}{2403.15717} 
\end{tabular}
} \clearpage

\footnotetext[12]{
\begin{tabular}[t]{@{}ll@{}}
arXiv ID:
\href{https://arxiv.org/abs/2310.17674}{2310.17674},
\href{https://arxiv.org/abs/2311.02848}{2311.02848} 
\end{tabular}
}
 \clearpage
 \clearpage
\begin{table*}[!t]
    \caption{
        In-domain performance of Intra-GA Recommendation models across scientific fields.
        0.5 $\uparrow$ indicates the proportion of semantically justifiable predictions (CAR@5 $>$ 0.5). 
        The \texttt{cond-mat} domain shows the highest CAR@5 values, suggesting that GA selection is more consistent and easier to model in this field.
    }
    \label{tbl:cross_domain}
    \centering
    \small
    \begin{tabular}{lrccccccc}
        \toprule
        \multirow{2}{*}{\textbf{Research Fields}} &
        \multirow{2}{*}{\textbf{Data Size}} &
        \multirow{2}{*}{\textbf{R@1}} &
        \multirow{2}{*}{\textbf{R@2}} &
        \multirow{2}{*}{\textbf{R@3}} &
        \multirow{2}{*}{\textbf{MRR}} &
        \multirow{2}{*}{\textbf{nDCG@5}} &
        \multicolumn{2}{c}{\textbf{CAR@5}} \\
        \cmidrule(lr){8-9}
         & & & & & & & Mean & 0.5 $\uparrow$ \\
        
        \midrule
        cs       & 20,520 & 0.637 & 0.826 & 0.914 & 0.778 & 0.824 & 0.615 & 0.691 \\
        math     &  1,498 & 0.473 & 0.663 & 0.763 & 0.643 & 0.684 & 0.493 & 0.493 \\
        cond-mat &  3,323 & 0.640 & 0.805 & 0.871 & 0.767 & 0.805 & 0.639 & 0.700 \\
        astro-ph &  1,949 & 0.462 & 0.651 & 0.764 & 0.633 & 0.679 & 0.494 & 0.533 \\

        \bottomrule
    \end{tabular}
\end{table*}

 \clearpage

\footnotetext[13]{
\footnotesize
\begin{tabular}[t]{@{}l@{}}
(a) arXiv ID (Query):
\href{https://arxiv.org/abs/2309.16074}{2309.16074}, 
arXiv ID (Recommended):
\href{https://arxiv.org/abs/2207.00255}{2207.00255},
\href{https://arxiv.org/abs/2303.01488}{2303.01488},
\href{https://arxiv.org/abs/2311.11963}{2311.11963}
\end{tabular}
}

\footnotetext[14]{
\footnotesize
\begin{tabular}[t]{@{}l@{}}
(b) arXiv ID (Query): 
\href{https://arxiv.org/abs/2104.08712}{2104.08712},
arXiv ID (Recommended):
\href{https://arxiv.org/abs/2108.11626}{2108.11626},
\href{https://arxiv.org/abs/2209.12711}{2209.12711},
\href{https://arxiv.org/abs/2306.02053}{2306.02053}
\end{tabular}
}

\footnotetext[15]{
\footnotesize
\begin{tabular}[t]{@{}l@{}}
(c) arXiv ID (Query):
\href{https://arxiv.org/abs/2107.12519}{2107.12519},
arXiv ID (Recommended): 
\href{https://arxiv.org/abs/2109.11627}{2109.11627},
\href{https://arxiv.org/abs/2308.09886}{2308.09886},
\href{https://arxiv.org/abs/2208.13506}{2208.13506}
\end{tabular}
}

\begin{table*}[!t]
    \caption{
        Quantitative comparison of various approaches for Inter-GA Recommendation.
        For each method, we report the mean and standard deviation of its top-$k$ recommended GAs under four axes introduced in Sec. 4.1.2:
        Field Match, Semantic Coherence,
        Visual Coherence, and Aesthetic Quality.
        Higher values indicate stronger alignment along each respective axis, and larger standard deviations reflect greater diversity within the top-$k$ set.
        Best scores for each metric are shown in \textbf{bold}, and the highest standard deviations are \underline{underlined}.
    }
    \label{tbl:inter_results}
    \centering
    \small
    \setlength{\tabcolsep}{5pt}
    \resizebox{\textwidth}{!}{
        \begin{tabular}{llcccccccc}
            \toprule
            \multirow{2}{*}{\textbf{Method}} &
            \multirow{2}{*}{\textbf{Backbone}} &
            \multicolumn{2}{c}{\textbf{Field Match@$k$}} & 
            \multicolumn{2}{c}{\textbf{Semantic Coherence@$k$}} &
            \multicolumn{2}{c}{\textbf{Visual Coherence@$k$}} &
            \multicolumn{2}{c}{\textbf{Aesthetics Quality@$k$}} \\

            \cmidrule(lr){3-4} \cmidrule(lr){5-6} \cmidrule(lr){7-8} \cmidrule(lr){9-10}
            
            & &
            top-5 & top-10 &
            top-5 & top-10 &
            top-5 & top-10 &
            top-5 & top-10 \\

            \midrule
            (BL) Random Sampling & -- 
                & $ 0.338 $ & $ 0.345 $
                & $ 0.227 \pm 0.111 $ & $ 0.228 \pm 0.115 $
                & $ \underline{0.545 \pm 0.077} $ & $ \underline{0.545 \pm 0.081} $
                & $ \underline{0.125 \pm 0.248} $ & $ \underline{0.125 \pm 0.253} $ \\

            \cmidrule(lr){1-1}
            \cmidrule(lr){2-2}
            \cmidrule(lr){3-3}
            \cmidrule(lr){4-4}
            \cmidrule(lr){5-5}
            \cmidrule(lr){6-6}
            \cmidrule(lr){7-7}
            \cmidrule(lr){8-8}
            \cmidrule(lr){9-9}
            \cmidrule(lr){10-10}

            \multirow{5}{*}{(i) Abs2Cap}
            & ROUGE-L~\cite{lin2004rouge-l}
                & $ 0.502 $ & $ 0.486 $
                & $ 0.314 \pm 0.114 $ & $ 0.306 \pm 0.118 $
                & $ 0.579 \pm 0.066 $ & $ 0.578 \pm 0.069 $
                & $ 0.136 \pm 0.236 $ & $ 0.136 \pm 0.243 $ \\
                
            & METEOR~\cite{banerjee2005meteor}
                & $ 0.421 $ & $ 0.417 $
                & $ 0.268 \pm 0.110 $ & $ 0.264 \pm 0.112 $
                & $ 0.573 \pm 0.063 $ & $ 0.571 \pm 0.064 $
                & $ 0.130 \pm 0.240 $ & $ 0.130 \pm 0.249 $\\
                
            & CIDEr~\cite{vedantam2015cider}
                & $ 0.438 $ & $ 0.420$
                & $ 0.287 \pm 0.105 $ & $ 0.273 \pm 0.108 $
                & $ 0.579 \pm 0.064 $ & $ 0.577 \pm 0.066 $
                & $ 0.134 \pm 0.237 $ & $ 0.135 \pm 0.243 $ \\
                
            & BM25~\cite{robertson1994bm25}
                & $ 0.704 $ & $ 0.685 $
                & $ 0.489 \pm 0.105 $ & $ 0.468 \pm 0.111 $
                & $ 0.605 \pm 0.072 $ & $ 0.601 \pm 0.074 $
                & $ 0.151 \pm 0.233 $ & $ 0.148 \pm 0.240 $ \\
                
            & BERTScore~\cite{zhang2020bertscore}
                & $ 0.549 $ & $ 0.545 $
                & $ 0.360 \pm 0.107 $ & $ 0.351 \pm 0.109 $
                & $ 0.580 \pm 0.069 $ & $ 0.578 \pm 0.071 $
                & $ 0.147 \pm 0.232 $ & $ 0.145 \pm 0.239 $ \\
                
            \cmidrule(lr){1-1}
            \cmidrule(lr){2-2}
            \cmidrule(lr){3-3}
            \cmidrule(lr){4-4}
            \cmidrule(lr){5-5}
            \cmidrule(lr){6-6}
            \cmidrule(lr){7-7}
            \cmidrule(lr){8-8}
            \cmidrule(lr){9-9}
            \cmidrule(lr){10-10}
            
            \multirow{6}{*}{(iii) Abs2Fig} 
            & CLIP~\cite{radford2022clip}
                & $ 0.729 $ & $ 0.719 $
                & $ 0.455 \pm 0.105 $ & $ 0.444 \pm 0.109 $ 
                & $ 0.646 \pm 0.054 $ & $ 0.642 \pm 0.057 $
                & $ \mathbf{0.160 \pm 0.220} $ & $ \mathbf{0.162 \pm 0.227} $ \\

            & BLIP-2~\cite{li2023blip-2}
                & $ 0.683 $ & $ 0.674 $
                & $ 0.419 \pm 0.110 $ & $ 0.410 \pm 0.114 $ 
                & $ 0.622 \pm 0.063 $ & $ 0.620 \pm 0.065 $
                & $ 0.152 \pm 0.228 $ & $ 0.153 \pm 0.235 $ \\ 

            & X$\smash{{}\scriptstyle ^2}$-VLM~\cite{zeng2023x2-vlm}  
                & $ 0.418 $ & $ 0.402 $
                & $ \underline{0.263 \pm 0.116} $ & $ \underline{0.257 \pm 0.122} $ 
                & $ 0.461 \pm 0.032 $ & $ 0.451 \pm 0.033 $
                & $ 0.133 \pm 0.206 $ & $ 0.122 \pm 0.203 $ \\
                
            & OpenCLIP~\cite{cherti2023open-clip}
                & $ 0.720 $ & $ 0.710 $
                & $ 0.451 \pm 0.106 $ & $ 0.440 \pm 0.109 $ 
                & $ 0.632 \pm 0.058 $ & $ 0.630 \pm 0.061 $
                & $ 0.159 \pm 0.221 $ & $ 0.157 \pm 0.229 $ \\

            & SigLIP2~\cite{tschannen2025siglip2}
                & $ 0.631 $ & $ 0.620 $
                & $ 0.387 \pm 0.110 $ & $ 0.381 \pm 0.114 $
                & $ 0.598 \pm 0.065 $ & $ 0.597 \pm 0.068 $
                & $ 0.142 \pm 0.229 $ & $ 0.144 \pm 0.239 $ \\
                
            & Long-CLIP~\cite{zhang2024long-clip}
                & $ 0.726 $ & $ 0.717 $
                & $ 0.456 \pm 0.108 $ & $ 0.445 \pm 0.103 $ 
                & $ \mathbf{0.648 \pm 0.056} $ & $ \mathbf{0.644 \pm 0.060} $
                & $ 0.159 \pm 0.221 $ & $ 0.160 \pm 0.228 $ \\

            \cmidrule(lr){1-1}
            \cmidrule(lr){2-2}
            \cmidrule(lr){3-3}
            \cmidrule(lr){4-4}
            \cmidrule(lr){5-5}
            \cmidrule(lr){6-6}
            \cmidrule(lr){7-7}
            \cmidrule(lr){8-8}
            \cmidrule(lr){9-9}
            \cmidrule(lr){10-10}
            
            \multirow{6}{*}{(iv) Abs2Fig w/cap} 
            & CLIP~\cite{radford2022clip}
                & $ \mathbf{0.755} $ & $ \mathbf{0.742} $
                & $ 0.493 \pm 0.098 $ & $ 0.479 \pm 0.101 $ 
                & $ 0.614 \pm 0.067 $ & $ 0.611 \pm 0.071 $
                & $ 0.152 \pm 0.229 $ & $ 0.152 \pm 0.237 $ \\

            & BLIP-2~\cite{li2023blip-2}
                & $ 0.647 $ & $ 0.639 $
                & $ 0.390 \pm 0.105 $ & $ 0.382 \pm 0.109 $ 
                & $ 0.597 \pm 0.067 $ & $ 0.596 \pm 0.068 $
                & $ 0.147 \pm 0.228 $ & $ 0.149 \pm 0.236 $ \\ 

            & X$\smash{{}\scriptstyle ^2}$-VLM~\cite{zeng2023x2-vlm}
                & $ 0.415 $ & $ 0.399 $
                & $ 0.254 \pm 0.114 $ & $ 0.250 \pm 0.119 $ 
                & $ 0.555 \pm 0.067 $ & $ 0.552 \pm 0.072 $
                & $ 0.143 \pm 0.213 $ & $ 0.140 \pm 0.226 $ \\
                
            & OpenCLIP~\cite{cherti2023open-clip}
                & $ 0.749 $ & $ 0.737 $
                & $ 0.489 \pm 0.097 $ & $ 0.475 \pm 0.100 $ 
                & $ 0.615 \pm 0.066 $ & $ 0.611 \pm 0.069 $
                & $ 0.154 \pm 0.231 $ & $ 0.150 \pm 0.237 $ \\

            & SigLIP2~\cite{tschannen2025siglip2}
                & $0.235 $ & $ 0.336 $
                & $0.186 \pm 0.128 $ & $ 0.212 \pm 0.129 $
                & $0.462 \pm 0.076 $ & $ 0.467 \pm 0.079 $
                & $0.134 \pm 0.253 $ & $ 0.117 \pm 0.247 $ \\
            
            & Long-CLIP~\cite{zhang2024long-clip}
                & $ 0.753 $ & $ 0.737 $
                & $ \mathbf{0.498 \pm 0.098} $ & $ \mathbf{0.482 \pm 0.103} $  
                & $ 0.614 \pm 0.070 $ & $ 0.611 \pm 0.073 $
                & $ 0.148 \pm 0.231 $ & $ 0.145 \pm 0.240 $ \\
                
            \bottomrule
        \end{tabular}
    }
\end{table*}

\begin{table*}[!t]
    \caption{
        Additional quantitative results for Inter-GA Recommendation using DreamSim as an alternative perceptual similarity metric.
        We report (1) nDCG@$k$ using DreamSim-based pseudo-relevance labels, and (2) the mean and standard deviation of DreamSim similarities over the top-$k$ retrieved GAs.
        These results examine robustness to the choice of visual similarity metric.
        Best values are shown in \textbf{bold}, and the highest standard deviations are \underline{underlined}. 
    }
    \label{tbl:dreamsim}

    \centering
    \small
    \setlength{\tabcolsep}{5pt}
    \resizebox{0.65\textwidth}{!}{
        \begin{tabular}{llccccc}
            \toprule
            \multirow{2}{*}{\textbf{Method}} &
            \multirow{2}{*}{\textbf{Backbone}} &
            \multicolumn{3}{c}{\textbf{nDCG@$k$ (DreamSim)}} &
            \multicolumn{2}{c}{\textbf{DreamSim@$k$}} \\
            \cmidrule(lr){3-5} \cmidrule(lr){6-7}
            & &
            top-5 & top-10 & top-30 &
            top-5 & top-10 \\

            \midrule
            
            (BL) Random Sampling & -- 
                & $ 0.679 $ & $ 0.711 $ & $ 0.803 $
                & $ \underline{0.382 \pm 0.086} $ & $ 0.382 \pm 0.090 $ \\
            
            \cmidrule(lr){1-1}
            \cmidrule(lr){2-2}
            \cmidrule(lr){3-3}
            \cmidrule(lr){4-4}
            \cmidrule(lr){5-5}
            \cmidrule(lr){6-6}
            \cmidrule(lr){7-7}
            
            \multirow{5}{*}{(i) Abs2Cap}
            & ROUGE-L~\cite{lin2004rouge-l}
                & $ 0.722 $ & $ 0.750 $ & $ 0.830 $
                & $ 0.417 \pm 0.082 $ & $ 0.415 \pm 0.086 $ \\
                
            & METEOR~\cite{banerjee2005meteor}
                & $ 0.749 $ & $ 0.777 $ & $ 0.851 $
                & $ 0.433 \pm 0.067 $ & $ 0.431 \pm 0.071 $ \\
                
            & CIDEr~\cite{vedantam2015cider}
                & $ 0.742 $ & $ 0.771 $ & $ 0.847 $
                & $ 0.425 \pm 0.072 $ & $ 0.425 \pm 0.075 $ \\
                
            & BM25~\cite{robertson1994bm25}
                & $ 0.742 $ & $ 0.767 $ & $ 0.841 $
                & $ 0.443 \pm 0.081 $ & $ 0.438 \pm 0.083 $ \\
                
            & BERTScore~\cite{zhang2020bertscore}
                & $ 0.704 $ & $ 0.735 $ & $ 0.821 $
                & $ 0.406 \pm 0.083 $ & $ 0.406 \pm 0.086 $ \\
                
            \cmidrule(lr){1-1}
            \cmidrule(lr){2-2}
            \cmidrule(lr){3-3}
            \cmidrule(lr){4-4}
            \cmidrule(lr){5-5}
            \cmidrule(lr){6-6}
            \cmidrule(lr){7-7}
            
            \multirow{6}{*}{(iii) Abs2Fig} 
            & CLIP~\cite{radford2022clip}
                & $ \mathbf{0.753} $ & $ \mathbf{0.778} $ & $ \mathbf{0.853} $
                & $ 0.456 \pm 0.072 $ & $ 0.452 \pm 0.075 $ \\

            & BLIP-2~\cite{li2023blip-2}
                & $ 0.752 $ & $ \mathbf{0.778} $ & $ \mathbf{0.853} $
                & $ 0.456 \pm 0.070 $ & $ 0.453 \pm 0.073 $ \\

            & X$\smash{{}\scriptstyle ^2}$-VLM~\cite{zeng2023x2-vlm}  
                & $ 0.438 $ & $ 0.437 $ & $ 0.652 $
                & $ 0.245 \pm 0.036 $ & $ 0.232 \pm 0.037 $ \\
                
            & OpenCLIP~\cite{cherti2023open-clip}
                & $ 0.748 $ & $ 0.776 $ & $ 0.851 $
                & $ 0.453 \pm 0.070 $ & $ 0.451 \pm 0.073 $ \\
                
            & SigLIP2~\cite{tschannen2025siglip2}
                & $ 0.726 $ & $ 0.756 $ & $ 0.838 $
                & $ 0.432 \pm 0.073 $ & $ 0.431 \pm 0.076 $ \\
                
            & Long-CLIP~\cite{zhang2024long-clip}
                & $ 0.752 $ & $ \mathbf{0.778} $ & $ 0.852 $
                & $ \mathbf{0.457 \pm 0.072} $ & $ \mathbf{0.454 \pm 0.075} $ \\
                
            \cmidrule(lr){1-1}
            \cmidrule(lr){2-2}
            \cmidrule(lr){3-3}
            \cmidrule(lr){4-4}
            \cmidrule(lr){5-5}
            \cmidrule(lr){6-6}
            \cmidrule(lr){7-7}
            
            \multirow{6}{*}{(iv) Abs2Fig w/cap} 
            & CLIP~\cite{radford2022clip}
                & $ 0.739 $ & $ 0.766 $ & $ 0.843 $
                & $ 0.447 \pm 0.077 $ & $ 0.443 \pm 0.080 $ \\

            & BLIP-2~\cite{li2023blip-2}
                & $ 0.764 $ & $ 0.789 $ & $ 0.860 $
                & $ 0.453 \pm 0.065 $ & $ 0.450 \pm 0.069 $ \\

            & X$\smash{{}\scriptstyle ^2}$-VLM~\cite{zeng2023x2-vlm}
                & $ 0.697 $ & $ 0.726 $ & $ 0.816 $
                & $ 0.387 \pm 0.074 $ & $ 0.385 \pm 0.078 $ \\
                
            & OpenCLIP~\cite{cherti2023open-clip}
                & $ 0.744 $ & $ 0.772 $ & $ 0.847 $
                & $ 0.449 \pm 0.074 $ & $ 0.446 \pm 0.077 $ \\

            & SigLIP2~\cite{tschannen2025siglip2}
                & $ 0.460 $ & $ 0.513 $ & $ 0.680 $
                & $ 0.255 \pm 0.073 $ & $ \underline{0.270 \pm 0.103} $ \\
            
            & Long-CLIP~\cite{zhang2024long-clip}
                & $ 0.738 $ & $ 0.765 $ & $ 0.841 $
                & $ 0.447 \pm 0.079 $ & $ 0.443 \pm 0.081 $ \\

            \bottomrule
        \end{tabular}
    }
\end{table*}

\end{document}